\documentclass[12pt]{article}
\usepackage{amsmath}
\usepackage{times}
\usepackage{graphicx,subcaption}
\usepackage{color}
\usepackage{multirow}
\usepackage{amsfonts}
\usepackage{amssymb}
\usepackage{apacite}
\usepackage{caption}

\usepackage[authoryear]{natbib}

\usepackage{algorithm}
\usepackage{algpseudocode}% http://ctan.org/pkg/algorithmicx
\usepackage{tabularx,booktabs}
\newcolumntype{C}{>{\centering\arraybackslash}X} % centered version of "X" type
\usepackage{lipsum}
\usepackage{etoolbox}
\usepackage{tikz}
\usetikzlibrary{tikzmark}
\usetikzlibrary{calc}
\usepackage[noadjust]{cite}
\usepackage{upgreek}%\usepackage{subcaption}

\usepackage{rotating}
\usepackage{bbm}
\usepackage{latexsym}
\newtheorem{definition}{Definition}

\newtheorem{conjecture}{Conjecture}
\newcommand{\captionfonts}{\normalsize}

\makeatletter  
\long\def\@makecaption#1#2{%
  \vskip\abovecaptionskip
  \sbox\@tempboxa{{\captionfonts #1: #2}}%2
  \ifdim \wd\@tempboxa >\hsize
    {\captionfonts #1: #2\par}
  \else
    \hbox to\hsize{\hfil\box\@tempboxa\hfil}%
  \fi
  \vskip\belowcaptionskip}
\makeatother

\begin{document}
\hspace{13.9cm}1

\ \vspace{20mm}\\

{\LARGE Towards a Kernel based Uncertainty Decomposition Framework for Data and Models}

\ \\
{\bf \large Rishabh Singh}\\
{\bf \large Jose C. Principe}\\
 %$^{\displaystyle 1, \displaystyle 2}$}\\
{Computational NeuroEngineering Laboratory, Department of Electrical and Computer Engineering, University of Florida, Gainesville, FL 32611, USA}\\
%{$^{\displaystyle 2}$Your second affiliation.}\\

%\ \\[-2mm]
{\bf Keywords:} Model uncertainty, RKHS, point-prediction, neural networks, quantum physics, moment decomposition
\thispagestyle{empty}
\markboth{}{NC instructions}
\ \vspace{-0mm}\\
\begin{center} {\bf Abstract} \end{center}
This paper introduces a new framework for quantifying predictive uncertainty for both data and models that relies on projecting the data into a Gaussian reproducing kernel Hilbert space (RKHS) and transforming the data probability density function (PDF) in a way that quantifies the flow of its gradient as a topological potential field quantified at all points in the sample space. This enables the decomposition of the PDF gradient flow by formulating it as a moment decomposition problem using operators from quantum physics, specifically the Schr\"odinger's formulation.  We experimentally show that the higher order modes systematically cluster the different tail regions of the PDF, thereby providing unprecedented discriminative resolution of data regions having high epistemic uncertainty. In essence, this approach decomposes local realizations of the data PDF in terms of uncertainty moments. We apply this framework as a surrogate tool for predictive uncertainty quantification of point-prediction neural network models, overcoming various limitations of conventional Bayesian based uncertainty quantification methods. Experimental comparisons with some established methods illustrate performance advantages exhibited by our framework.

\section{Introduction}
\subsection{Information Theory: Physics based Perspective}
The modern foundation of information theory lies in its ability to quantify uncertainty in a random variable $X$ using entropy (Shannon), which was later generalized by Renyi \citep{r2} among many others. It has become an indispensable tool in communication theory and machine learning as a predictor of value of information to optimize design engineering systems and as a descriptor for density estimation and other statistical evaluations that attempt to characterize the intrinsic generating functions of data \citep{r3, r4, r5}. There was an earlier definition of information in statistics proposed by Sir Ronald Fisher in the context of parameter estimation \citep{fish}, which measures the amount of information that an observable random variable $X$ carries about an unknown parameter $\theta$ of a distribution that models $X$. In statistical mechanics, Boltzmann’s entropy is the logarithm of the number of arrangements that a system can be configured in and it still remains consistent with the thermodynamic observables \citep{bol}. Boltzmann and Shannon information share exactly the same mathematical formula. Fisher information \citep{fish} has been well regarded as the cornerstone concept in measuring the gain of information from data and in quantifying the \textit{order} of a system \citep{roy1}, instead of disorder as is done by entropy. This presents a strong physics based interpretation of data analysis \citep{roy2}, and it has been used to formulate quantum physical models of stock market data in the field of econophysics \citep{q1, q2}. More recently in quantum computing, Von Neumann's quantum entropy \citep{vonn}, which is based on Shannon entropy, is considered to be the cornerstone to design quantum computer algorithms and communication channels. The Belavkin equation \citep{bel} is a stochastic differential equation describing the dynamics of a quantum system undergoing observation in continuous time, which reduces to Schr\"odinger's equation in special conditions. As this brief review shows, there is a conceptual equivalence in modeling information, both from the statistical mechanics and the machine learning perspectives that served as inspiration for our framework.

\subsection{Paper Structure}
Because of the novelty of the approach, we start the paper with a motivation and high level description of the methodology and contributions. The technical part starts in section 2 with the definition of the kernel mean embedding (KME) theory and its extension to the potential field interpretation of data PDF in the sample space. In section 3, we formulate a functional operator on the PDF using the Schr\"odinger's equation to define the quantum field potential of the data. In section 4, we show how to perform a moment decomposition of the quantum field potential to extract its uncertainty modes. Section 5 presents a step-by-step summary of the entire proposed framework. We first analyze our framework from a completely data driven perspective (without involving a model) and provide pedagogical examples in section 6 using implementation of the framework on time-series signals, which provides an intuitive understanding of how it characterizes data PDF. We subsequently present the application of the framework for quantifying predictive uncertainties of neural network models in section 7 where we also discuss how the proposed framework differs from existing Bayesian methods in terms of scalability, precision and computational cost. Related experimental results are provided in section 8 which initially consist of examples illustrating advantages offered by our framework compared to established methods such as Monte Carlo dropout and GP regression. We also present comparison results on a transfer learning application to evaluate scalability of our framework. Further results on some benchmark datasets provide quantified evidence of our framework's performance advantages.

\subsection{Motivation and Contributions}
\subsubsection*{Data Uncertainty Viewpoint}
In most real-world scenarios, observed signals are generated by systems controlled by a multitude of source processes and noise resulting in very complicated data dynamics. Current machine learning models and information theoretic divergence measures characterize the global data structure but fail to effectively characterize local uncertainties associated with such data. Quantum based formulations in physics, on the other hand, have been well known for providing high resolution multi-scale characterizations of system dynamics. This is achieved through a stochastic description of the system in terms of energy modes in a Hilbert space. The key quality of such formulations is their non-parametric nature and a complete multi-scale quantification of modes, leading to a description of the system at all points in space. We hypothesize that extending information theoretic measures in terms of such physics-based formulations could yield similar advantages in the characterization of data and models, thereby providing a multi-scale enhanced view of their distributions. This inspires us to develop nonparametric data characterization tools that utilize operators from physics to decompose PDF information while making minimal assumptions. Our conjecture is that the best way to achieve this goal is to begin with non-homogenous, local, \textit{data-induced} metric spaces that would be highly sensitive towards signal characterization \citep{corr}. Perhaps the best contender for this metric space is the Gaussian reproducing kernel Hilbert space (RKHS) which has been well established to provide universal characterization of data \citep{parze, gp, g1, g2}. Moreover, projecting data into an RKHS transforms them into Gaussian functions centered at the data coordinates, which obey the properties of a potential field \citep{prin}. Hence the RKHS makes it possible to obtain physical formulations of data properties with simplicity because of the uses of the Hilbert space.\par

 \begin{figure}[!b]
  \centering
   \includegraphics[scale=0.6]{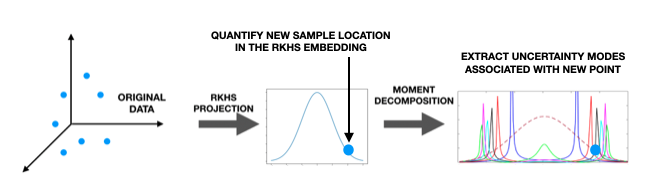}
    \caption{Basic depiction of the proposed framework}
    \label{fr1}
  \end{figure}

Therefore, towards the goal of effectively quantifying data uncertainty, we introduce an RKHS based information theoretic framework that utilizes a physical interpretation of the data space to extract its various uncertainty moments. This methodology has solid foundations because it guarantees, from the kernel mean embedding (KME) theory \citep{emb}, accurate estimations of the data PDF as a functional in the RKHS. We interpreted KME as a potential field and called it the information potential field (IPF), where there is a value associated with each data point in the feature space \citep{prin}. Here we extend this interpretation by taking the normalized Laplacian of the IPF over the space of samples to emphasize local contributions to the IPF. The Laplacian of the IPF is inspired by Schr\"odinger's equation and represents the gradient flow of density in the Hilbert space, hence the name quantum information potential field (QIPF). As in quantum physics, one can perform a moment decomposition of the QIPF using Hermite polynomials and obtain multi-scale projections of the QIPF at every point in space. Our uncertainty framework is depicted in fig. \ref{fr1} and can be summarized in terms of the following key steps.
\begin{enumerate}
\item \textbf{Field based interpretation of data PDF}: Projection of samples into the Gaussian RKHS to obtain a functional description of the data PDF over the space of the samples interpreted as the IPF.
\item \textbf{Quantifying the gradient flow of the IPF}: Use Schr\"odinger's equation to locally quantify the gradient flow of the IPF over the function space, which we called the QIPF.
\item \textbf{PDF Mode Extraction}: Solve the QIPF moment decomposition problem (formulated by the Schr\"odinger's equation) using Hermite polynomial projections to extract various moments of the QIPF that provide a high resolution selective quantification of uncertainty at different regions of the data distribution.
\end{enumerate}
 
Such a framework presents an entirely new characterization of the data that enables one to quantify the flow across local regions of the PDF by using only a few samples as initial reference. This is not possible by current statistical and Bayesian based methodologies. It also makes the framework very suitable for applications on online streaming data. Furthermore, the reliance of the framework on the Gaussian RKHS enables it to intrinsically extract a rich statistical representation of the data. The physics-based decomposition of the PDF results in high resolution, multi-scale descriptors of anisotropy, with the higher order moments being increasingly sensitive to samples in the tail regions of the distribution (where epistemic uncertainty is maximum). This is in stark contrast with the traditional moment decomposition methods of the PDF which describe central moments.

\subsubsection*{Model Uncertainty Viewpoint} 
Deep learning neural network models have made remarkable progress over the past two decades in a large variety of machine learning applications \citep{lec}. However, despite their success, such models only quantify the expected value of the model output, given the input, and do not provide any information related to their prediction uncertainties. This information is crucial for engineering applications and in sensitive application arenas such as personalized medicine and autonomous driving, especially given how prone neural networks are towards overfitting. Moreover, there have also been alarming revelations regarding the high susceptibility of such models towards adversarial attacks \citep{su, adv}.\par

We are therefore motivated to explore the utility of our proposed framework for the predictive uncertainty quantification of deep learning models. The main idea here is to find the uncertain regions in the input-output mapping distribution learnt by the model. In this case, instead of working on data PDF, we propose an uncertainty decomposition of the distribution learnt by the model parameters. One way to achieve this on a sample-by-sample basis is by evaluating the cross-entropy between the instantaneous test set outputs (learnt distribution realizations) of the internal layers and final layer (model output). The cross-entropy can be directed estimated from the cross information potential \citep{prin}, making all the proposed quantum mode decomposition directly applicable to model uncertainty quantification. Before delving into more details, we first review some established methods of uncertainty quantification (UQ).\par

Existing UQ methods can be broadly classified into \textit{forward UQ} and \textit{inverse UQ} methods from an implementation point of view \citep{uncdef1, uncdef2}. In uncertainty propagation (forward UQ), one attempts to directly characterize the model output uncertainty distribution from the implicit uncertainties present in the parameters. Inverse UQ, on the other hand, attempts to quantify uncertainty distributions over model parameters. In the context of machine learning, most of the focus has been on the latter category with a domination of Bayesian based inferencing methods which offer the most mathematically grounded approach to quantify model uncertainty by learning probability distributions of model weights \citep{mack, neal, bishop}. Early development of Bayesian based models revolved around Laplacian approximation \citep{mack}, Hamiltonian Monte Carlo \citep{neal}, Markov-chain Monte Carlo (MCMC) based Bayesian neural networks \citep{bishop}. Although such methods offer a principled approach of quantifying model uncertainty mainly by marginalizing over model parameters, they involve prohibitive computational costs and lack scalability towards large data and model architectures. Most of the recent work in this field has therefore been related to developing faster variational inference based approaches that offer more efficient ways of training Bayesian neural networks (BNNs) \citep{graves, jord, hof}. The high parameter dimensionality and the complexity of weight associations in modern neural networks still makes it very difficult for such variational inference approaches to adequately capture parameter dependencies \citep{projbnn}. Other methods involve surrogate modeling techniques that exploit the input-output mapping learnt by the model \citep{nag, sur1, sur2}. Here, computationally cheap approximations of models are used for easier extraction of the relevant information related to model uncertainty. Forward UQ methods include ensemble based methods where multiple instances of models with different initializations are trained on noisy data and the result is the aggregation of all model outputs \citep{tib, osb, pearce}. The variations of the results provide the necessary uncertainty information. A notable related work is that of Lakshminarayan \citep{laks} where authors use ensemble neural networks to implement forward UQ. Recent work of Gal and Ghahramani \citep{gal} has gained increased popularity due to its simplicity and effectiveness in quantifying predictive uncertainty. Here, authors propose Monte Carlo dropout where multiple instantiations of dropout are used during testing of models to obtain the uncertainty intervals associated with the model predictions. A critical disadvantage of this method in real-time applications is the requirement for multiple forwards passes during testing, which is not feasible for very large networks used in modern datasets.\par

\begin{figure}[!t]
  \centering
   \includegraphics[scale=0.6]{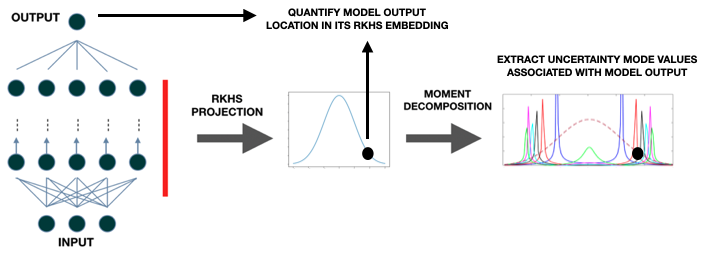}
    \caption{Implementation of proposed framework on a neural network.}
    \label{fr2}
  \end{figure}

We advocate an approach for predictive uncertainty quantification that is non-intrusive to the training process of a traditional deep learning model and relies solely on extracting information from the internal parameter distribution of the \textit{trained} model with respect to its output. In this regard, we hypothesize that the application of our framework as a forward UQ method could be advantageous. The idea is to create an alternate representation (or an embedding) of the model that quantifies epistemic uncertainty based on the location of its prediction with respect to its learnt parameter PDF, with the uncertainty being relatively higher for output locations closer to the tails of the parameter PDF. Towards this end, we utilize our framework as a surrogate uncertainty quantifier of a trained neural network that decomposes the model's parameter PDF (input-output mapping), realized at every test instance, in terms of uncertainty modes that provide a high resolution location of the model's output with respect to its PDF. The implementation is depicted in fig. \ref{fr2}. From a physics based analogical perspective, one can visualize the framework's representation of the model as a drum membrane and the output as a drumstick hitting the membrane during each test cycle. The resultant modes of membrane vibration (uncertainty modes of the model PDF) , that depend on where exactly the membrane was hit, quantify the uncertainties of the model with respect to the output. We submit various advantages offered by our framework over traditional model uncertainty quantification paradigms. 
 \begin{itemize}
 	\item Our framework is non-intrusive to the model's training process and provides a \textit{single-shot} estimation of the model's epistemic uncertainty during testing, unlike variational inference based methods that are dependent on random sampling of the model.
 	\item Instead of estimating central moments of uncertainty as is done by Bayesian methods, the proposed framework extracts high resolution \textit{local uncertainty moments} at every point in the sample space that better enhance the quantification of data regions unknown to the model.
 	\item We further posit that our framework quantifies epistemic uncertainty with much greater precision and scales better to larger models in modern applications than variational inference based methods.
 \end{itemize} 

\section{Quantifying Data PDF in the RKHS}
Functional statistics have a long history, starting with the work of Karhunen and Loeve, Grenander and Rao \citep{kar, loe, gre, rao}. In the early days the big practical issue was the infinite dimensionality of the space, which was conquered with subspace projections.
\subsection{Kernel Mean Embedding}
Kernel methods have been very popular and well established in the field of machine learning \citep{smola}. The crux of their success is largely owed to a powerful property of the reproducing kernel Hilbert space (RKHS) associated with positive definite kernels called the ``kernel trick" \citep{aron}, which allows one to pose any problem in an input set $X$ as a linear-algebraic problem in its RKHS, $H$, with a non-linear transformation (embedding) of $X$ into $H$ induced by a kernel $k: X $ x $ X \rightarrow R$. In other words, the RKHS, constructed by an appropriate kernel, allows one to simplify any non-linear relationship in an input space as a linear expression in a higher dimensional space. This property has led to the advent of many popular kernel based algorithms in machine learning \citep{hoff, jp}. Following similar intuition, another elegant property of the RKHS is the theory of kernel mean embedding (KME) which allows one to non-parametrically quantify a data distribution from the input space as an element of its associated RKHS \citep{embor}. For a detailed explanation of the metric, we refer the reader to \citep{emb}. Its definition is summarized as follows.

\begin{definition}[Kernel Mean Embedding]
 Suppose that the space $\mathcal{Z(X)}$ consists of all probability measures $\mathbb{P}$ on a measurable space $(\mathcal{X},\Sigma)$. The kernel mean embedding of probability measures in $\mathcal{Z(X)}$ into an RKHS denoted by $\mathcal{H}$ and characterized with a reproducing kernel $k : X \times X \rightarrow \mathbb{R}$ is defined by a mapping 
 \begin{equation*}
 \mu: \mathcal{Z(X)} \rightarrow	 \mathcal{H}, \ \ \mathbb{P} \mapsto \int k(x, .)d\mathbb{P}(x)
 \label{5}
 \end{equation*}
\end{definition}
Hence the kernel mean embedding (KME) represents the probability distribution in terms of a mean function by utilizing the kernel feature map in the space of the distribution. In other words,
\begin{equation}
\phi(\mathbb{P}) = \mu_{\mathbb{P}} = \int k(x, .)d\mathbb{P}(x)	
\label{kl}
\end{equation}
There are several useful properties associated with the KME. For instance, it is injective for characteristic kernels, meaning that $\mu_{\mathbb{P}} = \mu_{\mathbb{Q}}$ only when $\mathbb{P} = \mathbb{Q}$, thus allowing for unique characterizations of data distributions. It also makes minimal assumptions on the data generating process and enables extensions of most learning algorithms in the space of probability distributions.\par 
  
%\subsection{Empirical Estimate}
In most real world applications, there is no information available regarding the nature of $\mathbb{P}$. One must therefore resort to empirical estimation of the KME. The simplest method of empirically computing the KME is by computing its unbiased estimate given by
\begin{equation}
\hat\mu = \frac{1}{n}\sum_{t=1}^{n}k(x_t,.).
\label{6}
\end{equation}
Here $\hat\mu$ converges to $\mu$ for $n \rightarrow \infty$, in concordance with the law of large numbers.  One can intuit that the empirical KME is also a result of the general Dirac formulation assigning a mass of $1/n$ to every data sample. This also gives rise to the interpretation of the empirical KME as an instance of a point process \citep{emb}.\par

\subsection{Potential Field Interpretation}
Another important quantity, that extends KME to the space of samples, is the information potential (IP) of the data set \citep{ux}, which is simply the empirical mean value of the PDF over the space of samples in the RKHS. It appears as the argument of the logarithm, $\Psi(X)$, in Renyi's quadratic entropy \citep{r2} given by
\begin{equation}
H_2(X) = -log\int{p(x)^2 dx} = -log\Psi(X).
\end{equation}
One can estimate $\Psi(X)$ by using the Parzen density estimator \citep{parz} for estimating $p(x)$. Hence, assuming a Gaussian kernel window of kernel width $\sigma$, one can readily estimate directly from experimental data ${x_i, i=1, ..., N}$ the information potential as
\begin{equation}
\begin{aligned} 
\Psi(X) = \int p(x)^2 dx = \int\bigg(\frac{1}{N}\sum_{i=1}^{N}G_{\sigma}(x-x_i)\bigg)^2dx \\ = \frac{1}{N^2}\int
%\bigg(\sum_{i=1}^{N}\sum_{j=1}^{N}G_\sigma(x-x_j).G_\sigma(x-x_i)\bigg)dx \\ = \frac{1}{N^2}
%\sum_{i=1}^{N}\sum_{j=1}^{N}\int G_\sigma(x-x_j).G_\sigma(x-x_i)dx \\ = \frac{1}{N^2}
\sum_{i=1}^{N}\sum_{j=1}^{N} G_{\sigma/\sqrt{2}}(x_j - x_i)
\end{aligned}
\label{7}
\end{equation}
Therefore the IP is a number obtained by the double sum of the Gaussian functions centered at differences of samples with a larger kernel size. There is a physical interpretation of $\Psi(X)$ as the total information potential, if we think of the projected samples (the Gaussians) as particles in a potential field, hence the name information potential. The total potential can be written as
\begin{equation}
\Psi(X) = \frac{1}{N}\sum\limits_{j=1}^{N}\psi(x_j),
\end{equation} 
where,
\begin{equation}
\psi(x) = \frac{1}{N}\sum\limits_{i=1}^{N}G(x - x_i)
\label{cf}
\end{equation}  
represents the \textit{field} due the addition of Gaussians centered at each sample. We refer to $\psi(x)$ as the \textit{information potential field} (IPF), and it is the equivalent of the probability measure in RKHS, i.e. it gives us the probability in each point of the space induced by the data.

We see here that the main outcome of the RKHS based formulation of data PDF involves a representation over all the space populated by samples, where each sample becomes a Gaussian function. But notice that we still have not quantified the local anisotropy created by the Gaussian bumps, which changes across space and it is dependent upon the ``local density of samples". Having such a data description is beyond the concept of PDF. However, it is possible to define \textit{local operators} in the RKHS to formulate a \textit{field based} definition of the IPF that is governed by \textit{local interactions} between the Gaussian bumps. Similar to spectral theory, one can impose a local differential operator (the Laplacian) over the IPF to eventually formulate a decomposition in the Gaussian RKHS, which can be an indicator for uncertainty in that particular local region of the space. Indeed the approach of utilizing the Laplacian operator is the crux for describing many physical phenomena formed as a cumulation of smaller interacting states such as electric and gravitational potentials, wave propagation, etc. For instance, in Gauss's law formulation, the Laplacian operator is used for describing the local density associated with an electrostatic potential, which subsequently describes the overall charge distribution ($ q=-\epsilon_0\nabla^2 \upvarphi$). As another example, the Laplacian is also used for describing the density associated with a gravitational potential to subsequently describe the overall mass distribution. We similarly theorize that one can utilize the Laplacian operator here as well for describing the \textit{local sample density} associated with the information potential (IPF) around a data sample. 

\section{Schr\"odinger's Eigenvalue Problem over the IPF}
We know that in quantum physics, systems are characterized by discrete changes in their stochasticity or uncertainty described by the their various Eigenmodes (principal components), which describe our analysis goals very well. Moreover, there are well established mathematical procedures such as the Schr\"odinger's Eigenvalue problem to decompose the wave-function in moments that quantify uncertainty, which fits very well into our goals. Of course, we will have to make appropriate modifications to the theory since we are not working with a physical system, but an abstraction of ``particles'' as data samples, on a potential field defined by the Gaussian function in RKHS. To this end, we use a ``data-equivalent'' Schr\"odinger formulation over the IPF given by $H\psi(x) = E\psi(x)$ \citep{prin}. Here $H$ denotes the Hamiltonian over $\psi(x)$, and $E$ is the total energy over the sample space. All the physical constants have been lumped on the only variable in the Gaussian RKHS, the size of the kernel. Similar to a general quantum system, the Hamiltonian is constructed using two operators: the kinetic and potential energy operators. The kinetic energy operator here consists of a Laplacian function over the IPF, i.e. $-\frac{\sigma^2}{2}\nabla^2\psi(x)$, which essentially quantifies local sample density in our context as discussed in section 2.2. The potential energy operator will be denoted by $V_s$ which we refer to as the \textit{quantum information potential field} (QIPF). The overall formulation becomes an Eigenvalue problem over the data PDF and is given as:

\begin{equation}
H\psi(x) = \bigg(-\frac{\sigma^2}{2}\nabla^2 + V_s(x)\bigg)\psi(x) = E\psi(x)
\label{qsc}
\end{equation}

Here, the IPF, $\psi(x)$, is the probability measure and the equivalent of the wave-function (seen in general quantum systems) that is being decomposed by the Schr\"odinger's formulae. Note that $\psi(x)$ is the eigenfunction of $H$ and $E$ is the lowest eigenvalue of the operator, which corresponds to the ground state. Since, by convention in quantum systems, the probability measure is defined as the square of the wave-function ($p(x) = |\psi(x)|^2$), we rescale $\psi(x)$ in (\ref{qsc}) to be the square root of the IPF, i.e. $\psi(x) = \sqrt{\frac{1}{N}\sum\limits_{i=1}^{N}G_{\sigma}(x-x_i)}$. Rearranging the terms in (\ref{qsc}), we get:

\begin{equation}
V_s(x) = E + \frac{\sigma^2/2\nabla^2\psi(x)}{\psi(x)}
\label{sf}
\end{equation}

To determine the value of $V_s(x)$ uniquely, we require that $min(V_s(x)) = 0$, which makes $E = -min\frac{\sigma^2/2\nabla^2\psi(x)}{\psi(x)}$.

Hence, in summary, we formulated a Schr\"odinger's Eigenvalue problem over IPF (\ref{qsc}) to go beyond the data PDF in the form of $V_s(x)$ in (\ref{sf}). This is basically a variant of the ``field-based interpretation'' of the data PDF. Given the data set, we expect $V_s(x)$ to increase quadratically outside the data region and to exhibit local minima associated with the locations of highest sample density (clusters). This can be interpreted as clustering since the potential function attracts the data density function $\psi(x)$ to its minima, while the Laplacian drives it away, producing a complicated potential function in the space. We should remark that, in this framework, $E$ sets the scale at which the minima are observed. This derivation can be easily extended to multidimensional data. We can see that $V_s(x)$ in (\ref{sf}) is also a potential function that differs from the information potential ($\psi(x)$) formulation of the data PDF because it is now an energy based formulation associated with a quantum field description of the PDF.

\section{Extracting Uncertainty Moments from the QIPF}
Unlike the classical interpretation of systems, the quantum interpretation provides a much more detailed decomposition of the system dynamics by assuming it to consist of a large (potentially infinite) number of stochastic \textit{features}, given by the energy/uncertainty modes. Likewise, when applying this quantum field potential to data, the same interpretation holds. However, this is beyond the conventional description of measure and PDFs. Our results support the concept that the QIPF provides information about the gradient flow of PDF within local space neighborhoods, quantifying anisotropy, but it is still a global measure over the sample space. It is clear that the anisotropy will be higher in regions of the space with sparser number of samples, which corresponds to the tail regions of the PDF. For practical applications we would like to quantify anisotropy at different densities of samples. Therefore we further propose a mode decomposition of the QIPF that would further quantify the local gradient flow of the PDF with high resolution and precision (sensitivity and specificity). The Schr\"odinger's Eigenvalue formulation allows us perform this decomposition with relative ease. Moreover, since such a formulation has been well established in physics to yield uncertainty functionals at all operating points of a physical system, we expect its data-based form (\ref{qsc}) to yield similar moments in the sample space.\par

In quantum physics, the solution of a Schr\"odinger's equation describing some physical system (for instance, a quantum harmonic oscillator) yields successive Eigenstates of the wave-function (quantifying system uncertainties) that are related to each other by an orthogonal polynomial sequence called the \textit{Hermite polynomials}. Following this solution pattern, we extract Eigenstates (uncertainty moments) of the data-based Schr\"odinger's equation by projecting the wave-function (IPF), assumed to be in the ground state, into successively higher order Hermite polynomial spaces and then computing the corresponding QIPF (\ref{sf}).\par

The generating function of the physicist's version of Hermite polynomial sequence, first defined by Pierre-Simon Laplace \citep{lap} and later formalized by Charles Hermite \citep{hm}, is given as follows:
\begin{equation}
	H_n(x) = (-1)^ne^{x^2}\frac{d^n}{dx^n}e^{-x^2}
\end{equation}
which can be simplified into an explicit form as follows:
\begin{equation}
	H_n(x) = n!\sum_{m=0}^{\big[\frac{n}{2}\big]}\frac{(-1)^m}{m!(n-2m)!}\frac{x^{n-2m}}{2^m}
\end{equation}
Apart from occurring in the solutions of Eigenvalue problems in physics, the sequence is also seen in various other diverse fields such as combinatorics, numerical analysis (as Gaussian quadrature) and probability (Edgeworth series). Interestingly, Hermite functions also form an $L^2$ orthonormal basis which diagonalizes the Fourier transform operator. Following are some of the useful properties of the series:
\subsubsection*{Orthogonality}
Hermite polynomials are orthogonal with respect to the weight function $w(x) = e^{-x^2}$, i.e.
\begin{equation}
	\int_{-\infty}^{\infty} H_m(x)H_n(x)e^{-x^2}dx = \sqrt{\pi}2^nn!\delta_{nm}
\end{equation}

\subsubsection*{Recurrence Relation}
The physicist's Hermite polynomial sequence also satisfy the following recursion formula:
\begin{equation}
	H_{n+1}(x) = 2xH_n(x) - H'_n(x)
\end{equation}

Hence we formalize the solution for the moment decomposition of the QIPF equation by using the following conjecture:
 
\begin{conjecture}[Extraction of QIPF Uncertainty Modes]
	Consider the QIPF of the data samples $x$ as $V_s(x) = E + \frac{\sigma^2/2\nabla^2\psi(x)}{\psi(x)}$ with the associated ground state wave-function given by $\psi(x) = \sqrt{\frac{1}{n}\sum\limits_{i=1}^{n}k(x_i, t)}$. The approximate higher order uncertainty (or energy) modes of $\psi(x)$ can be extracted by projecting the ground state wave-function into the corresponding order Hermite polynomial given by $\psi_k(x) = H^*_{k}(\psi(x))$, where $H^*_k$ denotes the normalized $k^{th}$ order Hermite polynomial, normalized so that $H^*_k = \int\limits_{x=-\infty}^{\infty}e^{-x^2}[H_k(x)]^2dx = 1$. This leads to the evaluation of the higher order QIPF states as
	\begin{equation}
	\begin{aligned}
		V_s^k(x) = E_k + \frac{\sigma^2/2\nabla^2H^*_k(\psi(x))}{H^*_k(\psi(x))}\\\\ = E_k + \frac{\sigma^2/2\nabla^2\psi_k(x)}{\psi_k(x)}
		\label{vs}
	\end{aligned}
	\end{equation}
	where $k$ denotes the order number and $E_k$ denotes the corresponding eigenvalues of the various modes and is given by
	\begin{equation}
		E_k = -min\frac{\sigma^2/2\nabla^2\psi_k(x)}{\psi_k(x)}
		\label{ek}
	\end{equation}
\end{conjecture}
The extracted modes of the data QIPF given by $V^k_s(x)$ are thus stochastic functionals depicting the different moments of uncertainty of the data at any point $x$. This is different from the IPF formulation of (\ref{cf}) and it can also be visualized as an energy based metric resembling the potential energy operator in a quantum harmonic oscillator at various energy levels (Eigenstates) depicted by $E_k$.

\section{Summarizing the Framework}
\begin{figure*}[!b]
\centering
\includegraphics[scale = 0.5]{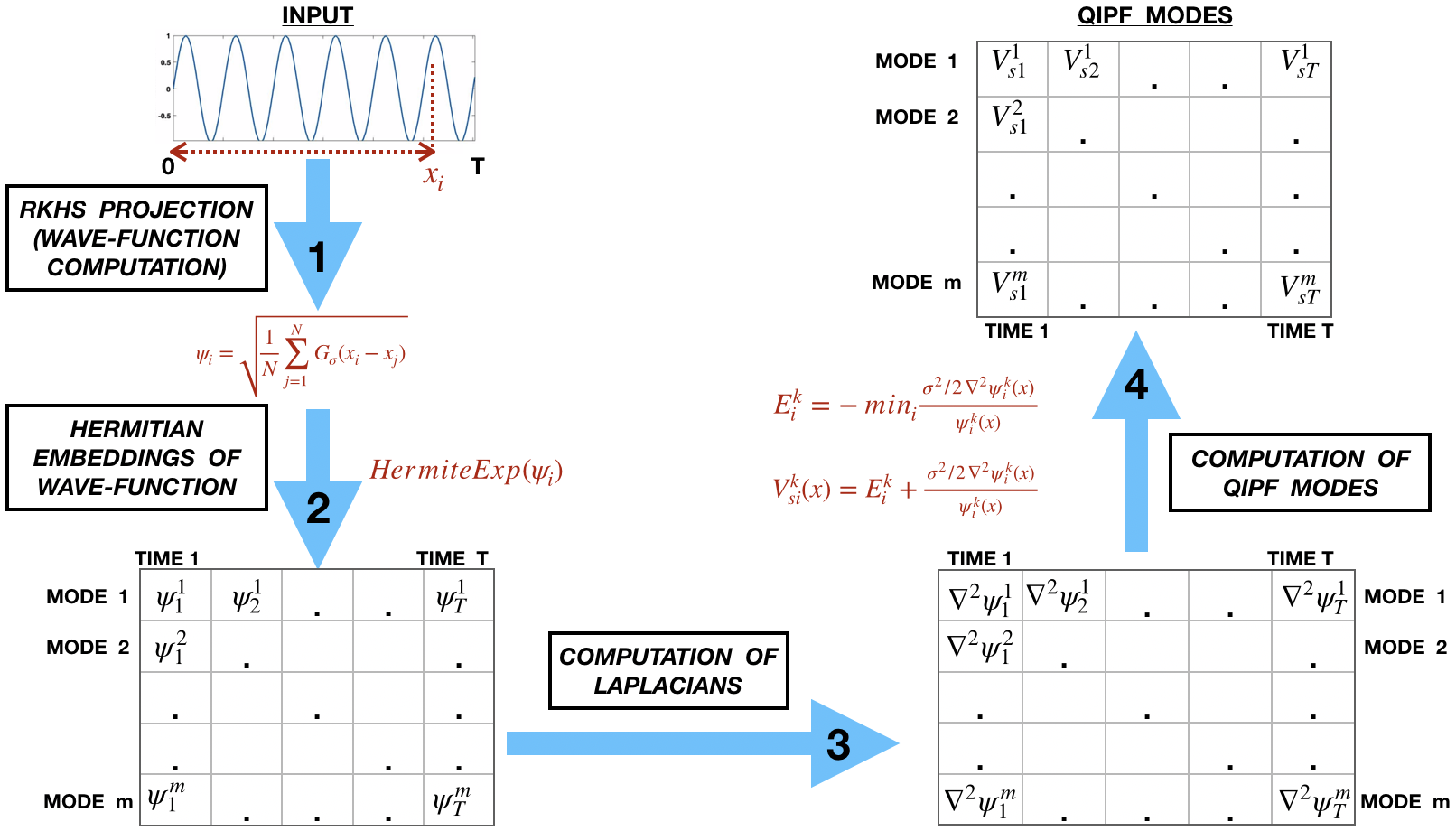}
\caption{Proposed framework for extraction of quantum uncertainty states}
\label{frf}
\end{figure*}

The implementation details of the framework are illustrated in figure \ref{frf} and the key aspects are shortly summarized as follows.
\begin{itemize}
	\item \textbf{Quantifying Data PDF in the Gaussian RKHS:} We project data using the Gaussian RKHS and provide a non-parametric characterization of the implicit PDF of the data in an incremental manner individual gaussian bumps associated with data points in the RKHS (quantified as the IPF).
	\item \textbf{Eigenvalue Problem Formulation of Data PDF:} We quantify the local gradient flow of the data PDF by utilizing the Laplacian operator in the form of the QIPF expression, which is a data-based Schr\"odinger's Eigenvalue formulation.
	\item \textbf{Extraction of Data Uncertainty Modes:} We extract uncertainty moments (or modes) of the local gradient flow of data PDF, now expressed as an Eigenvalue problem (QIPF), by implementing orthogonal Hermite polynomial projections of the ground state wave-function and finding the corresponding QIPF state.
	\end{itemize}
	
The mathematical foundations of the Gaussian RKHS, specifically the kernel trick, which leads to the kernel mean embedding theory, guarantees that our framework (through information potential computation) is able to non-parametrically and universally estimate the PDF associated with any kind of function. This is established by the universal approximation property of kernel methods. In addition to utilizing this property, The QIPF framework, through its unique uncertainty moment decomposition paradigm, is able to point to different regions of the function’s PDF and quantify how well the function is defined at such regions. Therefore, while this serves as a valuable property for quantifying epistemic uncertainty, the QIPF framework can also be utilized as an efficient signal processing tool for decomposing time-series signals. The uncertainty modes can be utilized here for clustering different samples of the signal based on how well they fit in the overall PDF. We therefore formalize the utility of the QIPF framework for time-series signals as follows:
\begin{conjecture}[Time-Series Decomposition using QIPF Framework]
	A given time series signal $x(t)$ can be characterized in terms of its intrinsic uncertainty modes extracted through the QIPF framework. This is given by:
		\begin{equation}
	\begin{aligned}
		V_s^k(x_t) = E_k + \frac{\sigma^2/2\nabla^2\psi_k(x_t)}{\psi_k(x_t)}
		\label{vst}
	\end{aligned}
	\end{equation}
	where $\psi_k(x_t)$ is the wave-function evaluation of sample at time $t$ in an RKHS field formed by using all previous samples in time, $x_0, x_1,...,x_{t-1}$, as centers. $V_s^k(x_t)$ is the $k^{th}$ QIPF mode of signal sample at time $t$ with respect to previous samples. This is equivalent to spectral decomposition of the signal on a sample-by-sample basis.
\end{conjecture}

\section{Mode Decomposition of Time Series}
\begin{figure}[!t]
  \begin{subfigure}{.45\linewidth}
    \centering\includegraphics[width=7.2cm, height=3.3cm]{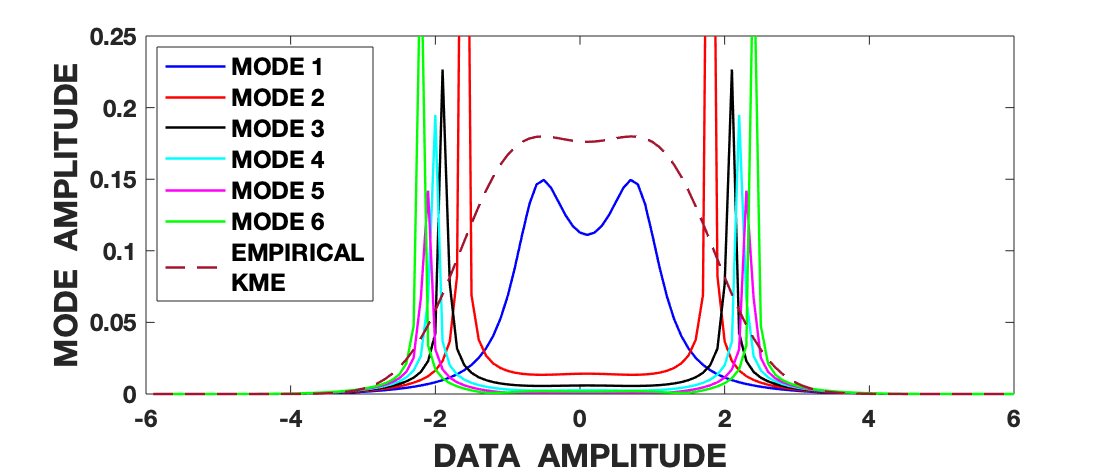}
    \caption{Kernel width = 0.6}
  \end{subfigure}
  \begin{subfigure}{.45\linewidth}
    \centering\includegraphics[width=8cm, height=3.3cm]{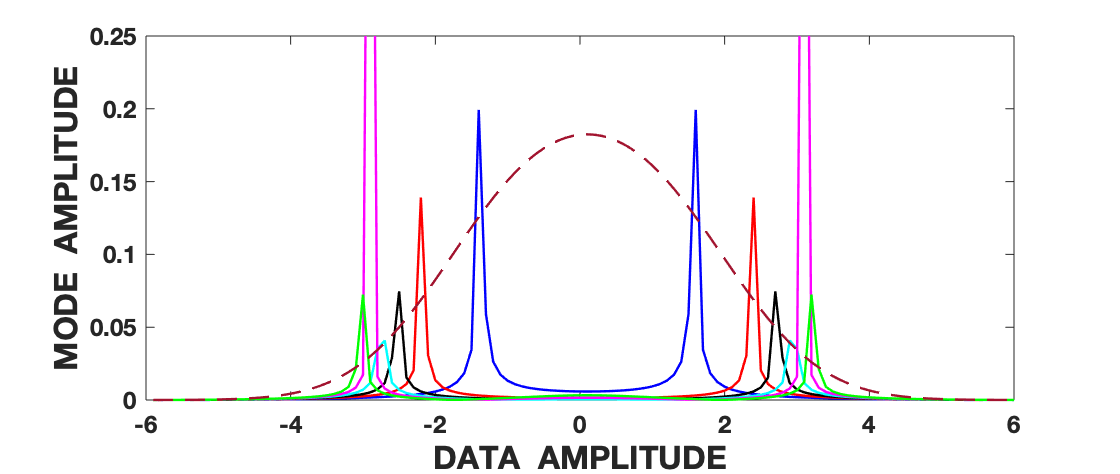}
    \caption{Kernel width = 1.2}
  \end{subfigure}
  \bigskip
  
  \begin{subfigure}{.45\linewidth}
    \centering\includegraphics[width=7.2cm, height=3.3cm]{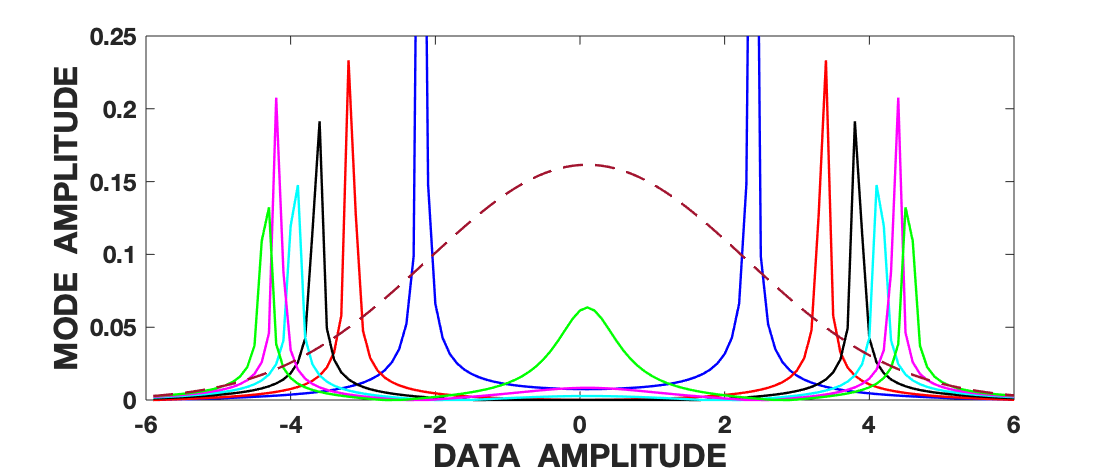}
    \caption{Kernel width = 1.8}
  \end{subfigure}
  \begin{subfigure}{.45\linewidth}
    \centering\includegraphics[scale = 0.2]{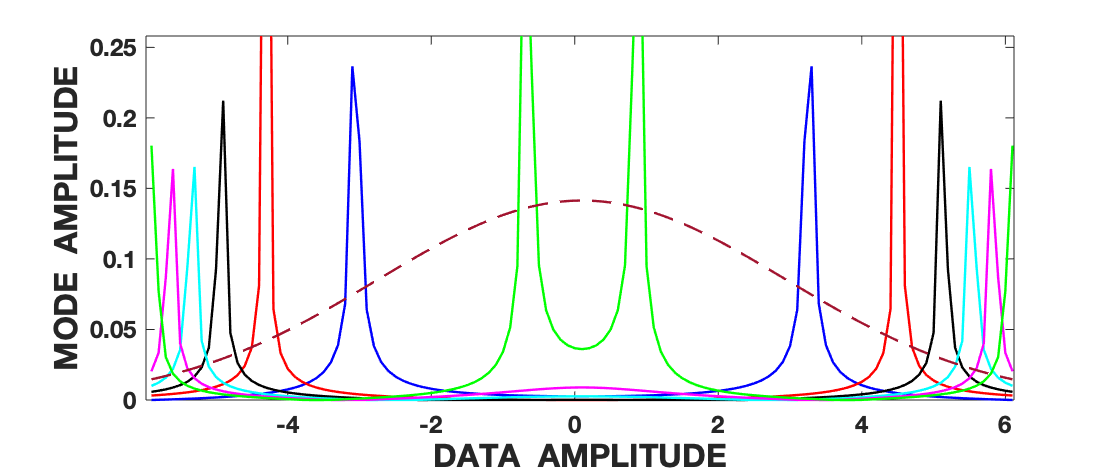}
    \caption{Kernel width = 2.6}
  \end{subfigure}
  \caption{Analysis of mode locations in the data space using different kernel widths. Solid colored lines represent the different QIPF modes. Dashed line represents the empirical KME (IPF).}
  \label{space}
\end{figure}

We begin our analysis of the proposed framework by studying how it characterizes time series signals. We used MATLAB R2019a to obtain the results shown in this section. For an intuitive understanding of how the different QIPF modes get configured in the space of data, we extracted the first 6 modes of a simple sine wave signal. We generated 3000 samples of a 50 Hz unweighted sine wave signal sampled at the rate of 6000 samples per second to mimic a continuous signal. The signal was also normalized to zero mean and unit standard deviation. We used all 3000 samples as centers to construct the wave-function (square root of the IPF) and then evaluated it at each point in the data space range $x = (-6, 6)$ using a step size of 0.1. We then evaluated the Hermite projections of the wave-function value at each point to subsequently extract 6 QIPF modes using the formulation given by (\ref{vs}). This was done for four different kernel widths whose corresponding QIPF plots (represented by solid color lines) are shown in fig. \ref{space}. The dashed line represents the empirical KME estimate (or simply the IPF) given by $p(x) = \psi^2(x) = \frac{1}{N}\sum\limits_{i=1}^{N}\kappa(x, x_i)$, which basically gives an estimate of the data PDF. All plots were normalized for easier visualization. Perhaps the most important property of the extracted QIPF modes that can be observed from the plots is that, for all kernel widths, they systematically signify the more uncertain
 regions of the data space closer to the tails of the data PDF. In fact, one can observe the significant increase in the density (or clustering) of the extracted QIPF modes as one moves farther away from the mean ($x=0$) and towards the PDF tails. Furthermore, we observe here that the modes appear sequentially based on their orders, with the lower order modes signifying regions closer to the mean and the higher order modes clustering together at the PDF tails. An interesting observation that must also be noted is that, for larger kernel widths (1.8 and 2.6), which exceed the dynamic range of the signal, we can see that some high order modes begin to emerge in the region around the mean. This behavior is remarkably similar to physical systems. If we consider the same drum membrane analogy we used in section I, one can visualize the space of the samples here as the membrane. If we increase the tension of the membrane and hit it, the drum will vibrate for a long time. In our potential field, the stiffness is controlled by the kernel size. If the kernel size is large, the QIPF becomes stiffer leading to the energy in the higher QIPF modes to increase. If one decreases the kernel size, the membrane becomes more elastic leading to many local modes that decay much faster.\par

As a pedagogical demonstration to understand how the framework characterizes different dynamical data structures, we implement it to compare the extracted uncertainty modes of a simple sine wave oscillator and a Lorenz series. The sine wave represents one of the simplest time series with a single generating function. The Lorenz series on the other hand is a chaotic deterministic dynamical system with complex state space defined by the following mutually coupled differential equations (with $\sigma$, $\rho$ and $\beta$ as system parameters) governing its dynamics:
\begin{equation}
\frac{dx}{dy} = \sigma(y-x),\> 
\frac{dy}{dt} = x(\rho - z) - y,\>
\frac{dz}{dt} = xy - \beta{z}.
\label{lorenz}
\end{equation}

\begin{figure}[!t]
  \centering
  \begin{subfigure}[t]{.3\linewidth}
    \centering\includegraphics[scale=0.12]{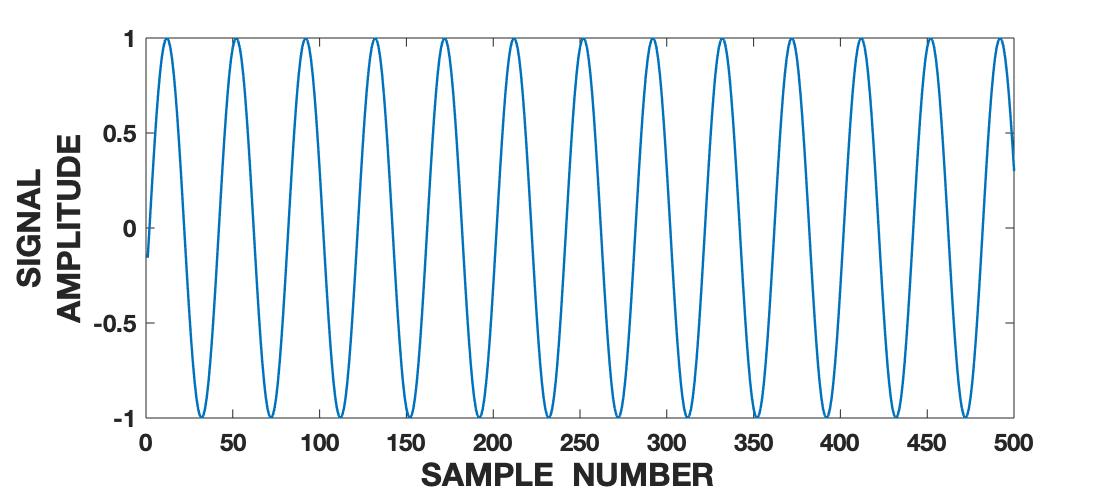}
%    \caption{This is a sub-caption}
  \end{subfigure}
  \begin{subfigure}[t]{.3\linewidth}
    \centering\includegraphics[scale=0.12]{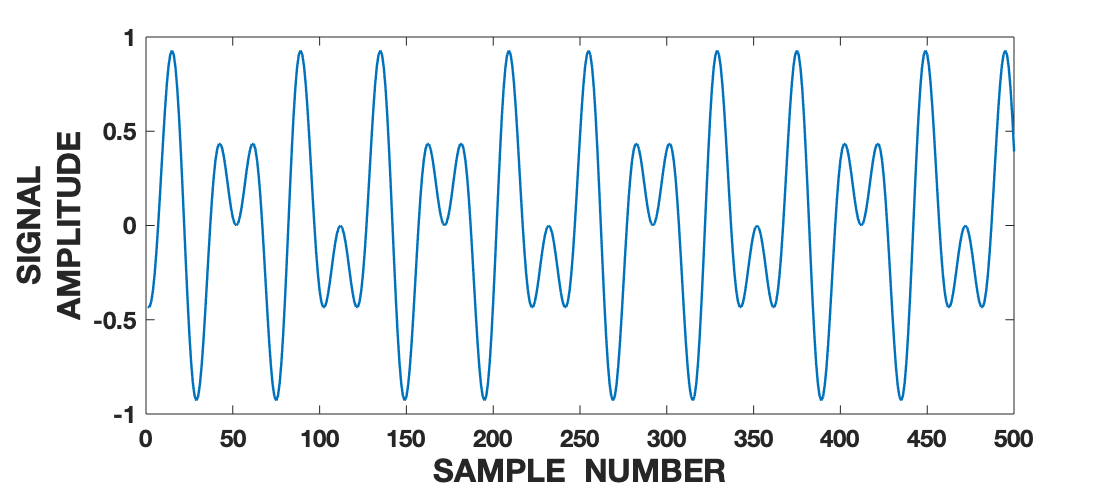}
%    \caption{This is a sub-caption.}
  \end{subfigure}
  \begin{subfigure}[t]{.3\linewidth}
    \centering\includegraphics[width=4.75cm, height = 2.1cm]{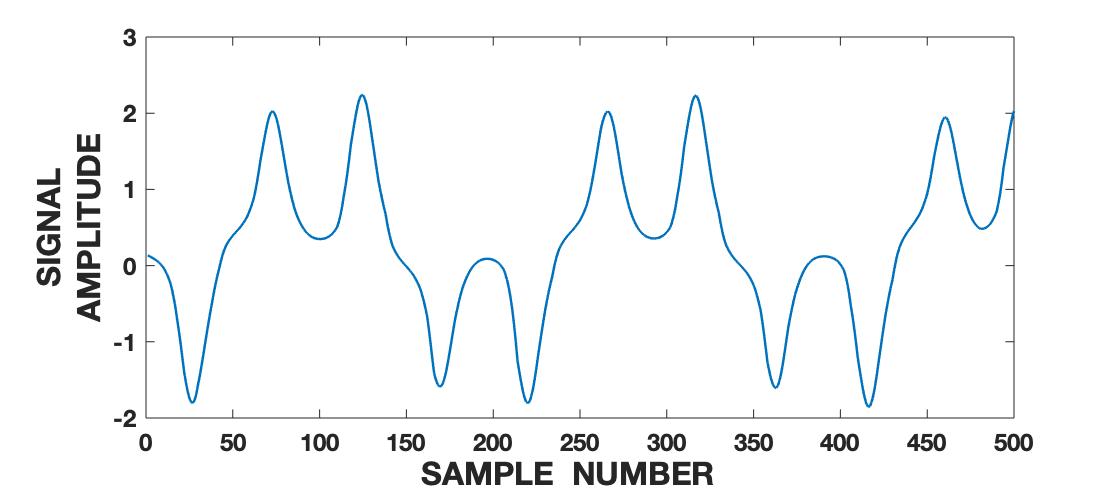}
%    \caption{This is a sub-caption}
  \end{subfigure}

\bigskip

  \begin{subfigure}{.3\linewidth}
    \centering\includegraphics[scale=0.12]{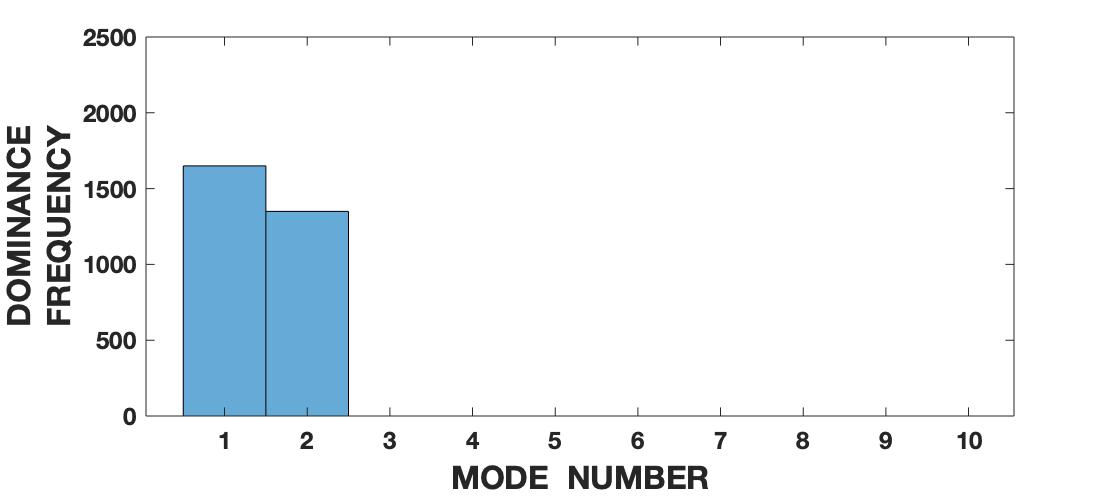}
    \caption{Sine wave (150 Hz)}
  \end{subfigure}
  \begin{subfigure}{.3\linewidth}
    \centering\includegraphics[scale=0.12]{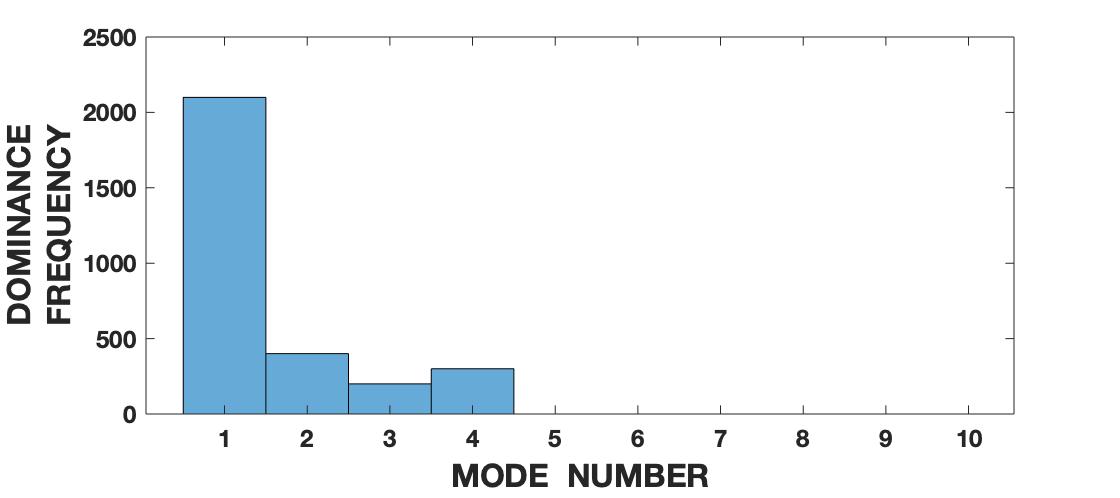}
    \caption{Sine Wave (Mixed Freq.)}
  \end{subfigure}
  \begin{subfigure}{.3\linewidth}
    \centering\includegraphics[scale=0.12]{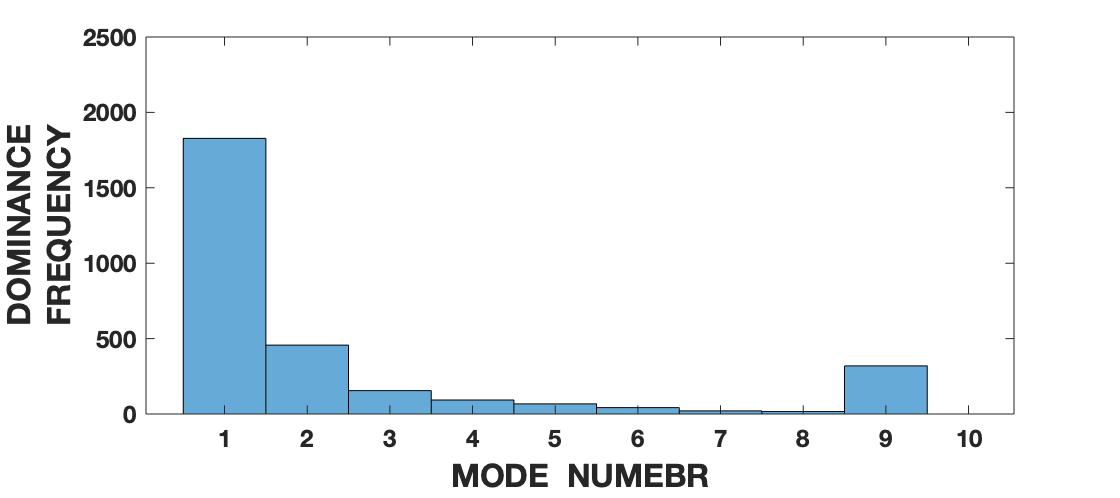}
    \caption{Lorenz Series}
  \end{subfigure}
  \caption{Characterization of different signals in terms of QIPF uncertainty moment composition. \textit{Top row:} Generated signals from different dynamical systems. \textit{Bottom row:} Dominance frequencies of QIPF modes (the number of times a particular QIPF mode has a larger value than all the other modes in the signal span) corresponding to each signal.}
  \label{dom}
\end{figure}

We generated 3000 samples of Lorenz series after setting the parameters as $\sigma = 10$, $\rho = 28$ and $\beta = 8/3$ and the initial conditions as $x_1 = 0$, $y_1 = 1$ and $z_1 = 1.05$. The signal was also normalized to zero mean and unit variance. We generated two sine wave signals with the first one having a fundamental frequency of 150 Hz and the second one with an added frequency component (150 Hz + 250 Hz). The signals were sampled at a rate of 6000 samples per second and were normalized to zero mean and unit variance. We extracted the first 10 QIPF modes using (\ref{vs}) and (\ref{ek}) to encode the different signals. The kernel width used for doing so was fixed to a moderate value of 1.2 for all signals. Fig. \ref{dom} shows the signals (top row) and the corresponding histogram plots (bottom row) of the number of times the value of each QIPF mode dominated over the others throughout the durations of the signals. As can be seen in fig. \ref{dom}(a), there are only two dominant modes in case of the single frequency sine wave (modes 2 and 3). Addition of a frequency component leads to increase in the number of modes contributing to the signal dynamics to four (fig. \ref{dom}(b)). The dominant modes in the case of Lorenz series, on the other hand, are more spread out (across all 10 modes) thus indicating a more complex data dynamical structure. These trends are quite similar to what we would expect from a frequency decomposition of the signals, except that here we are able to perform this decomposition on a sample-by-sample basis.\par

\section{Model Uncertainty Quantification}
\subsection{QIPF Framework Implementation}
The QIPF mode extraction framework can so be naturally extended for implementation on machine learning models. The fundamental idea here is to create a continuous RKHS embedding of the trained model (represented by its internal weights/activation outputs) that represents its intrinsic distribution. We can then extract the QIPF uncertainty modes associated with the interactions between model's embedding and its output. We expect this to quantify the extent to which the model's test output falls within the scope of its learnt predictive distribution. As evidenced in the previous section, we expect higher order QIPF modes to cluster in data regions where the model has not been trained, represented by the tails of the input-output mapping PDF learnt by the network, thereby providing a sensitive uncertainty characterization of data spaces unknown to the model (epistemic uncertainty). In this regard, we specifically focus on neural network models due to a recent surge in its research interest.\par
\begin{figure*}[!t]
\centering
\includegraphics[scale = 0.5]{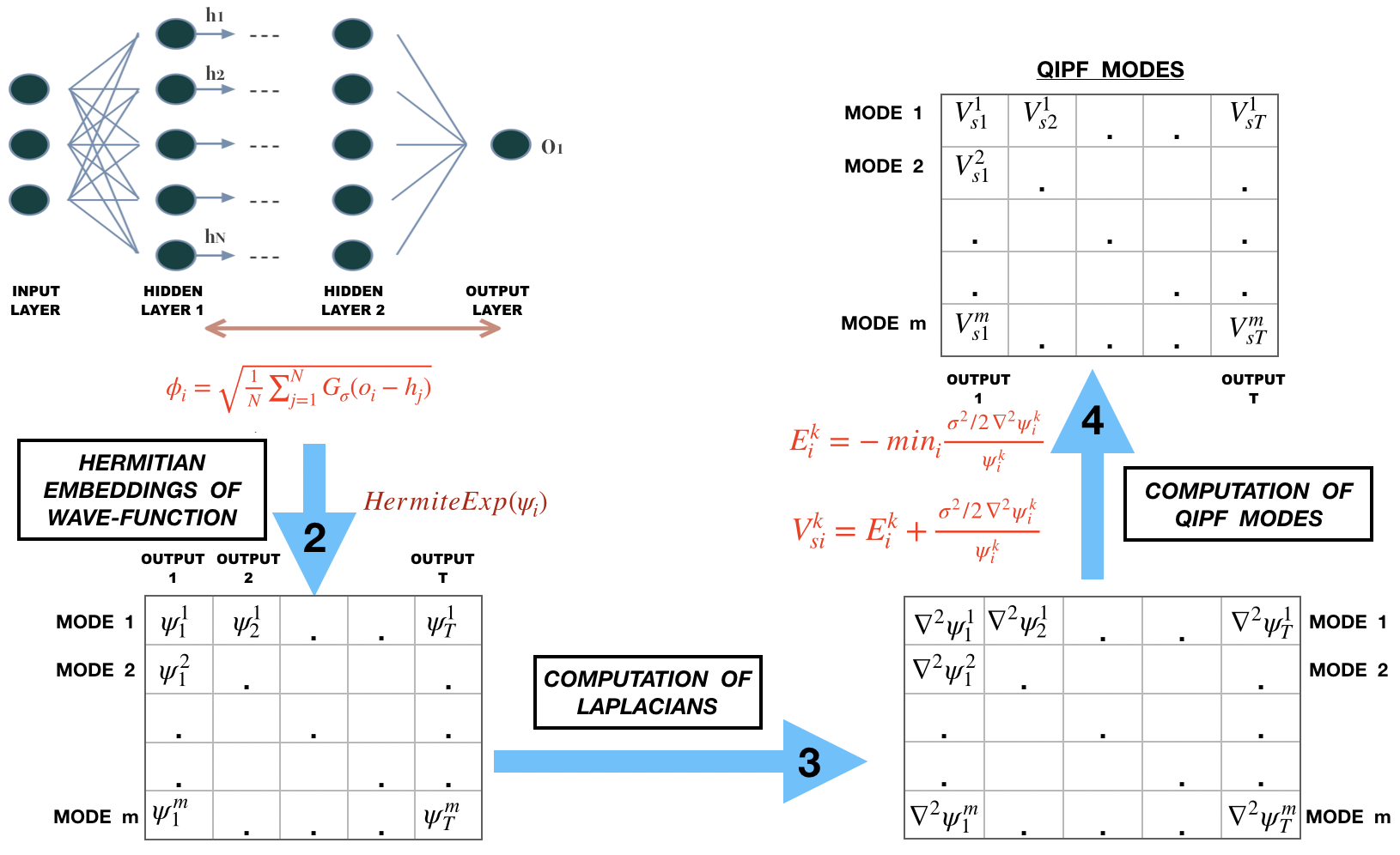}
\caption{Implementation of the proposed framework on a feed-forward artificial neural network.}
\label{frann}
\end{figure*}
The basic implementation strategy of the QIPF decomposition framework on a trained neural network is the same as the data-based implementation, except for the way in which we construct the information potential field (IPF). In this case, we aim to decompose the interactions between the RKHS fields of different pairs of network layers (with one of them typically being the output layer), thereby obtaining a multi-scale uncertainty representation of the implicit mapping between them. Intuitively, this quantifies the \textit{probabilistic similarity} between two layers of the network during each test cycle. The RKHS field of each layer is represented by the kernel feature map constructed by its corresponding node activation outputs. We evaluate the \textit{cross information potential} (CIP) which measures interactions between two information potentials \citep{prin}. One can represent the cross information potential between two layers of the neural network using a generalized form of the kernel mean embedding formulation. Let us consider two layers of an ANN whose node outputs are represented by the random variables $L_1$ and $L_2$. The kernel feature map of $L_1$ can then be represented in the same form as (\ref{kl}) as
\begin{equation}
	\mu_{\mathbb{P}_{L_1}} = \int k(l_1, .)\mathbb{P}_{L_1}(l_1)dl_1.
\end{equation}
The mean evaluation of $\mu_{L_1}$ at a point $l_2 \in L_2$ is the value of the information potential field created by $L_1$ at the point $l_2$ and can be represented as
\begin{equation}
	\mu_{\mathbb{P}_{L_1}}(l_2)=V_c({{\mathbb{P}_{L_1}},l_2}) = \int k(l_1, .)k(l_2,.)\mathbb{P}_{L_1}(l_1)dl_1,
	\label{km2}
\end{equation}
Its empirical evaluation leads to
\begin{equation}
	\hat{V_c}({{\mathbb{P}_{L_1}},l_2}) = \frac{1}{n}\sum\limits_{i=1}^{n}k\big(l_1(i)-l_2\big).
	\label{mod}
\end{equation}
which we refer to as the \textit{cross information potential field} at $l_2$.
The quantum decomposition can then be performed in the same way as before (summarized in fig. \ref{frann}), thus leading to extraction of multi-scale uncertainty features associated with the layer-layer interactions of the neural network. We will typically consider $L_2$ to be the output layer in most empirical evaluations, thus measuring the probabilistic interactions between the model's output and one or more of the hidden layers. One can either consider the model weights or its activation outputs to create the RKHS embedding of the model. Projection of weights in the RKHS provides a stable quantification of the model's PDF. The projections of activation outputs, on the other hand, quantify instantaneous realizations of the model's PDF at each test input and often provides a more computationally feasible approach. In the experiments (section 9), we show examples of both approaches. We expect the QIPF modes (resulting from the information potential interactions between the hidden layers weights/activations and the prediction output) to be more densely clustered in test data regions unknown to the model thereby characterizing its epistemic uncertainty at each test iteration.

\subsection{Comparison with Bayesian Approaches}
\subsubsection*{Bayesian Uncertainty Extraction}
The basic approach of Bayesian based predictive uncertainty quantification methods involves probabilistic descriptions of model parameters given a fixed set of data by using variational inference. For instance, consider a set of model variables $\omega = [\hat{M_i}]_{i=1}^L$ for a model with $L$ layers. One can formulate then formulate the predictive distribution for a new input point $x^*$ as:
\begin{equation}
	p(y^*|x^*, X, Y) = \int{p(y^*|x^*, \omega)p(\omega|X,Y)d\omega}.
\end{equation}

Here $y^* \in \mathbb{R}^D$. To approximate $p(\omega|X,Y)$, variational inference methods work by defining a distribution $q(\omega)$ and minimizing the Kullback-Leibler (KL) divergence between the two distributions, i.e. $KL(q(\omega)|p(\omega|X,Y))$, which results in the following formulation of approximate predictive distribution:
\begin{equation}
	q(y^*|x^*) = \int{p(y^*|x^*, \omega)q(\omega)d\omega}.
	\label{bay}
\end{equation}

In recent years, Monte Carlo Dropout \citep{gal} has become a very popular method to approximate the model's predictive distribution. Here, authors showed that regularizing any network using dropout and $L2$ regularization is equivalent to performing variational inference to determine $q(y^*|x^*)$ and Monte Carlo sampling of the network during \textit{testing} leads to an approximation of its first two moments summarized by the following expectations.

\begin{equation}
	E_q(y^*) \approx \frac{1}{T}\sum_{t=1}^{T}\hat{y}(x^*, \hat{M_1}^t, \hat{M_2}^t...\hat{M_L}^t)
	\label{mcd1}
\end{equation}

\begin{equation}
	E_q((y^*)^T(y^*)) \approx \tau^{-1}I_D + \frac{1}{T}\sum_{t=1}^{T}\hat{y}(x^*, \hat{M_1}^t, \hat{M_2}^t...\hat{M_L}^t)^T\hat{y}(x^*, \hat{M_1}^t, \hat{M_2}^t...\hat{M_L}^t)
	\label{mcd2}
\end{equation}

with the total number of stochastic forward passes (Monte Carlo samples) being equal to $T$, $\tau$ defining the model precision and $\hat{y}$ being the current output of the $t^{th}$ forward pass parameterized by the current input and the model variable realizations in different layers at that particular pass. Hence quantifying predictive uncertainty using MC dropout becomes equivalent to performing multiple forward passes through the network and averaging the results.\par

\subsubsection*{Scalability, Precision and Computational Cost}
Hence Bayesian based inferencing methods, which involve posterior estimation of the model's predictive distribution, require a large number of model realizations to quantify the predictive uncertainty with respect to its output. As is seen from (\ref{bay}), such methods intrinsically require marginalization over all model variables to give the most ideal estimates of uncertainty. Considering a neural network with $L$ layers and $n$ neurons in each layer, this would translate to a complexity of $\mathbf{O}(n^L)$, assuming we are marginalizing over each neuron (considered as a random variable) to obtain the complete predictive distribution of the model. This is intractable for modern applications involving large datasets and networks. Monte Carlo dropout considerably reduces this cost by only marginalizing over only a few neurons selected through random sampling during testing and hence reduces the cost to approximately $\mathbf{O}(d*n^L)$ $ \forall{0<d<1}$, where $d$ is the dropout rate. Although the extracted predictive distribution $q(\omega)$ through variational inference contains multi-modal information as pointed out in \citep{gal}, in practice, one is only limited to extracting the first two \textit{central moments} given by (\ref{mcd1}) and (\ref{mcd2}) to maintain a reasonably low requirement for the number of forward passes ($T$). In big data applications, however, where the number of model layers and neurons increase considerably, this too becomes intractable. The proposed QIPF framework differs from the variational inference approaches in a major way by relying on \textit{prior} estimation of the model's predictive distribution in the RKHS, thereby offering a \textit{single-shot} estimation of the model's predictive distribution with respect to its output. This is enabled by the kernel trick and the mathematical guarantees provided by the kernel mean embedding theory and therefore makes the QIPF framework a more feasible solution for real-time quantification of model uncertainty while testing.  The computational bottleneck in the QIPF framework comes from the computation of the information potential field, i.e. $\Psi(X) = \frac{1}{N}\sum\limits_{j=1}^{N}\psi(x_j)$, for data-based implementation or equivalently, from the computation of the cross information potential (CIP) given by (\ref{mod}) for model implementation. Considering the same network ($L$ layers and $n$ neurons per layer), the computational complexity of the CIP becomes $\mathbf{O}(mnL)$ where $m$ is the number of output layer neurons. Here we consider the CIP evaluation at $m$ outputs from the last layer in an RKHS field created by $nL$ activation outputs as centers. The framework therefore is significantly cheaper in terms of computations when compared with variational inference methods, but its cost is still dependent on the size of the model network (in terms of $n$ and $L$). Several ways of curbing the computational growth of the information potential with respect to the increasing data size are summarized in \citep{prin}. One notable method of achieving this is by using the \textit{fast Gauss transform (FGT)} \citep{fgt} which leverages the shifting property of the Gaussian kernel to make the computation of weighted sums of Gaussians more efficient. It reduces the computation of CIP from $\textbf{O}(mnL)$  to  $\textbf{O}(m) + \textbf{O}(nL)$. Another effective approach is by implementing \textit{incomplete Cholesky decomposition} \citep{icd} on the CIP gram matrix (augmented to make it symmetric). This reduces the computational complexity of the CIP from $\mathbf{O}(mnL)$ to  $\mathbf{O}(mD^2)$ where $D$ represents the reduced form of the gram matrix associated with $nL$ centers. Other methods such as quantization \citep{qua} involve efficient reduction of the number of RKHS centers ($nL$ in this example) for computing any kernel based metric such as the CIP in a more computational friendly manner.\par

A major property that makes the QIPF framework significantly advantageous over Bayesian approaches is the way in which the uncertainty moments are computed. Instead of computing central moments of predictive uncertainty as is done in variational inference methods, we compute the \textit{local} moments associated with the gradient flow of the data PDF at every point in the sample space. The formulation of the eigenfunction problem through the Schr\"odinger's equation allows us to potentially extract infinite moments without any sampling restrictions (as is seen in Bayesian approaches) because we simply use successive Hermite polynomial projections in the RKHS for moment decomposition. This enables a much higher resolution of quantified uncertainty than is possible through variational inference. Moreover, the Schr\"odinger's equation naturally tailors each moment to only quantify the successively more uncertain data regions with respect to its PDF.

\section{Experiments and Analysis}
We present simulation results to illustrate and compare performance of the QIPF framework with respect to currently popular approaches for the problem of predictive uncertainty quantification. All simulations described in this section were performed using python 3.6. Being a kernel based approach, we compare the QIPF framework's performance with that of Gaussian process regression (GPR) \citep{gp} which is a widely famed kernel method for machine learning that is known for providing reliable uncertainty estimates associated with its predictions. We also provide comparisons with Monte Carlo dropout \citep{gal} that has gained recent popularity as a an approximate variational inference based method for uncertainty quantification of neural networks.\par
\subsection{Regression}
\subsubsection*{Datasets}
We generate two different regression datasets as didactic examples for experimental comparisons and analysis. The idea of such datasets is to simulate real world scenarios where tasks in machine learning encounter test data from outside the training domain or have to face external noise or outliers in their training set. Indeed these synthesized datasets have also been used in various uncertainty quantification literature for demonstration of methods \citep{osband, gal}. The first dataset consists of 60 regression pairs $x_i$, $y_i$ from the following weighted sine signal:
\begin{equation}
y_i = {x_i}sin(x_i).
\end{equation}

\begin{figure}[!t]
  \centering
  \begin{subfigure}[t]{.32\linewidth}
    \centering\includegraphics[scale=0.29]{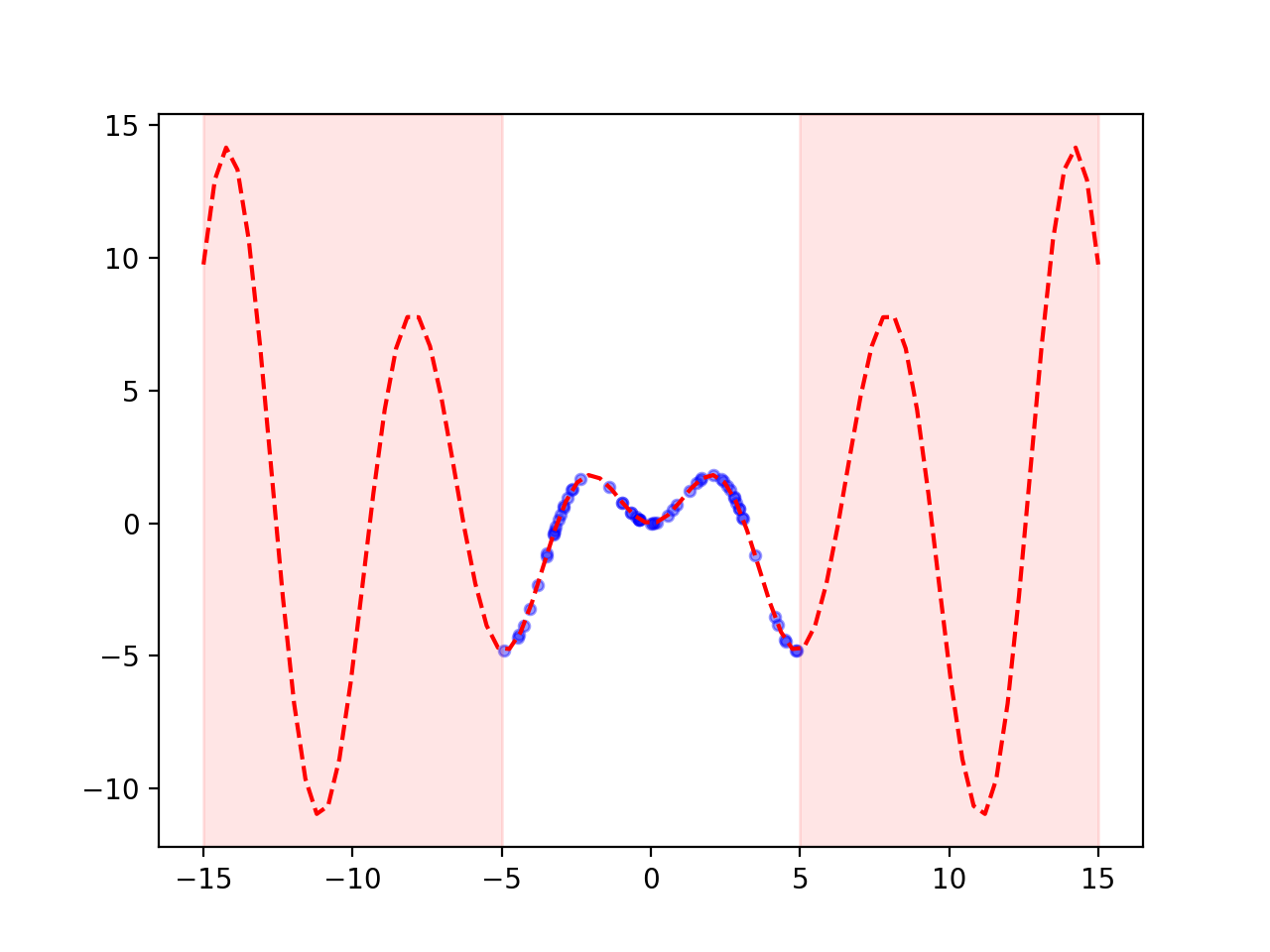}
    \caption{Synthesized data I}
  \end{subfigure}
  \begin{subfigure}[t]{.32\linewidth}
    \centering\includegraphics[scale=0.29]{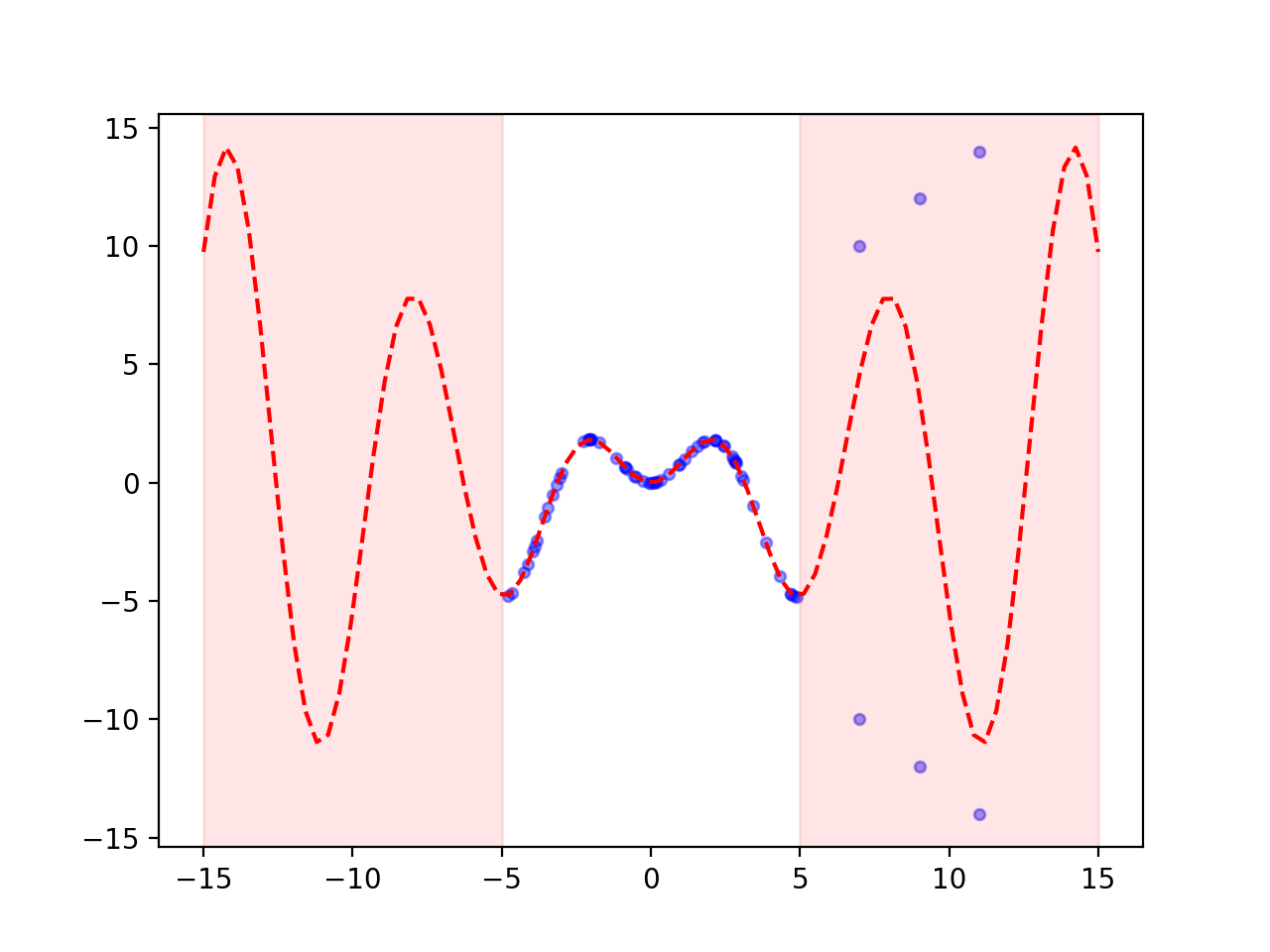}
    \caption{Synthesized data I (with outliers added)}
  \end{subfigure}
   \begin{subfigure}[t]{0.32\linewidth}
    \centering\includegraphics[scale=0.29]{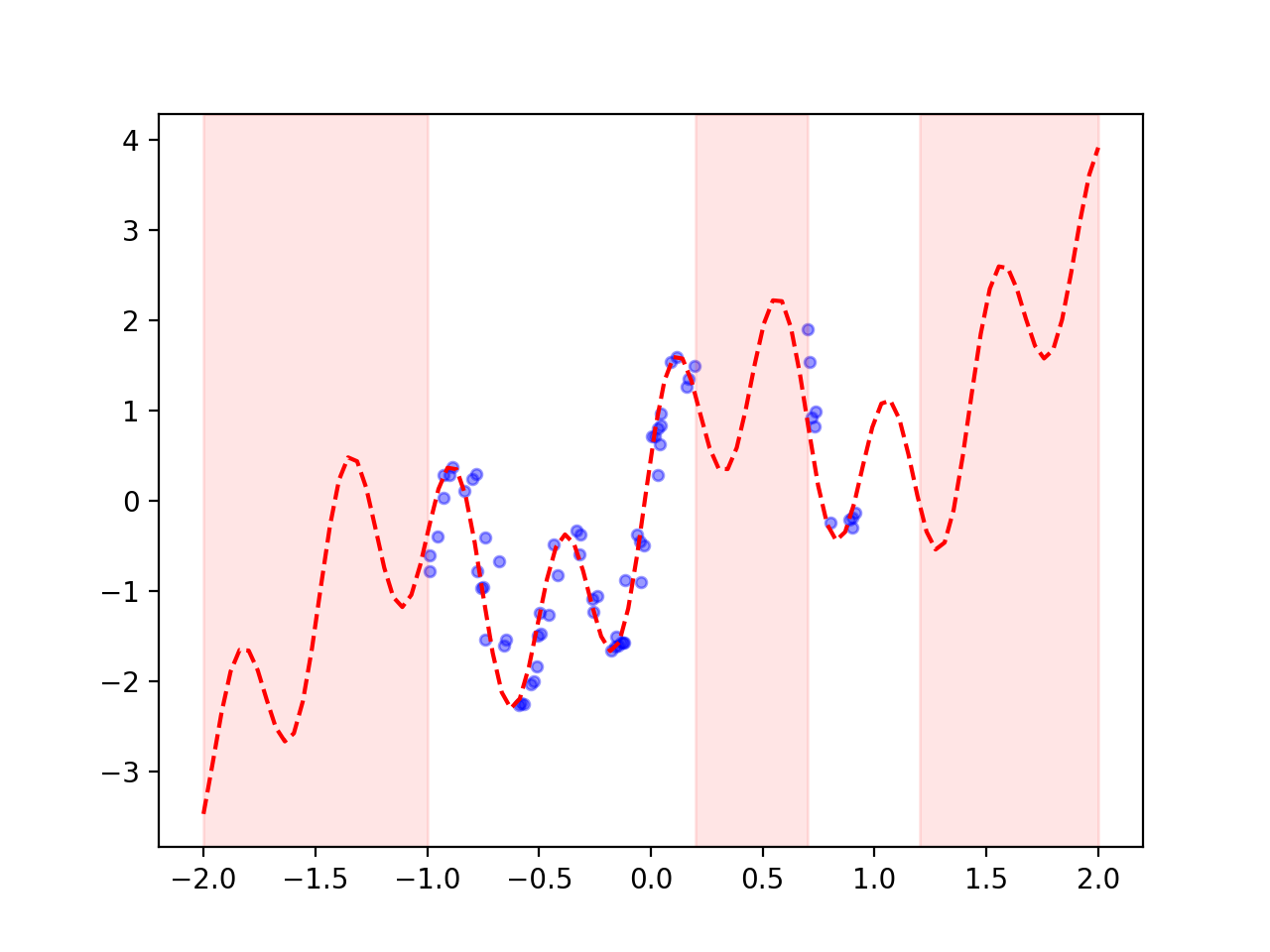}
    \caption{Synthesized data II}
  \end{subfigure}
  \caption{Synthesized datasets for experimental evaluations. Blue circles depict the sampled training data and red dashed lines represent their associated generating functions. Pink bands represent regions with no training samples.}
  \label{dats}
  \end{figure}
Here, the training inputs $x_i$ are drawn uniformly from $(-5, 5)$. The dataset is shown in fig. \ref{dats}a as synthesized dataset I. The blue circles represent the training samples and the red dotted line represents its underlying governing function in the region $(-15, 15)$. Although the training pairs are sampled only from a specific region, testing (for all algorithms) is performed in the entire data region by sampling 120 test data pairs uniformly from the region $(-15, 15)$. The pink bands represent test data regions for which training data has not been provided. We therefore expect high predictive uncertainties in these regions. As part of the analysis, we also add 6 widely varying outlier samples (not lying on the governing function) to the training set of synthesized dataset I as shown in fig. \ref{dats}b. Synthesized data II consists of noisy regression pairs $x_i$, $y_i$ sampled from the following signal: 
\begin{equation}
y_i = x_i + sin(\alpha(x_i + w_i)) + sin(\beta(x_i + w_i)) + w_i.
\end{equation}
Here, we set $\alpha = 4$, $\beta = 13$ and $w_i\sim N(\mu=0, \sigma^2=0.03^2)$. We draw 40 input samples for training uniformly from $(-1, 0.2)$ and 10 from $(0.7, 1)$, leaving the region $(0.2, 0.7)$ as blank. For model testing, we draw 120 test sample pairs uniformly from $(-2, 2)$. The dataset is depicted in fig \ref{dats}c. In addition to these datasets, we also perform model extrapolation experiments on the Mauna Loa CO2 dataset which consists of atmospheric CO2 concentrations measured from in situ air samples collected at the Mauna Loa Observatory, Hawaii \citep{keeling}.

\subsubsection*{Model Implementations}
For implementing the MC dropout and QIPF framework on synthesized data I, we use a small fully connected and ReLu activated neural network with 3 hidden layers containing 20 neurons each. We train the network on the given training samples in the region $(-5, 5)$. Since the network is very small, the dropout rate for training was set to 0.05 (similar to that recommended in \citep{gal} for a similar network size) and we used 100 epochs with the batch size equal to the number of training samples. Thereafter we tested the network on 120 input points sampled uniformly from the entire data region $(-15, 15)$. For implementing MC dropout, the test dropout rate was set to 0.2 and we used 100 forward stochastic runs to quantify the uncertainty interval at each test point. We implemented the QIPF framework by extracting 5 cross-QIPF states of the prediction point with respect to the activation outputs of each hidden layer. We used a kernel width based on the values of hidden layer activation outputs and fixed it to 20 times the Silverman's thumb rule for bandwidth estimation \citep{sil}. The criteria for kernel width depends on the range of data space on which the QIPF framework is to be implemented as well as the desired resolution of modes. Since we are operating on a relatively small neural network over a limited data span, we speculated that 5 QIPF modes would be enough for our implementation. Thereafter we quantified the uncertainty interval by measuring the standard deviation of the extracted cross-QIPF states at each point of prediction. For synthesized data II, we used a 2-layer network with 50 neurons each for training. The training dropout rate was set to 0.1 and we used 100 training epochs with the batch size equal to the total training data. We implemented MC dropout and QIPF frameworks in the same manner as before on the entire data region $(-2, 2)$ using 120 uniformly generated test samples. In the case of the CO2 data, we used a relatively larger fully connected ReLu network of 5 layers with 100 neurons each. Training dropout rate in this case was set as 0.2. We test the network over the entire training region as well as an extrapolated region outside of the it. We also fitted the Gaussian Process regression model on all datasets. In each case, the parameters associated with the covariance kernel function were chosen using grid search so as to maximize the log marginal likelihood of the data.\par
\begin{figure}[!t]
  \centering

  \begin{subfigure}{\linewidth}
    \centering\includegraphics[scale=0.13]{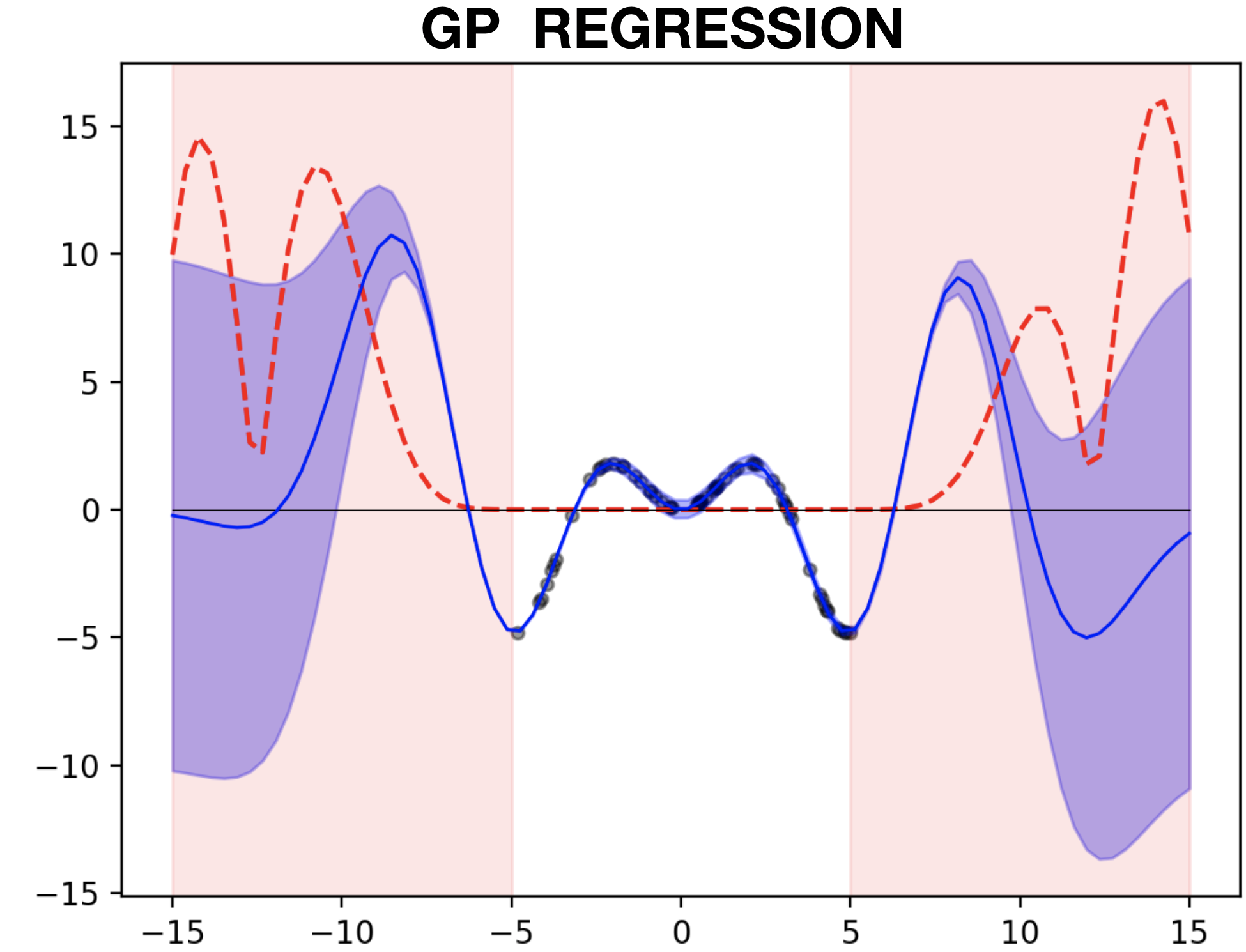}
    \centering\includegraphics[scale=0.13]{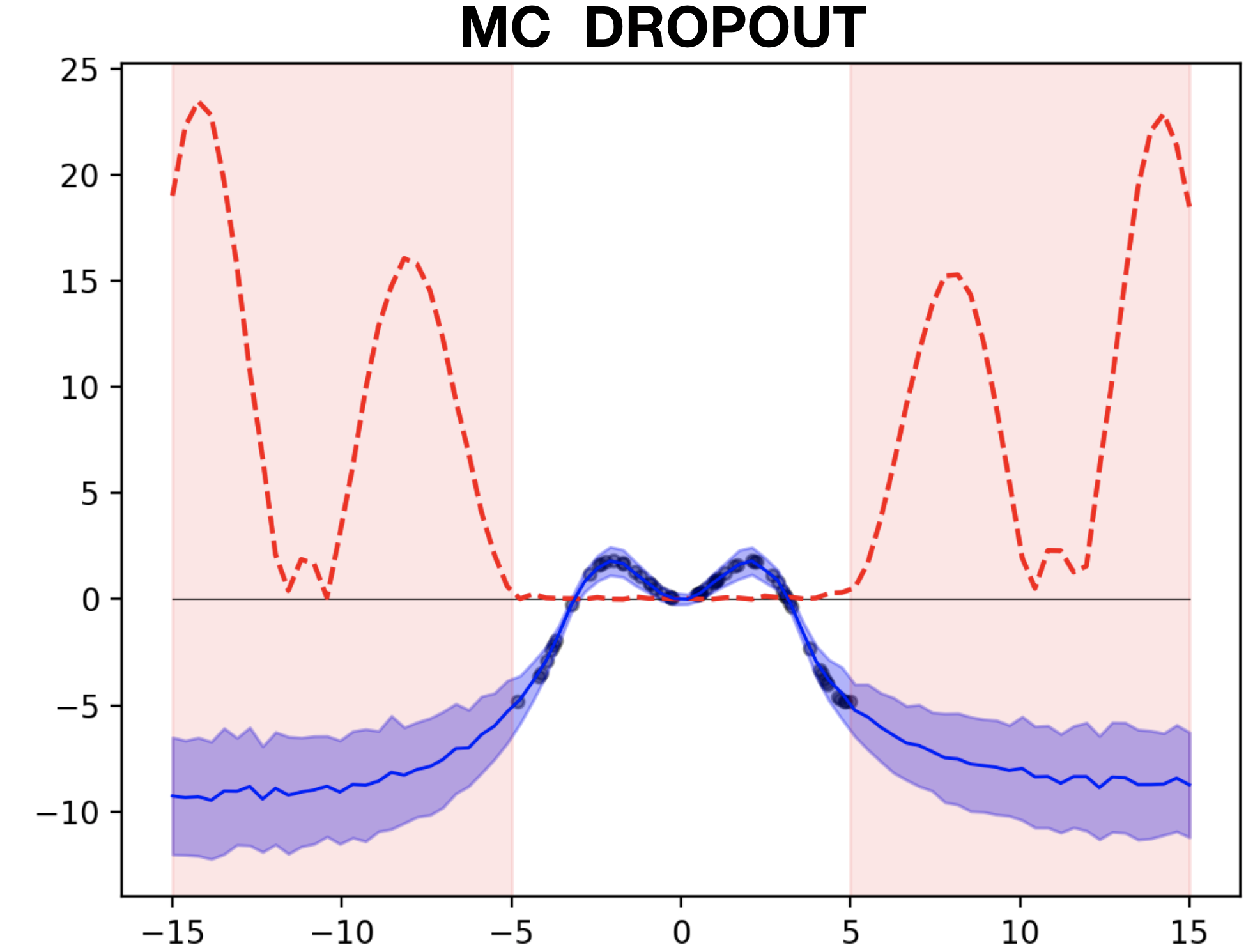}
    \centering\includegraphics[scale=0.13]{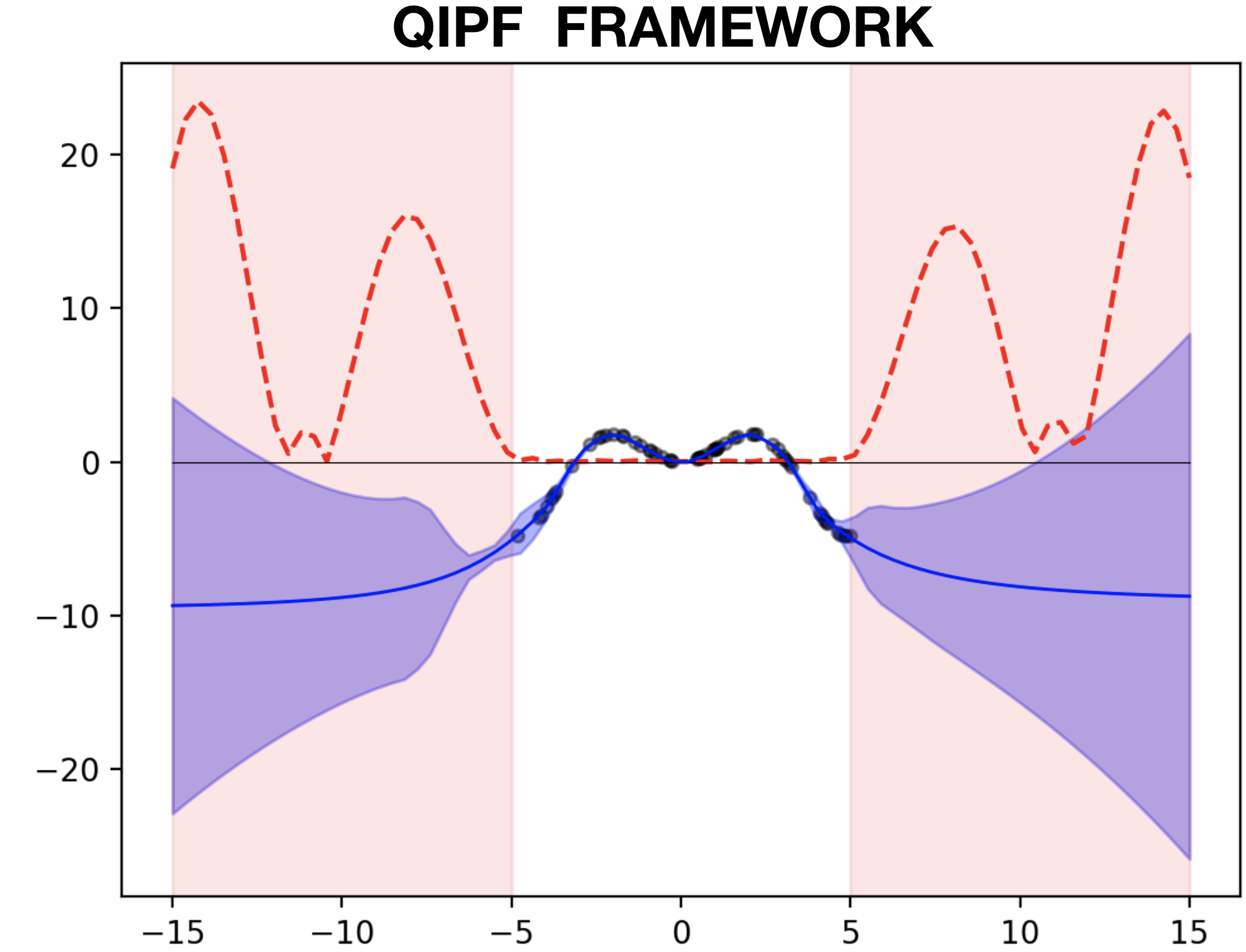}
    \caption{Synthesized data - I}
  \end{subfigure}
  
  \bigskip
  
  \begin{subfigure}{\linewidth}
    \centering\includegraphics[scale=0.129]{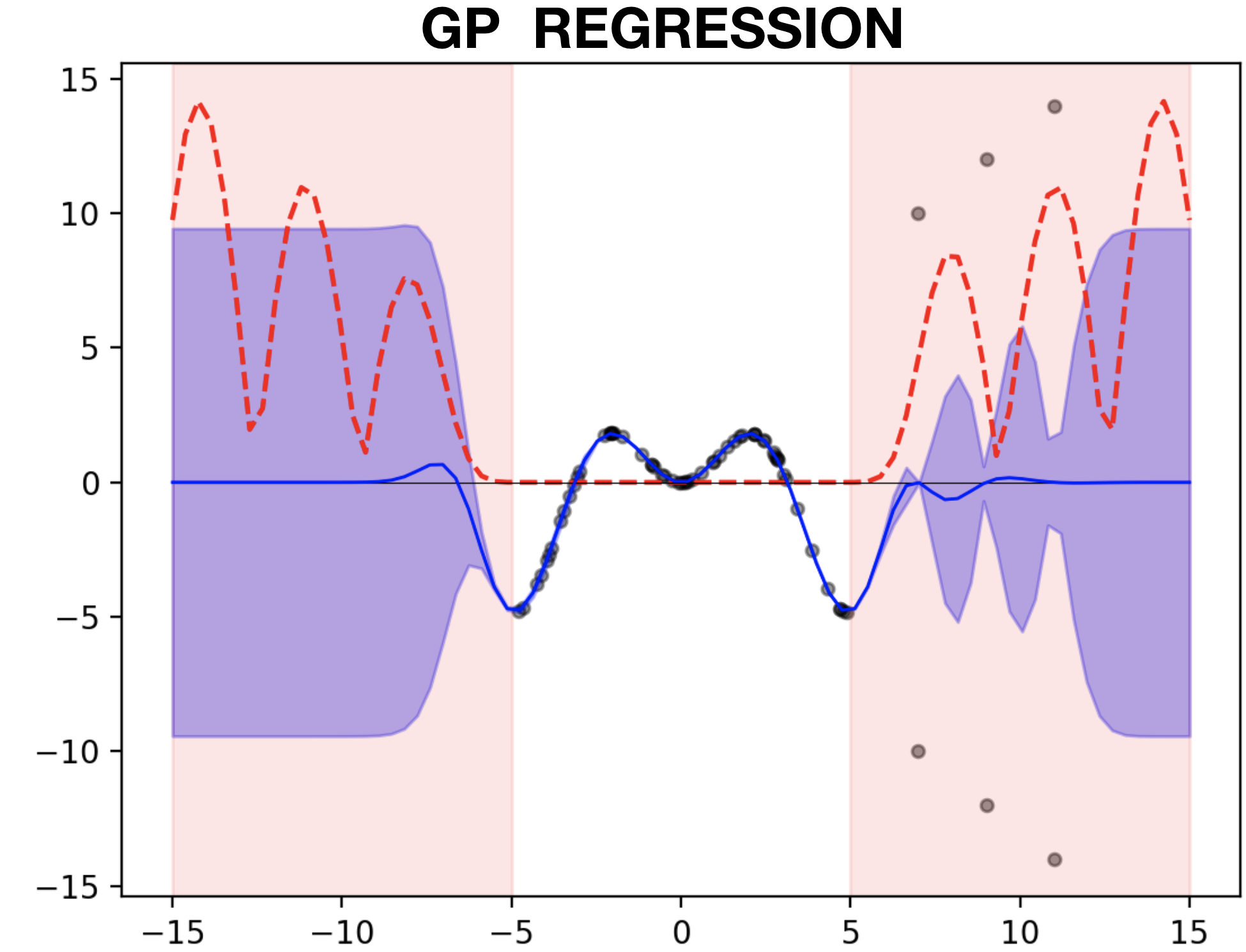}
    \centering\includegraphics[scale=0.107]{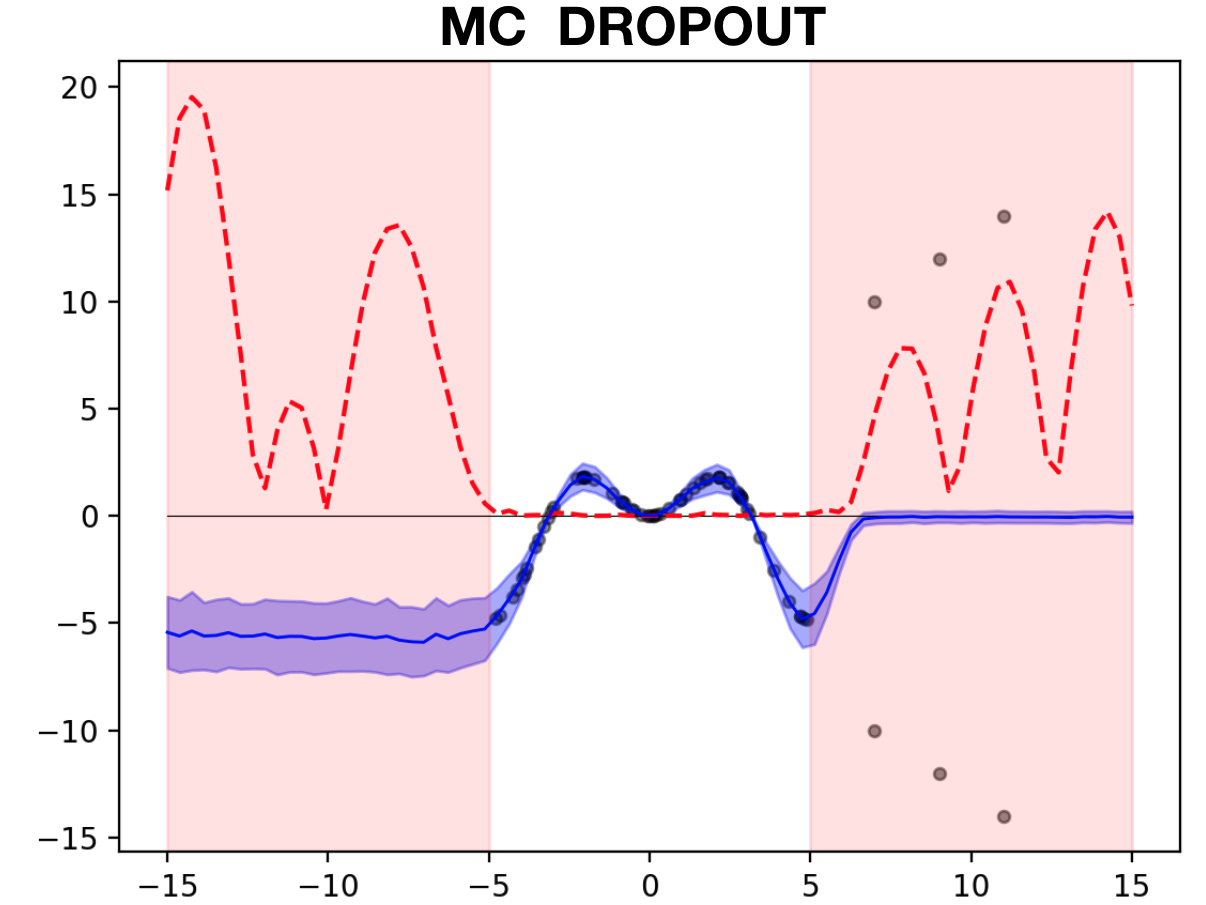}
    \centering\includegraphics[scale=0.129]{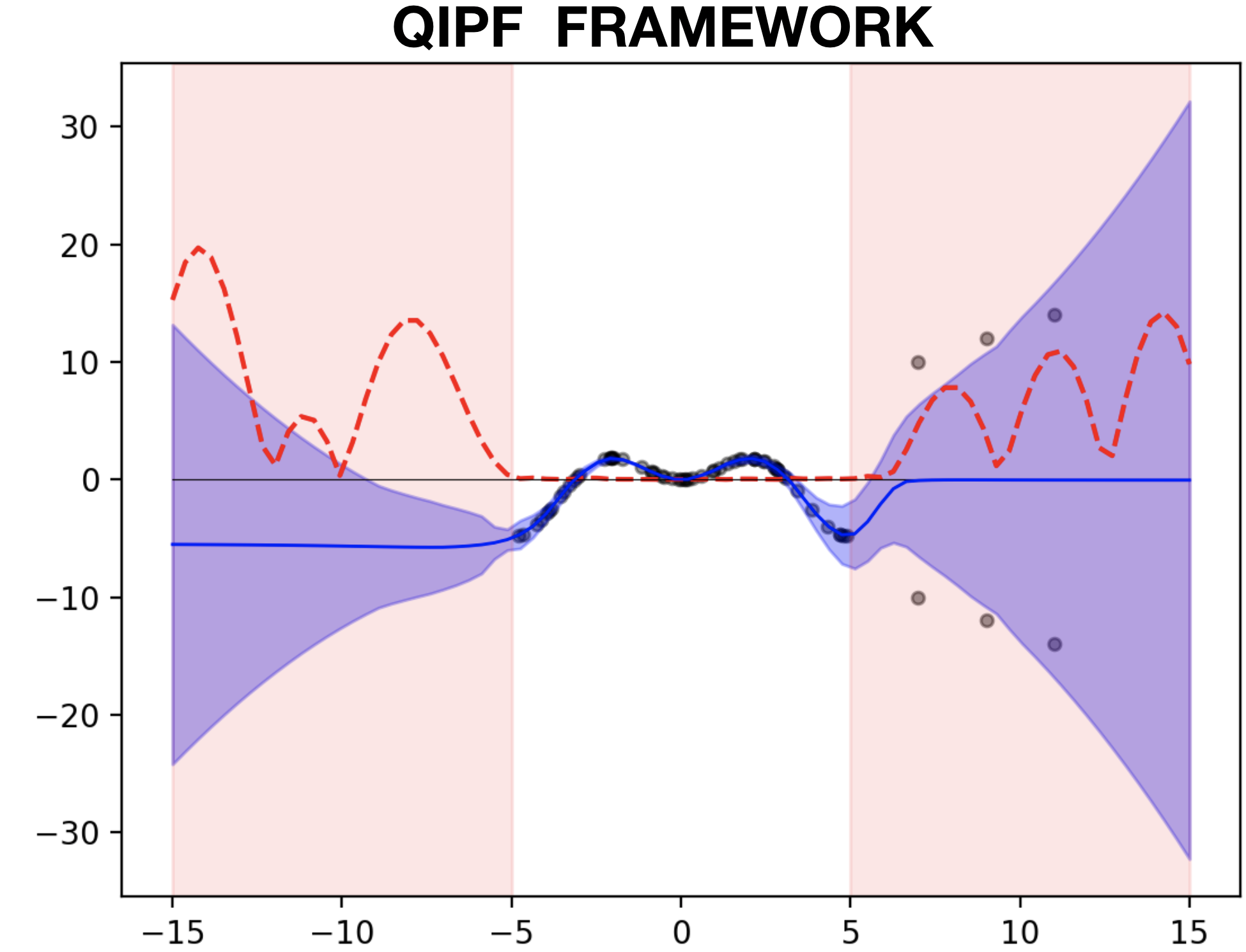}
 \caption{Synthesized data - I (with outliers added)}
  \end{subfigure}
  
  \bigskip
  
    \begin{subfigure}{\linewidth}
    \centering\includegraphics[scale=0.13]{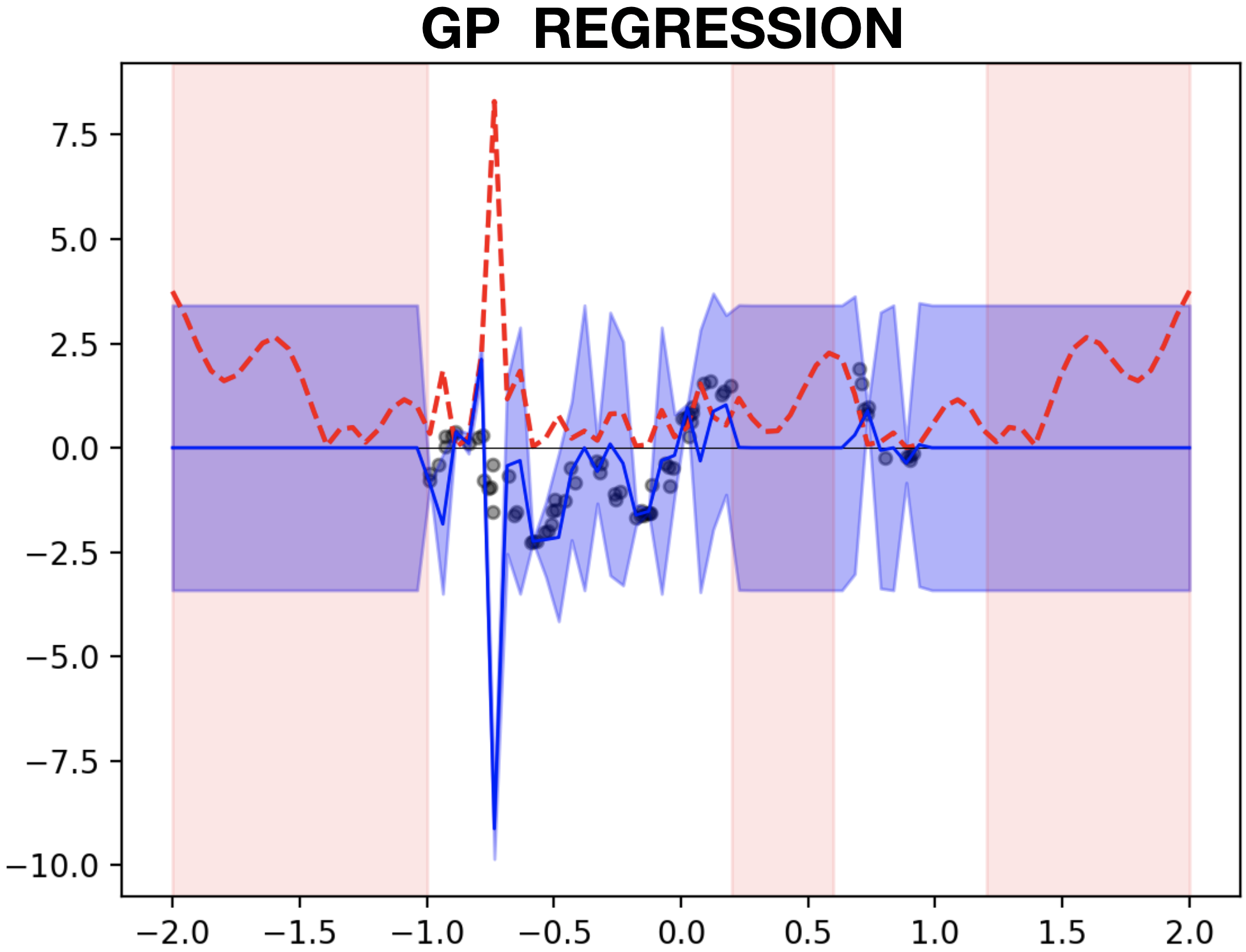}
    \centering\includegraphics[scale=0.13]{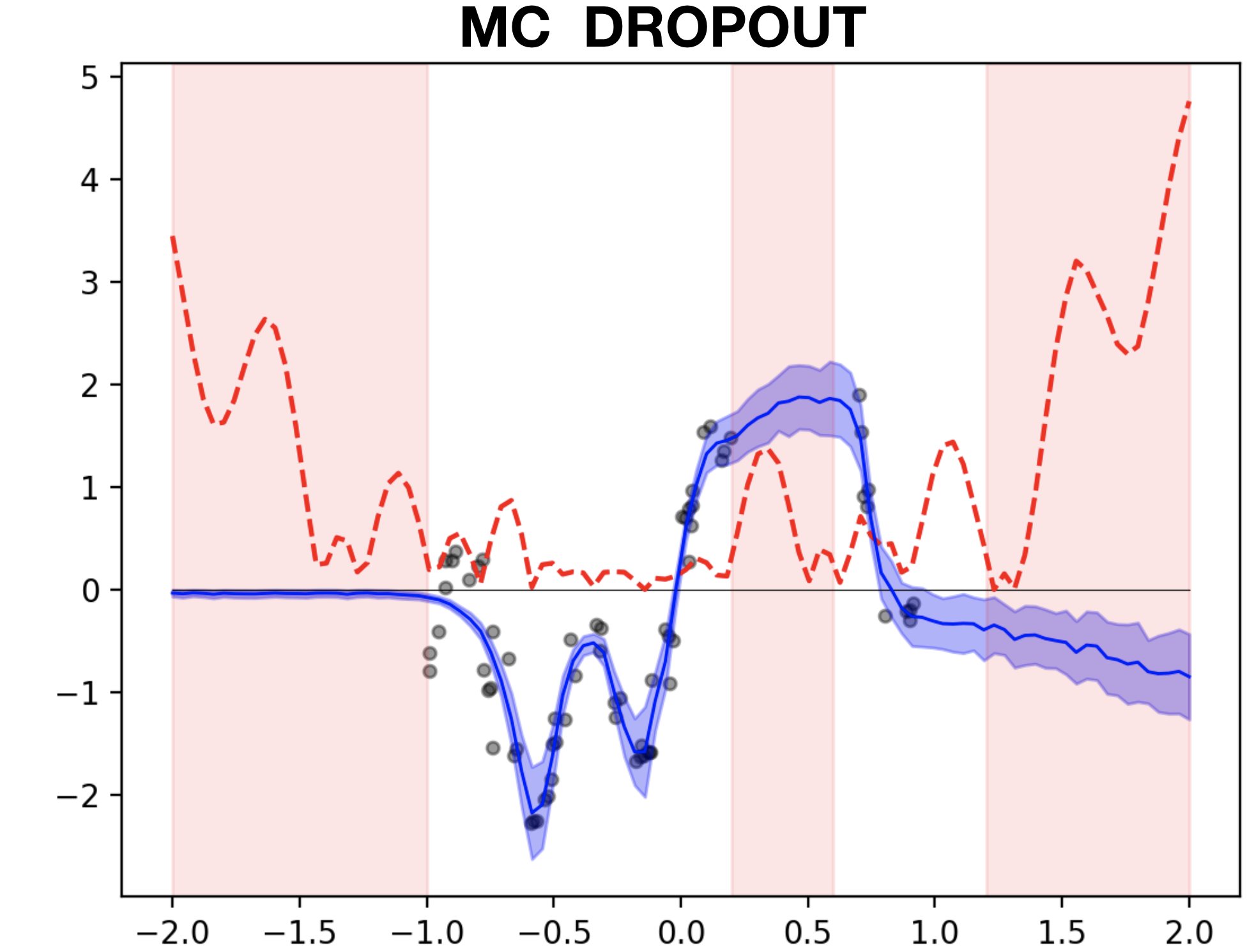}
    \centering\includegraphics[scale=0.13]{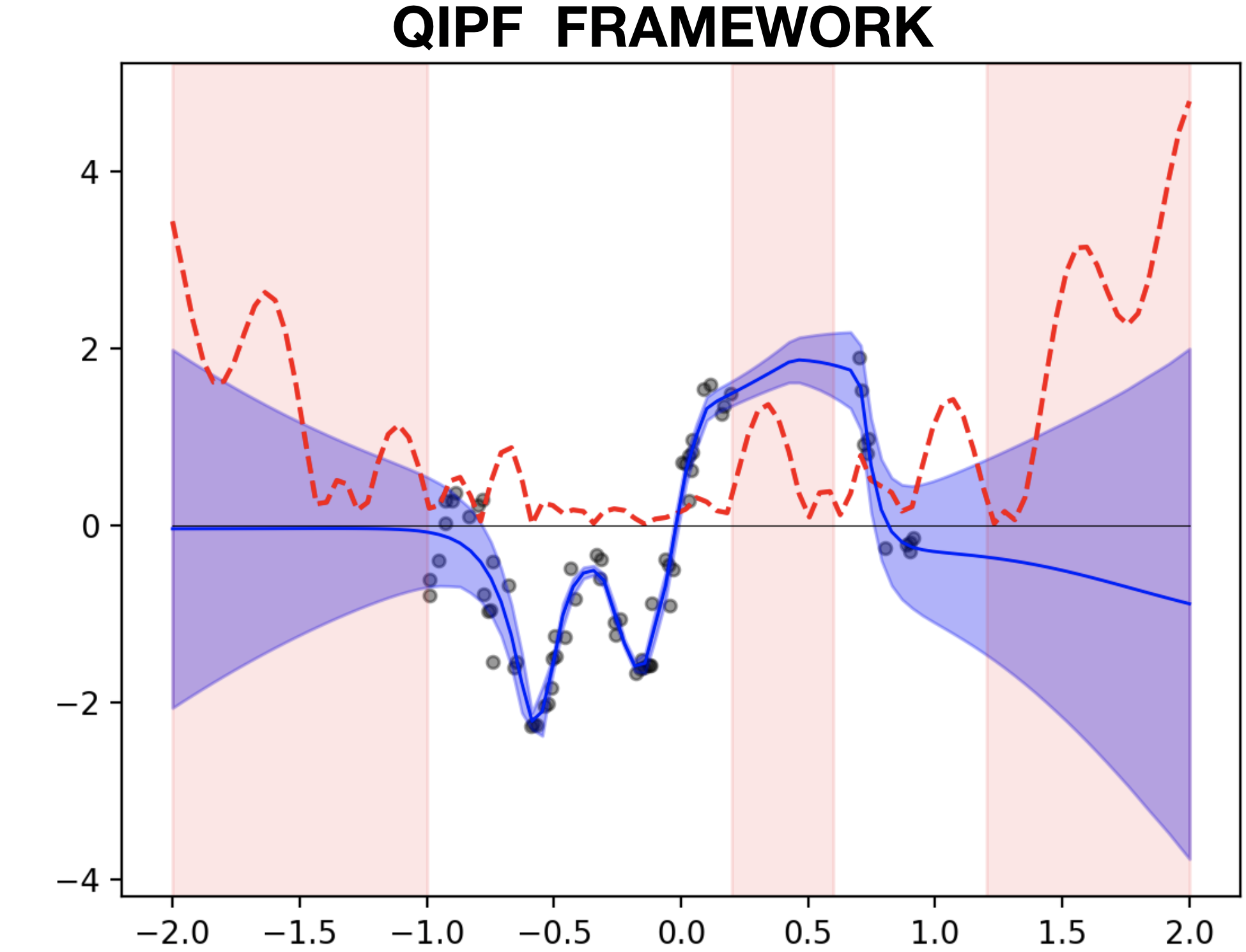}
    \caption{Synthesized data - II}
  \end{subfigure}

\bigskip

  \begin{subfigure}{\linewidth}
    \centering\includegraphics[scale=0.106]{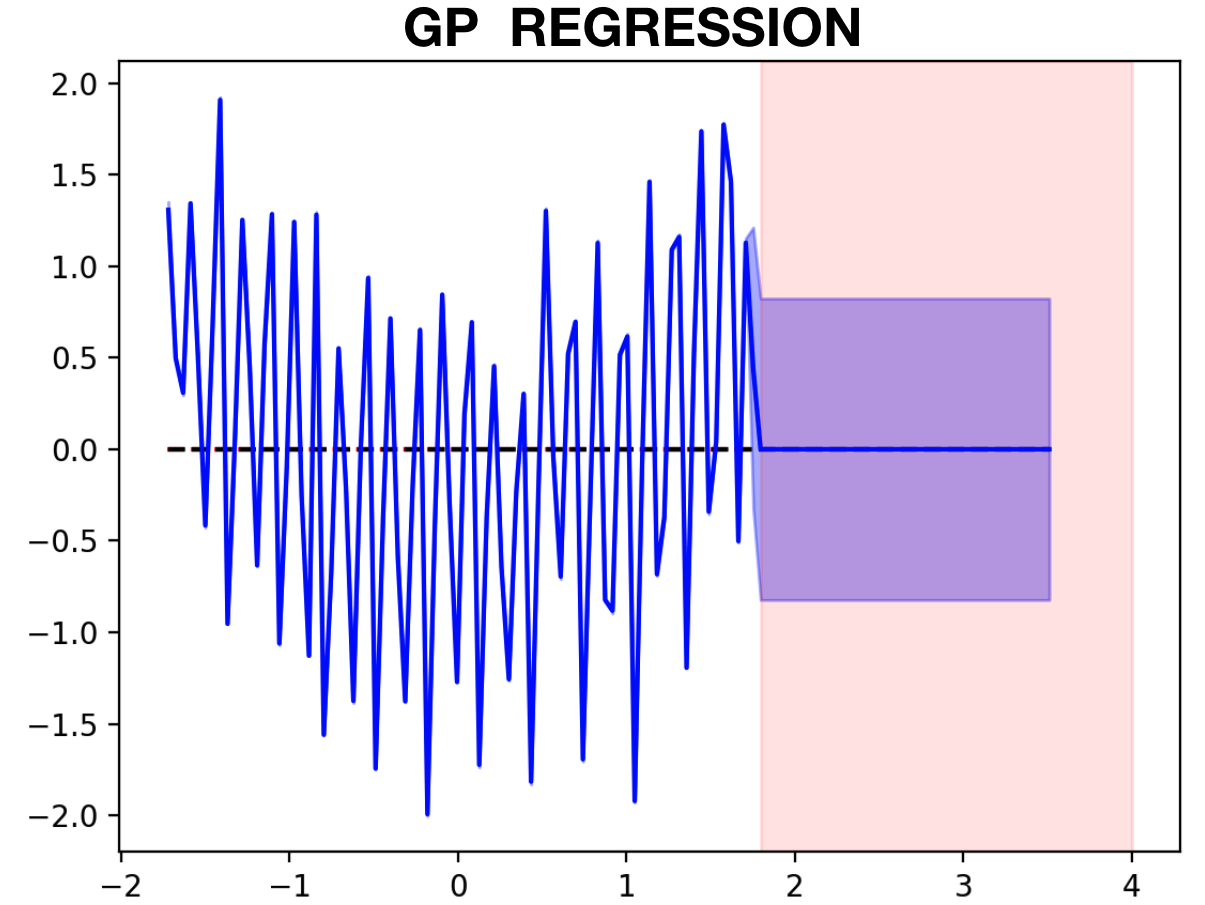}
    \centering\includegraphics[scale=0.106]{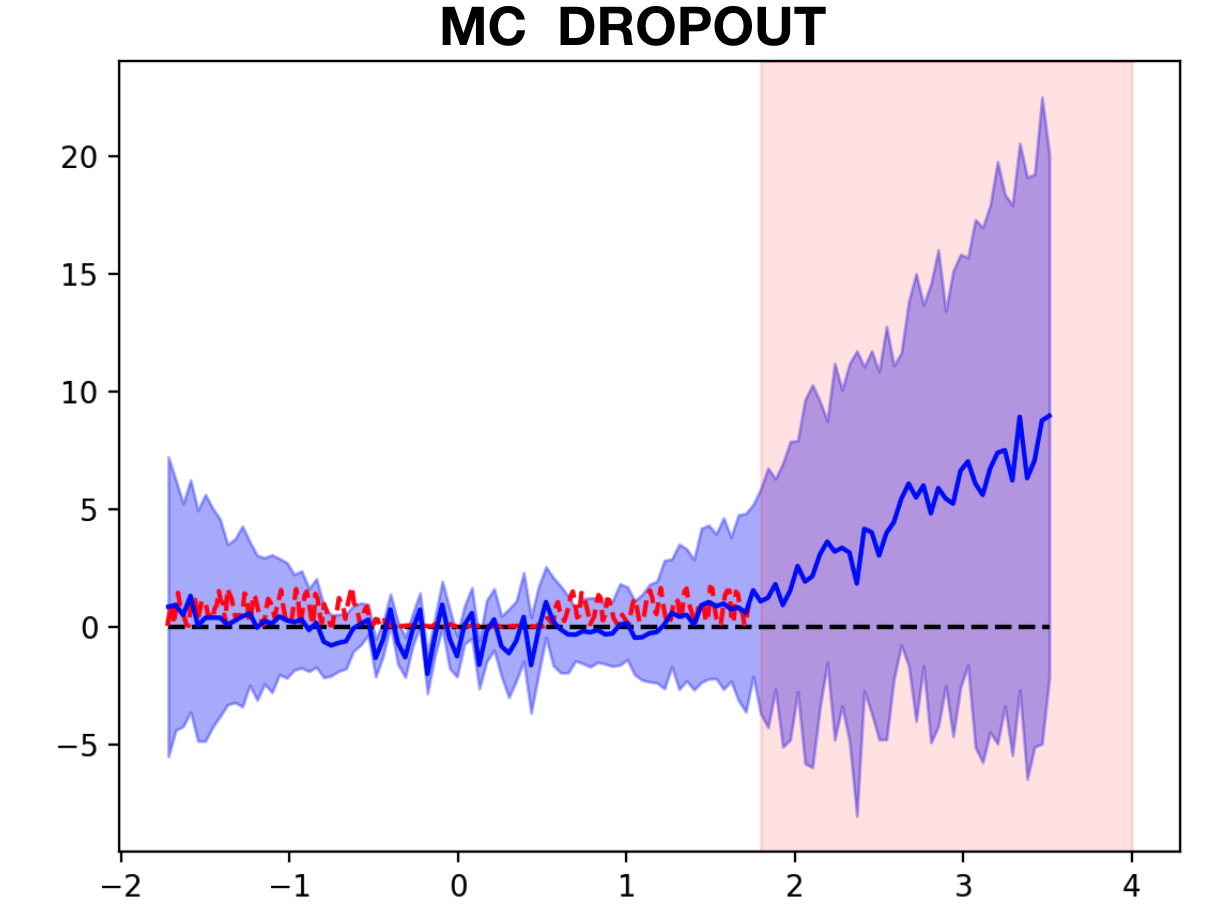}
    \centering\includegraphics[scale=0.106]{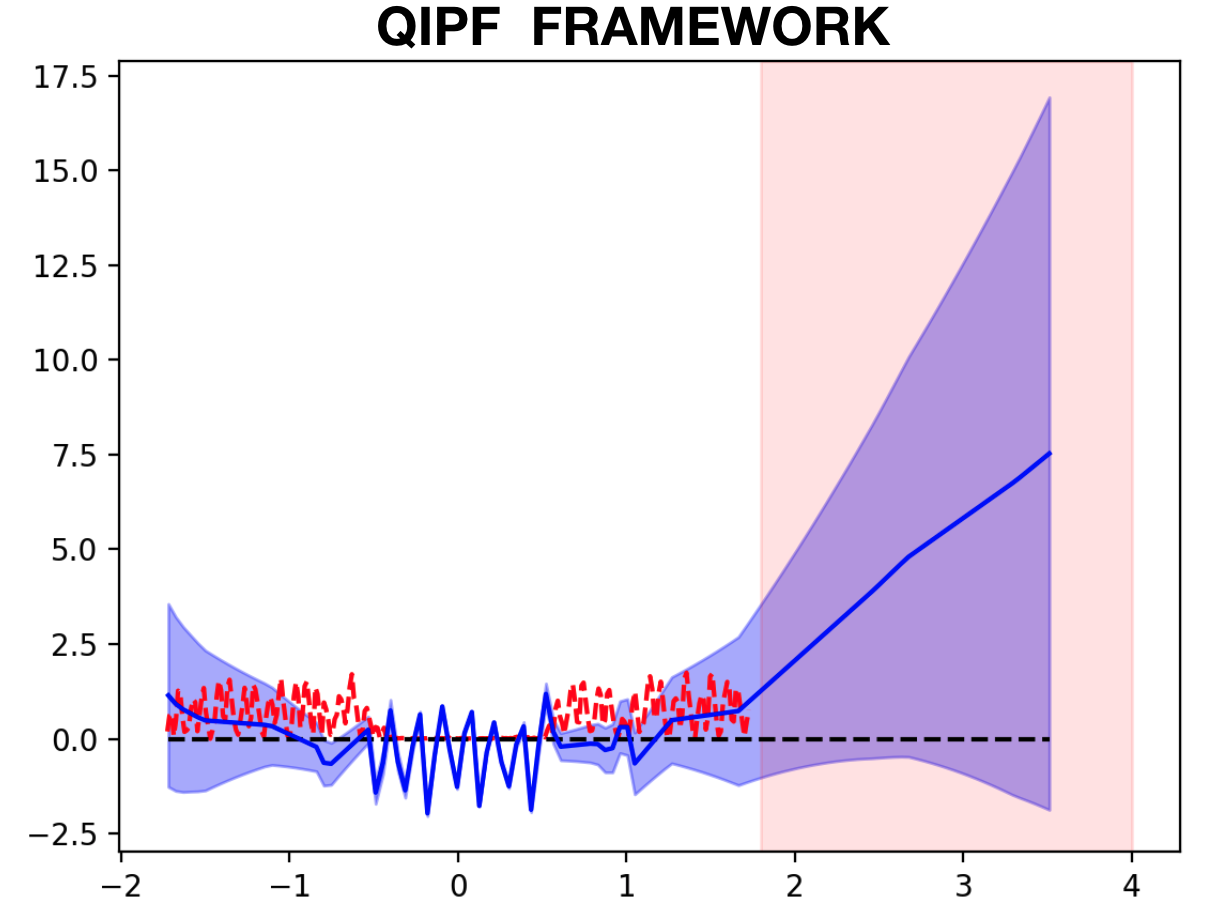}
	\caption{Mauna Loa CO2 data}
  \end{subfigure}
  \caption{Comparison of predictive UQ methods. Blue lines are model predictions. Blue shaded areas are their associated uncertainty ranges. Blue dots are training samples. Red dot lines are test prediction errors. Pink bands are untrained regions.}
  \label{mainres}
\end{figure}

\subsubsection*{Analysis of Results} 
The results of the different uncertainty quantifiers are shown in fig. \ref{mainres}. The blue line represents the predictions and the blue shaded regions depict the uncertainty ranges (standard deviation) at the prediction points quantified by the different methods (GPR, MC dropout and QIPF). Red dotted line indicates the test set prediction errors with respect to the generating function values at those points and the pink bands represent regions in the input space where no training samples were generated. For synthesized data I in fig. \ref{mainres}a, it can be observed that the GP regression model is able to better identify the generating signal dynamics than the neural network, producing low predictive errors for some distance outside of the training set. This is expected for small non-linear datasets where kernel methods outperform ANNs. It also produces well calibrated uncertainty ranges that are seen to be roughly proportional to the predictive errors. The QIPF framework can also be seen to produce uncertainty ranges that scale more proportionally with respect to the test error. MC dropout, on the other hand, produces comparatively less realistic uncertainties that soon converge to a constant level showing no change with respect to prediction errors thereafter. We observe wide disparities between the uncertainty quantification of the three algorithms in fig. \ref{mainres}b when the outliers are added to the training data (synthesized data I). We observe here that all models converge to the center of the outlier data when making predictions at those points. This is expected since the mean of such widely varying data points would represent the lowest error region for most learning models. However, we also notice that MC dropout becomes very overconfident with low uncertainties associated with its predictions at the outlier regions, which is opposite of what we would ideally expect. This is also reported in \citep{hern}. GP regression model also shows unrealistically low uncertainties in its predictions at the particular outlier points though it still produces increased uncertainty range around them. Also, unlike before, the GP model can be observed to converge to an unrealistic constant level of uncertainty as one goes outside of the training region, regardless of the increase in predictive errors. The QIPF framework, on the other hand, shows a remarkable property of \textit{increasing} its uncertainty range at the outlier regions, which is the ideal behavior. It is also seen to maintain its property of increasing proportionally with the predictive errors in all outside data regions. This indicates the sensitivity of the QIPF framework towards data variances and outliers in the training set. For synthesized data II (which consists of added normal noise), similar observations can be made related to the nature of uncertainties quantified by the different methods outside of the training domain. As before, the QIPF framework is seen to produce uncertainty estimates that increase more realistically outside the training domain and more proportional to the predictive errors when compared to MC dropout and GP regression. Both QIPF framework and MC dropout can be seen to be sensitive in their characterizations of uncertainty in the thin middle untrained band (given by the region (0.2, 0.7)) . However, we also observe here that, like before, MC dropout becomes unrealistically overconfident due to large variances in the training data pairs (on the left side of its corresponding graph in fig. \ref{mainres}c). Similar analysis on the CO2 data (fig. \ref{mainres}d) reveals GP regression to fit better than the other two models in the training region with very little error. However, it continues to be insensitive to predictive error outside the training domain by exhibiting a constant level of uncertainty when extrapolated during testing. The same trend of GP regression is reported in \citep{gal}. Both MC dropout and QIPF framework extrapolate more realistically in terms of uncertainty. We summarize the observations related to the QIPF framework from these results as follows.

\begin{itemize}
	\item The framework is observed to be robust towards training set outliers and is able to effectively capture the model's associated uncertainties with respect to them during testing. This can be attributed to the ability of the Gaussian RKHS in better capturing the true data distributions.	
	\item For all datasets, the QIPF framework is observed to produce uncertainty estimates that are more consistent with predictive errors in all regions of the data domain consequently exhibiting realistic uncertainty ranges for both model interpolation and extrapolation applications.
	\item The framework also exhibits increased sensitivity towards inherent data variances.
\end{itemize}

\subsection{Classification}
\subsubsection*{Toy Example: MNIST digit rotation}
\begin{figure}[!t]
  \begin{subfigure}{\linewidth}
  \centering\includegraphics[scale=0.4]{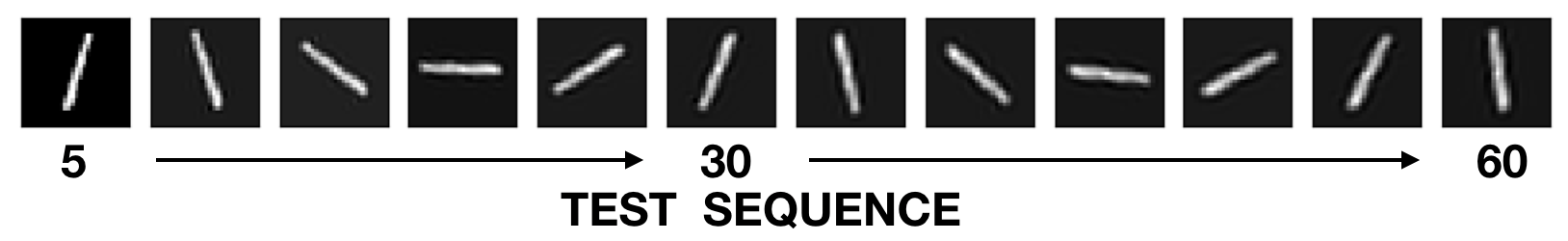}
    \caption{Testing sequence obtained by rotating MNIST digit of one.}
    \label{ro}
  \end{subfigure}
 \begin{subfigure}{\linewidth}
  \centering
   \includegraphics[scale=0.4]{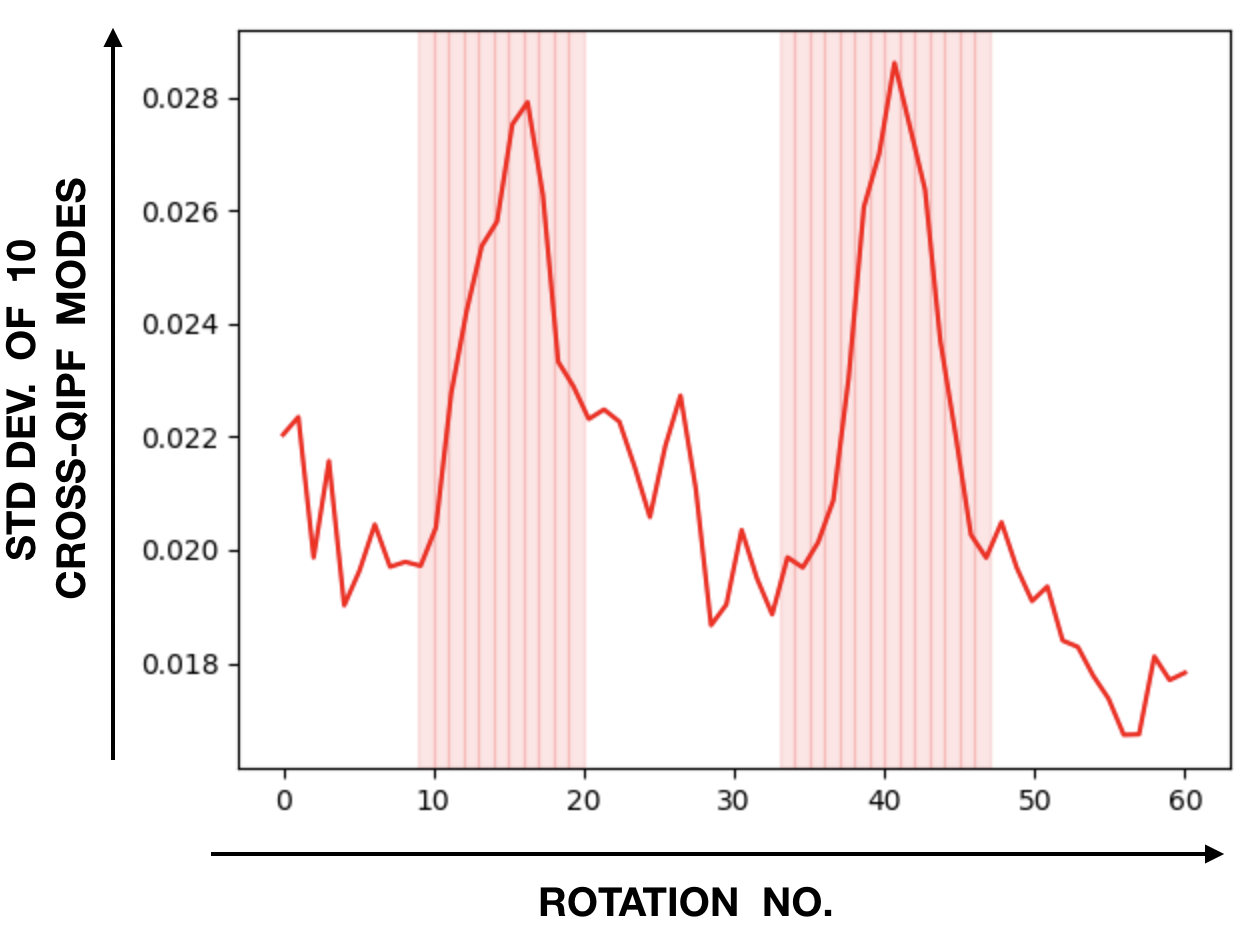}
    \caption{Uncertainty results of QIPF framework tested on rotated sequence of a digit.}
    \label{mn}
  \end{subfigure}
  \caption{Behavior of QIPF framework towards classification errors (pink bands)}
  \end{figure}

We also demonstrate the ability of the QIPF framework quantify uncertainties related to classification problems. We start with a toy example of training a network on MNIST dataset \citep{mnist}. The network consists of a ReLu activated and fully connected MLP with 3 hidden layers with 512 - 256 - 128 neurons respectively (from first to last hidden layers). We train the network without the implementation of dropout for 10 epochs using a batch size of 100. During testing, we rotated a single digit of 1 gradually (60 times uniformly) and fed each rotated version to the trained NN. During each test instance, we extracted the first 10 cross QIPF modes of the average node input value of the last layer (before thresholding) with respect to the activation outputs of hidden layer 1. In this application, we extracted the QIPF modes twice using different kernel widths (20x and 30x the Silverman bandwidth of the hidden layer 1 activation outputs) and considered the average of the two runs. Fig. \ref{ro} shows samples of the gradually rotated test sequence and fig. \ref{mn} shows the graph of the standard deviation of the 10 average QIPF modes at each test input. The pink bands represent the rotations at which the network produced incorrect classification results. One can observe the sharp rise of the standard deviation of the uncertainty modes at the misclassified test regions thus indicating that the framework produces uncertainty results consistent with the model classification errors.

\subsubsection*{Transfer Learning using VGG-16}
\begin{figure*}[!t]
\centering
\includegraphics[scale = 1]{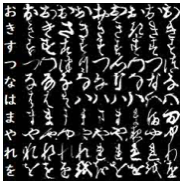}
\caption{Samples of the Kuzushiji-MNIST dataset. Each row corresponds to a different class and columns show samples corresponding to each class.}
\label{km}
\end{figure*}

In the recent years, there has been a surge of research interest in transfer learning, where large networks, pre-trained on  benchmark datasets, are used for efficient feature extraction from other datasets. They present a promising paradigm for increasing generalizability of deep learning models while requiring very low training to adapt to new datasets. This therefore presents a good venue for the utilization of epistemic model uncertainty quantification methods since we effectively want to explore the how well our framework quantifies predictive uncertainty in new data regions outside of the model's training system. To this end, we implement our framework on a VGG-16 CNN architecture \citep{vgg} pre-trained on the ImageNet dataset \citep{imgnt} and fine-tuned for a classification task on the Kuzushiji-MNIST dataset \citep{kuz}, which was recently introduced as a more challenging replacement to the MNIST dataset and consists of 10 classes of historical cursive Japanese letters (a subset of the full Kuzushiji-49 dataset). Some samples are shown in fig. \ref{km}.\par

\begin{figure*}[!t]
\centering
\includegraphics[scale = 0.7]{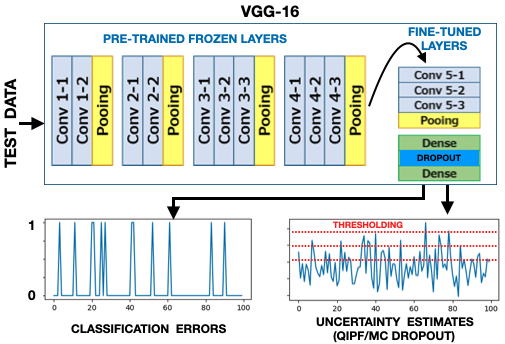}
\caption{Uncertainty quantification for transfer learning: QIPF framework and MC dropout are implemented on a Pre-trained and fine-tuned VGG-16 model while testing. The uncertainty estimates are then thresholded to detect classification errors.}
\label{arch}
\end{figure*}

\begin{figure}[!t]
\centering
\includegraphics[scale = 0.6]{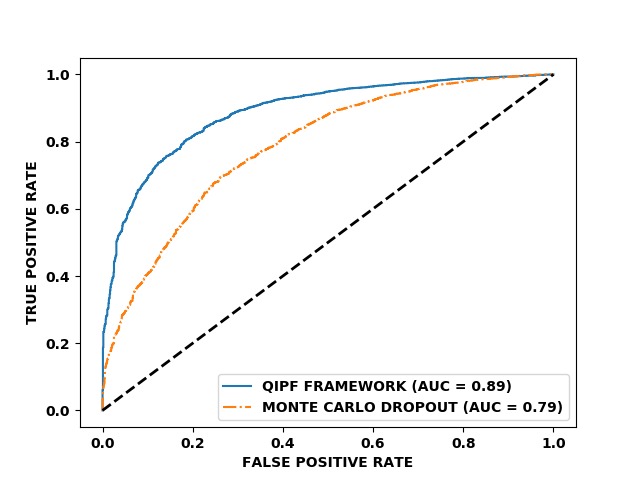}
\caption{Receiver operating characteristics (ROC) for detection of classification errors using QIPF and MC dropout uncertainty estimates. The QIPF framework is seen to have greater sensitivity and specificity with area under the curve as 0.89 compared to MC dropout's 0.79.}
\label{rs1}
\end{figure}

The K-MNIST dataset consists of 10 classes, a training set of 60000 samples and a test set of 10000 samples. We used CNN blocks of VGG-16 network pre-trained on the ImageNet dataset and froze all of its layers to training except for the last convolution block consisting of three CNN layers. For fine-tuning to K-MNIST, we further connected two dense layers with the first one consisting of 1024 neurons and the second one being the output layer with 10 neurons. Additionally we added a dropout layer in between them  with the rate set as 0.5. The overall architecture is depicted in fig. \ref{arch}. The total number of parameters of the network was 15,250,250 with the total number of trainable parameters being 7,614,986 (for fine-tuning). We fine-tuned the network on the training set of K-MNIST. We used the Adam optimizer and categorical cross-entropy loss function and trained the network for only 12 epochs with the batch size set as 128. We obtained a training accuracy of 99.50\% and a test accuracy of 96.96\%. We tested the network twice, first using MC dropout and then using the QIPF framework to obtain the predictive uncertainty associated with each test output. For MC dropout implementation, we inserted a dropout layer after each network layer. We used a small dropout rate of 0.1 for all CNN layers and rate of 0.2 for both the dense layers and implemented 500 forward passes at each test iteration. The standard deviation of the 500 network predictions was used as quantified predictive uncertainty value at each test iteration. For QIPF implementation, we used the \textit{weights} (instead of activation outputs) of the each layer of the network, except the last, as centers for computing the CIP associated with the last layer node output corresponding to the predicted class label (before softmax). We used a kernel width of 1 and slightly downsized the number of centers by using average pooling of weights at the layers so that the total number of centers were equal to 33736. Finally, we extracted 8 QIPF modes and measured their average to quantify uncertainty. For the uncertainties quantified by each method (MC dropout and QIPF framework), we performed thresholding (depicted in fig. \ref{arch}) to classify the obtained uncertainty values as either a classification error detection or no error detection. We then measured the true positive and false positive rates corresponding to the detections (for different threshold values) by comparing them with the actual classification errors of the network and plotted the receiver operating characteristics for both MC dropout and QIPF (fig. \ref{rs1}).\par

Hence we can see from fig. \ref{rs1} that the QIPF framework significantly outperforms the Monte Carlo dropout in terms of sensitivity and specificity in the detection of classification errors of the K-MNIST data for all threshold values. Moreover, the computation time of the QIPF framework was observed to be considerably faster (0.204 seconds for 33736 RKHS centers) compared to that of MC dropout (4.755 seconds for 500 runs) per test iteration. This shows that the QIPF framework is able to scale better to larger networks and datasets (both in terms of time complexity and performance) when compared to MC dropout. Furthermore, the results also reveal a potential utility of the QIPF framework in the domain of transfer learning. We intend to explore this application in the future.

\subsection{Benchmark Datasets}
We quantify and compare the quality of uncertainty estimates of our method with MC dropout over UCI datasets that are typically used as benchmarking data in various uncertainty quantification literature. We measure the quality of uncertainty estimate by quantifying how calibrated the uncertainty estimates are with respect to prediction errors. We chose datasets with diverse numbers of samples in order to test the framework on different forms of non-linearities. We train neural networks with 50 neurons in each hidden layer on 20 randomly generated train-test splits of the normalized UCI datasets (similar to the experimental framework of \citep{gal, hern}. A single kernel width of 1 was used in this case for extracting 10 QIPF states. We measured the RMSE of the uncertainty range (std. deviation of the QIPF modes at each test sample) with respect to the test error in each train-test split. This was done to measure how calibrated and scaled the estimated uncertainty range was with respect to the error. The average RMSE as well as its standard deviation for the 20 test splits are presented in table \ref{tab} for the framework's implementation on neural network architectures consisting of 1, 2 and 3 hidden layers respectively. It can be observed that the QIPF framework has lower RMSE values than MC dropout for most datasets in all network configurations thereby indicating that the estimated uncertainty using QIPF is more realistic.

\begin{table*}[!]
\begin{subtable}{\linewidth}
\centering
%\begin{center}
    \begin{tabular}{c c c c c c c c c c c c c c}
    \toprule
%    \addlinespace
    \multicolumn{2}{c}{\textbf{UCI Datasets}} & & & \textbf{N}  & \textbf{Q} & \textbf{MC Dropout} & & \textbf{QIPF} & \textbf{Significance}\\
%    \addlinespace
    \midrule
    \addlinespace
    \multicolumn{2}{c}{Yatch Hydrodynamics} & & & 308 &  6 & 0.332 +- 0.051 & & \textbf{0.204 +- 0.050} & 0.128\\ 
%    \addlinespace
   \multicolumn{2}{c}{Boston housing} & & & 506 &  13  & 0.246 +- 0.038 & & \textbf{0.234 +- 0.042} & 0.012\\ 
%   \addlinespace
   \multicolumn{2}{c}{Power Plant} & & & 9568 & 4 & 0.170 +- 0.035 & & \textbf{0.124 +- 0.035} & 0.046\\
%   \addlinespace
    \multicolumn{2}{c}{Concrete Strength} & & & 1030 & 8 & 0.234 +- 0.035 & & \textbf{0.221 +- 0.044} & 0.013\\
%    \addlinespace
    \multicolumn{2}{c}{Energy Efficiency} & & & 768 & 8 & 0.238 +- 0.028 & & 0.268 +- 0.061 & 0.030\\
%    \addlinespace
%    \multicolumn{2}{c}{Wine Quality Red} & & & & 1599 & & 11 & & 0.218 +- 0.028 & & 0.243 +- 0.031\\
%    \addlinespace
%    \addlinespace
    \bottomrule
    \end{tabular}
    \caption{1-hidden layer NN}
     \label{t1}
%    \end{center}
 \end{subtable}

%\begin{table*}[!]
%\begin{center}
	\begin{subtable}{\linewidth}
	\centering
    \begin{tabular}{c c c c c c c c c c c c c c}
    \toprule
%%    \addlinespace
%    \multicolumn{2}{c}{\textbf{UCI Datasets}} & & &  \textbf{N} & \textbf{Q} & \textbf{MC Dropout} & & \textbf{QIPF} & & \textbf{Significance Value}\\
%%    \addlinespace
%    \midrule
%%    \addlinespace
    \multicolumn{2}{c}{Yatch Hydrodynamics} & & & 308 & 6 & 0.294 +- 0.093 & & \textbf{0.169 +- 0.046} && 0.125\\ 
%    \addlinespace
   \multicolumn{2}{c}{Boston housing} & & & 506 & 13 & 0.222 +- 0.041 & & 0.234 +- 0.049 && 0.012\\ 
%   \addlinespace
   \multicolumn{2}{c}{Power Plant} & & & 9568 & 4 & 0.151 +- 0.050 & & \textbf{0.150 +- 0.057} && 0.001 \\
%   \addlinespace
    \multicolumn{2}{c}{Concrete Strength} & & & 1030 & 8 & 0.218 +- 0.036 & & \textbf{0.211 +- 0.043} &&  0.007\\
%    \addlinespace
    \multicolumn{2}{c}{Energy Efficiency} & & & 768 & 8 & 0.235 +- 0.032 & & 0.274 +- 0.046 &&  0.039\\
    \addlinespace
%    \multicolumn{2}{c}{Wine Quality Red} & & & & 1599 & & 11 & & - & & -\\
%    \addlinespace
%    \addlinespace
    \bottomrule
    \end{tabular}
    \caption{2-hidden layer NN}
        \label{t2}
%    \end{center}
 \end{subtable}
%\begin{table*}[!]
\begin{subtable}{\linewidth}
\centering
%\begin{center}
    \begin{tabular}{c c c c c c c c c c c c c c}
    \toprule
%%    \addlinespace
%    \multicolumn{2}{c}{\textbf{UCI Datasets}} & & & & \textbf{N} & & \textbf{Q} & & \textbf{MC Dropout} & & \textbf{QIPF}\\
%%    \addlinespace
%    \midrule
%%    \addlinespace
    \multicolumn{2}{c}{Yatch Hydrodynamics} & & & 308 & 6 & 0.226 +- 0.063 & & \textbf{0.167 +- 0.038} && 0.059\\ 
%    \addlinespace
   \multicolumn{2}{c}{Boston housing} & & & 506 & 13 & 0.223 +- 0.023 & & 0.234 +- 0.041 && 0.011\\ 
%   \addlinespace
   \multicolumn{2}{c}{Power Plant} & & & 9568 & 4 & 0.146 +- 0.038 & & \textbf{0.144 +- 0.05} && 0.002\\
%   \addlinespace
    \multicolumn{2}{c}{Concrete Strength} & & & 1030 & 8 & 0.220 +- 0.030 & & \textbf{0.204 +- 0.035} && 0.016\\
%    \addlinespace
    \multicolumn{2}{c}{Energy Efficiency} & & & 768 & 8 & 0.263 +- 0.034 & & \textbf{0.240 +- 0.048} && 0.023\\
%    \addlinespace
%    \multicolumn{2}{c}{Wine Quality Red} & & & & 1599 & & 11 & & 0.218 +- 0.028 & & 0.243 +- 0.031\\
%    \addlinespace
%    \addlinespace
    \bottomrule
    \end{tabular}
    \caption{3-hidden layer NN}
    \label{t3}
%    \end{center}
 \end{subtable}
 \caption{Normalized RMSE between the standard deviation of quantified uncertainty and the test error. \textbf{N} denotes the number of samples and \textbf{Q} denotes data dimensionality.}
 \label{tab}
 \end{table*}

\section*{Conclusion}
In this paper, we introduced a new information theoretic approach for quantifying uncertainty that is inspired by quantum physical principles and concepts. We formulated a new uncertainty moment decomposition framework for data   and models through a systematic and effective quantification gradient flow of data PDF by utilizing the structure and mathematical guarantees provided by the RKHS. The key advantages offered by our framework include its significantly high precision in detecting epistemic uncertainty at all points in the data/model space, ability to provide a single-shot quantification of uncertainty and ability to scale better to larger networks and models in terms of performance and computational cost. We gave pedagogical examples to show how our model provides a multi-scale characterization of the tails of the data PDF where epistemic uncertainty is maximum. We demonstrated how our framework can be utilized as a powerful tool for the predictive uncertainty quantification of models in both regression and classification tasks involving diverse and differently sized networks and datasets. In the future, we intend to explore the performance of the framework on larger model architectures in the domain of uncertainty quantification and transferability. We also intend to conduct more data-based implementations of the framework in order to explore its utility in the signal processing domain.

\subsection*{Acknowledgments}
We would like to acknowledge the support provided for this work by ONR under agreement number N00014-18-1- 2306 and DARPA under agreement number FA9453-18-1-0039.

\bibliographystyle{apacite}

\begin{thebibliography}{}

\end{thebibliography}


\begin{thebibliography}{}

\bibitem [\protect \citeauthoryear {%
Ahn%
, Choi%
, Dai%
, Sohn%
\BCBL {}\ \BBA {} Yang%
}{%
Ahn%
\ \protect \BOthers {.}}{%
{\protect \APACyear {2018}}%
}]{%
q2}
\APACinsertmetastar {%
q2}%
\begin{APACrefauthors}%
Ahn, K.%
, Choi, M.%
, Dai, B.%
, Sohn, S.%
\BCBL {}\ \BBA {} Yang, B.%
\end{APACrefauthors}%
\unskip\
\newblock
\APACrefYearMonthDay{2018}{}{}.
\newblock
{\BBOQ}\APACrefatitle {Modeling stock return distributions with a quantum
  harmonic oscillator} {Modeling stock return distributions with a quantum
  harmonic oscillator}.{\BBCQ}
\newblock
\APACjournalVolNumPages{EPL (Europhysics Letters)}{120}{3}{38003}.
\PrintBackRefs{\CurrentBib}

\bibitem [\protect \citeauthoryear {%
Aronszajn%
}{%
Aronszajn%
}{%
{\protect \APACyear {1950}}%
}]{%
aron}
\APACinsertmetastar {%
aron}%
\begin{APACrefauthors}%
Aronszajn, N.%
\end{APACrefauthors}%
\unskip\
\newblock
\APACrefYearMonthDay{1950}{}{}.
\newblock
{\BBOQ}\APACrefatitle {Theory of reproducing kernels} {Theory of reproducing
  kernels}.{\BBCQ}
\newblock
\APACjournalVolNumPages{Transactions of the American mathematical
  society}{68}{3}{337--404}.
\PrintBackRefs{\CurrentBib}

\bibitem [\protect \citeauthoryear {%
Belavkin%
}{%
Belavkin%
}{%
{\protect \APACyear {1989}}%
}]{%
bel}
\APACinsertmetastar {%
bel}%
\begin{APACrefauthors}%
Belavkin, V.%
\end{APACrefauthors}%
\unskip\
\newblock
\APACrefYearMonthDay{1989}{}{}.
\newblock
{\BBOQ}\APACrefatitle {A new wave equation for a continuous nondemolition
  measurement} {A new wave equation for a continuous nondemolition
  measurement}.{\BBCQ}
\newblock
\APACjournalVolNumPages{Physics letters A}{140}{7-8}{355--358}.
\PrintBackRefs{\CurrentBib}

\bibitem [\protect \citeauthoryear {%
Bergman%
}{%
Bergman%
}{%
{\protect \APACyear {1970}}%
}]{%
g1}
\APACinsertmetastar {%
g1}%
\begin{APACrefauthors}%
Bergman, S.%
\end{APACrefauthors}%
\unskip\
\newblock
\APACrefYear{1970}.
\newblock
\APACrefbtitle {The kernel function and conformal mapping} {The kernel function
  and conformal mapping}\ (\BVOL~5).
\newblock
\APACaddressPublisher{}{American Mathematical Soc.}
\PrintBackRefs{\CurrentBib}

\bibitem [\protect \citeauthoryear {%
Berlinet%
\ \BBA {} Thomas-Agnan%
}{%
Berlinet%
\ \BBA {} Thomas-Agnan%
}{%
{\protect \APACyear {2011}}%
}]{%
embor}
\APACinsertmetastar {%
embor}%
\begin{APACrefauthors}%
Berlinet, A.%
\BCBT {}\ \BBA {} Thomas-Agnan, C.%
\end{APACrefauthors}%
\unskip\
\newblock
\APACrefYear{2011}.
\newblock
\APACrefbtitle {Reproducing kernel Hilbert spaces in probability and
  statistics} {Reproducing kernel hilbert spaces in probability and
  statistics}.
\newblock
\APACaddressPublisher{}{Springer Science \& Business Media}.
\PrintBackRefs{\CurrentBib}

\bibitem [\protect \citeauthoryear {%
Bishop%
}{%
Bishop%
}{%
{\protect \APACyear {1995}}%
}]{%
bishop}
\APACinsertmetastar {%
bishop}%
\begin{APACrefauthors}%
Bishop, C\BPBI M.%
\end{APACrefauthors}%
\unskip\
\newblock
\APACrefYearMonthDay{1995}{}{}.
\newblock
{\BBOQ}\APACrefatitle {Bayesian methods for neural networks} {Bayesian methods
  for neural networks}.{\BBCQ}
\newblock

\PrintBackRefs{\CurrentBib}

\bibitem [\protect \citeauthoryear {%
Boltzmann%
}{%
Boltzmann%
}{%
{\protect \APACyear {1877}}%
}]{%
bol}
\APACinsertmetastar {%
bol}%
\begin{APACrefauthors}%
Boltzmann, L.%
\end{APACrefauthors}%
\unskip\
\newblock
\APACrefYear{1877}.
\newblock
\APACrefbtitle {{\"U}ber die Beziehung zwischen dem zweiten Hauptsatze des
  mechanischen W{\"a}rmetheorie und der Wahrscheinlichkeitsrechnung, respective
  den S{\"a}tzen {\"u}ber das W{\"a}rmegleichgewicht} {{\"U}ber die beziehung
  zwischen dem zweiten hauptsatze des mechanischen w{\"a}rmetheorie und der
  wahrscheinlichkeitsrechnung, respective den s{\"a}tzen {\"u}ber das
  w{\"a}rmegleichgewicht}.
\newblock
\APACaddressPublisher{}{Kk Hof-und Staatsdruckerei}.
\PrintBackRefs{\CurrentBib}

\bibitem [\protect \citeauthoryear {%
Chen%
, Zhao%
, Zhu%
\BCBL {}\ \BBA {} Pr{\'\i}ncipe%
}{%
Chen%
\ \protect \BOthers {.}}{%
{\protect \APACyear {2011}}%
}]{%
qua}
\APACinsertmetastar {%
qua}%
\begin{APACrefauthors}%
Chen, B.%
, Zhao, S.%
, Zhu, P.%
\BCBL {}\ \BBA {} Pr{\'\i}ncipe, J\BPBI C.%
\end{APACrefauthors}%
\unskip\
\newblock
\APACrefYearMonthDay{2011}{}{}.
\newblock
{\BBOQ}\APACrefatitle {Quantized kernel least mean square algorithm} {Quantized
  kernel least mean square algorithm}.{\BBCQ}
\newblock
\APACjournalVolNumPages{IEEE Transactions on Neural Networks and Learning
  Systems}{23}{1}{22--32}.
\PrintBackRefs{\CurrentBib}

\bibitem [\protect \citeauthoryear {%
Clanuwat%
\ \protect \BOthers {.}}{%
Clanuwat%
\ \protect \BOthers {.}}{%
{\protect \APACyear {2018}}%
}]{%
kuz}
\APACinsertmetastar {%
kuz}%
\begin{APACrefauthors}%
Clanuwat, T.%
, Bober-Irizar, M.%
, Kitamoto, A.%
, Lamb, A.%
, Yamamoto, K.%
\BCBL {}\ \BBA {} Ha, D.%
\end{APACrefauthors}%
\unskip\
\newblock
\APACrefYearMonthDay{2018}{}{}.
\newblock
{\BBOQ}\APACrefatitle {Deep learning for classical Japanese literature} {Deep
  learning for classical japanese literature}.{\BBCQ}
\newblock
\APACjournalVolNumPages{arXiv preprint arXiv:1812.01718}{}{}{}.
\PrintBackRefs{\CurrentBib}

\bibitem [\protect \citeauthoryear {%
Deng%
\ \protect \BOthers {.}}{%
Deng%
\ \protect \BOthers {.}}{%
{\protect \APACyear {2009}}%
}]{%
imgnt}
\APACinsertmetastar {%
imgnt}%
\begin{APACrefauthors}%
Deng, J.%
, Dong, W.%
, Socher, R.%
, Li, L\BHBI J.%
, Li, K.%
\BCBL {}\ \BBA {} Fei-Fei, L.%
\end{APACrefauthors}%
\unskip\
\newblock
\APACrefYearMonthDay{2009}{}{}.
\newblock
{\BBOQ}\APACrefatitle {Imagenet: A large-scale hierarchical image database}
  {Imagenet: A large-scale hierarchical image database}.{\BBCQ}
\newblock
\BIn{} \APACrefbtitle {2009 IEEE conference on computer vision and pattern
  recognition} {2009 ieee conference on computer vision and pattern
  recognition}\ (\BPGS\ 248--255).
\PrintBackRefs{\CurrentBib}

\bibitem [\protect \citeauthoryear {%
Fang%
, Li%
\BCBL {}\ \BBA {} Sudjianto%
}{%
Fang%
\ \protect \BOthers {.}}{%
{\protect \APACyear {2005}}%
}]{%
sur1}
\APACinsertmetastar {%
sur1}%
\begin{APACrefauthors}%
Fang, K\BHBI T.%
, Li, R.%
\BCBL {}\ \BBA {} Sudjianto, A.%
\end{APACrefauthors}%
\unskip\
\newblock
\APACrefYear{2005}.
\newblock
\APACrefbtitle {Design and modeling for computer experiments} {Design and
  modeling for computer experiments}.
\newblock
\APACaddressPublisher{}{Chapman and Hall/CRC}.
\PrintBackRefs{\CurrentBib}

\bibitem [\protect \citeauthoryear {%
Fisher%
}{%
Fisher%
}{%
{\protect \APACyear {1922}}%
}]{%
fish}
\APACinsertmetastar {%
fish}%
\begin{APACrefauthors}%
Fisher, R\BPBI A.%
\end{APACrefauthors}%
\unskip\
\newblock
\APACrefYearMonthDay{1922}{}{}.
\newblock
{\BBOQ}\APACrefatitle {On the mathematical foundations of theoretical
  statistics} {On the mathematical foundations of theoretical
  statistics}.{\BBCQ}
\newblock
\APACjournalVolNumPages{Philosophical Transactions of the Royal Society of
  London. Series A, Containing Papers of a Mathematical or Physical
  Character}{222}{594-604}{309--368}.
\PrintBackRefs{\CurrentBib}

\bibitem [\protect \citeauthoryear {%
Forrester%
, Sobester%
\BCBL {}\ \BBA {} Keane%
}{%
Forrester%
\ \protect \BOthers {.}}{%
{\protect \APACyear {2008}}%
}]{%
sur2}
\APACinsertmetastar {%
sur2}%
\begin{APACrefauthors}%
Forrester, A.%
, Sobester, A.%
\BCBL {}\ \BBA {} Keane, A.%
\end{APACrefauthors}%
\unskip\
\newblock
\APACrefYear{2008}.
\newblock
\APACrefbtitle {Engineering design via surrogate modelling: a practical guide}
  {Engineering design via surrogate modelling: a practical guide}.
\newblock
\APACaddressPublisher{}{John Wiley \& Sons}.
\PrintBackRefs{\CurrentBib}

\bibitem [\protect \citeauthoryear {%
Frieden%
}{%
Frieden%
}{%
{\protect \APACyear {2004}}%
}]{%
roy2}
\APACinsertmetastar {%
roy2}%
\begin{APACrefauthors}%
Frieden, B\BPBI R.%
\end{APACrefauthors}%
\unskip\
\newblock
\APACrefYear{2004}.
\newblock
\APACrefbtitle {Science from Fisher information: a unification} {Science from
  fisher information: a unification}.
\newblock
\APACaddressPublisher{}{Cambridge University Press}.
\PrintBackRefs{\CurrentBib}

\bibitem [\protect \citeauthoryear {%
Frieden%
\ \BBA {} Hawkins%
}{%
Frieden%
\ \BBA {} Hawkins%
}{%
{\protect \APACyear {2010}}%
}]{%
roy1}
\APACinsertmetastar {%
roy1}%
\begin{APACrefauthors}%
Frieden, B\BPBI R.%
\BCBT {}\ \BBA {} Hawkins, R\BPBI J.%
\end{APACrefauthors}%
\unskip\
\newblock
\APACrefYearMonthDay{2010}{}{}.
\newblock
{\BBOQ}\APACrefatitle {Quantifying system order for full and partial coarse
  graining} {Quantifying system order for full and partial coarse
  graining}.{\BBCQ}
\newblock
\APACjournalVolNumPages{Physical Review E}{82}{6}{066117}.
\PrintBackRefs{\CurrentBib}

\bibitem [\protect \citeauthoryear {%
Gal%
\ \BBA {} Ghahramani%
}{%
Gal%
\ \BBA {} Ghahramani%
}{%
{\protect \APACyear {2016}}%
}]{%
gal}
\APACinsertmetastar {%
gal}%
\begin{APACrefauthors}%
Gal, Y.%
\BCBT {}\ \BBA {} Ghahramani, Z.%
\end{APACrefauthors}%
\unskip\
\newblock
\APACrefYearMonthDay{2016}{}{}.
\newblock
{\BBOQ}\APACrefatitle {Dropout as a bayesian approximation: Representing model
  uncertainty in deep learning} {Dropout as a bayesian approximation:
  Representing model uncertainty in deep learning}.{\BBCQ}
\newblock
\BIn{} \APACrefbtitle {international conference on machine learning}
  {international conference on machine learning}\ (\BPGS\ 1050--1059).
\PrintBackRefs{\CurrentBib}

\bibitem [\protect \citeauthoryear {%
Graves%
}{%
Graves%
}{%
{\protect \APACyear {2011}}%
}]{%
graves}
\APACinsertmetastar {%
graves}%
\begin{APACrefauthors}%
Graves, A.%
\end{APACrefauthors}%
\unskip\
\newblock
\APACrefYearMonthDay{2011}{}{}.
\newblock
{\BBOQ}\APACrefatitle {Practical variational inference for neural networks}
  {Practical variational inference for neural networks}.{\BBCQ}
\newblock
\BIn{} \APACrefbtitle {Advances in neural information processing systems}
  {Advances in neural information processing systems}\ (\BPGS\ 2348--2356).
\PrintBackRefs{\CurrentBib}

\bibitem [\protect \citeauthoryear {%
Greengard%
\ \BBA {} Strain%
}{%
Greengard%
\ \BBA {} Strain%
}{%
{\protect \APACyear {1991}}%
}]{%
fgt}
\APACinsertmetastar {%
fgt}%
\begin{APACrefauthors}%
Greengard, L.%
\BCBT {}\ \BBA {} Strain, J.%
\end{APACrefauthors}%
\unskip\
\newblock
\APACrefYearMonthDay{1991}{}{}.
\newblock
{\BBOQ}\APACrefatitle {The fast Gauss transform} {The fast gauss
  transform}.{\BBCQ}
\newblock
\APACjournalVolNumPages{SIAM Journal on Scientific and Statistical
  Computing}{12}{1}{79--94}.
\PrintBackRefs{\CurrentBib}

\bibitem [\protect \citeauthoryear {%
Grenander%
}{%
Grenander%
}{%
{\protect \APACyear {1950}}%
}]{%
gre}
\APACinsertmetastar {%
gre}%
\begin{APACrefauthors}%
Grenander, U.%
\end{APACrefauthors}%
\unskip\
\newblock
\APACrefYearMonthDay{1950}{}{}.
\newblock
{\BBOQ}\APACrefatitle {Stochastic processes and statistical inference}
  {Stochastic processes and statistical inference}.{\BBCQ}
\newblock
\APACjournalVolNumPages{Arkiv f{\"o}r matematik}{1}{3}{195--277}.
\PrintBackRefs{\CurrentBib}

\bibitem [\protect \citeauthoryear {%
Hahn%
\ \BBA {} Shapiro%
}{%
Hahn%
\ \BBA {} Shapiro%
}{%
{\protect \APACyear {1967}}%
}]{%
r5}
\APACinsertmetastar {%
r5}%
\begin{APACrefauthors}%
Hahn, G\BPBI J.%
\BCBT {}\ \BBA {} Shapiro, S\BPBI S.%
\end{APACrefauthors}%
\unskip\
\newblock
\APACrefYearMonthDay{1967}{}{}.
\newblock
\APACrefbtitle {Statistical models in engineering.} {Statistical models in
  engineering.}\ \APACbVolEdTR{}{\BTR{}}.
\PrintBackRefs{\CurrentBib}

\bibitem [\protect \citeauthoryear {%
Hermite%
}{%
Hermite%
}{%
{\protect \APACyear {1864}}%
}]{%
hm}
\APACinsertmetastar {%
hm}%
\begin{APACrefauthors}%
Hermite, M.%
\end{APACrefauthors}%
\unskip\
\newblock
\APACrefYear{1864}.
\newblock
\APACrefbtitle {Sur un nouveau d{\'e}veloppement en s{\'e}rie des fonctions}
  {Sur un nouveau d{\'e}veloppement en s{\'e}rie des fonctions}.
\newblock
\APACaddressPublisher{}{Imprimerie de Gauthier-Villars}.
\PrintBackRefs{\CurrentBib}

\bibitem [\protect \citeauthoryear {%
Hern{\'a}ndez-Lobato%
\ \BBA {} Adams%
}{%
Hern{\'a}ndez-Lobato%
\ \BBA {} Adams%
}{%
{\protect \APACyear {2015}}%
}]{%
hern}
\APACinsertmetastar {%
hern}%
\begin{APACrefauthors}%
Hern{\'a}ndez-Lobato, J\BPBI M.%
\BCBT {}\ \BBA {} Adams, R.%
\end{APACrefauthors}%
\unskip\
\newblock
\APACrefYearMonthDay{2015}{}{}.
\newblock
{\BBOQ}\APACrefatitle {Probabilistic backpropagation for scalable learning of
  bayesian neural networks} {Probabilistic backpropagation for scalable
  learning of bayesian neural networks}.{\BBCQ}
\newblock
\BIn{} \APACrefbtitle {International Conference on Machine Learning}
  {International conference on machine learning}\ (\BPGS\ 1861--1869).
\PrintBackRefs{\CurrentBib}

\bibitem [\protect \citeauthoryear {%
Hoffman%
, Blei%
, Wang%
\BCBL {}\ \BBA {} Paisley%
}{%
Hoffman%
\ \protect \BOthers {.}}{%
{\protect \APACyear {2013}}%
}]{%
hof}
\APACinsertmetastar {%
hof}%
\begin{APACrefauthors}%
Hoffman, M\BPBI D.%
, Blei, D\BPBI M.%
, Wang, C.%
\BCBL {}\ \BBA {} Paisley, J.%
\end{APACrefauthors}%
\unskip\
\newblock
\APACrefYearMonthDay{2013}{}{}.
\newblock
{\BBOQ}\APACrefatitle {Stochastic variational inference} {Stochastic
  variational inference}.{\BBCQ}
\newblock
\APACjournalVolNumPages{The Journal of Machine Learning
  Research}{14}{1}{1303--1347}.
\PrintBackRefs{\CurrentBib}

\bibitem [\protect \citeauthoryear {%
Hofmann%
, Sch{\"o}lkopf%
\BCBL {}\ \BBA {} Smola%
}{%
Hofmann%
\ \protect \BOthers {.}}{%
{\protect \APACyear {2008}}%
}]{%
hoff}
\APACinsertmetastar {%
hoff}%
\begin{APACrefauthors}%
Hofmann, T.%
, Sch{\"o}lkopf, B.%
\BCBL {}\ \BBA {} Smola, A\BPBI J.%
\end{APACrefauthors}%
\unskip\
\newblock
\APACrefYearMonthDay{2008}{}{}.
\newblock
{\BBOQ}\APACrefatitle {Kernel methods in machine learning} {Kernel methods in
  machine learning}.{\BBCQ}
\newblock
\APACjournalVolNumPages{The annals of statistics}{}{}{1171--1220}.
\PrintBackRefs{\CurrentBib}

\bibitem [\protect \citeauthoryear {%
Karhunen%
}{%
Karhunen%
}{%
{\protect \APACyear {1946}}%
}]{%
kar}
\APACinsertmetastar {%
kar}%
\begin{APACrefauthors}%
Karhunen, K.%
\end{APACrefauthors}%
\unskip\
\newblock
\APACrefYearMonthDay{1946}{}{}.
\newblock
{\BBOQ}\APACrefatitle {Zur spektraltheorie stochastischer prozesse} {Zur
  spektraltheorie stochastischer prozesse}.{\BBCQ}
\newblock
\APACjournalVolNumPages{Ann. Acad. Sci. Fennicae, AI}{34}{}{}.
\PrintBackRefs{\CurrentBib}

\bibitem [\protect \citeauthoryear {%
Keeling%
\ \BBA {} Whorf%
}{%
Keeling%
\ \BBA {} Whorf%
}{%
{\protect \APACyear {1991}}%
}]{%
keeling}
\APACinsertmetastar {%
keeling}%
\begin{APACrefauthors}%
Keeling, C.%
\BCBT {}\ \BBA {} Whorf, T.%
\end{APACrefauthors}%
\unskip\
\newblock
\APACrefYearMonthDay{1991}{}{}.
\newblock
{\BBOQ}\APACrefatitle {Mauna Loa atmospheric CO2—modern record} {Mauna loa
  atmospheric co2—modern record}.{\BBCQ}
\newblock
\APACjournalVolNumPages{Trends}{91}{}{12--15}.
\PrintBackRefs{\CurrentBib}

\bibitem [\protect \citeauthoryear {%
Kullback%
\ \BBA {} Leibler%
}{%
Kullback%
\ \BBA {} Leibler%
}{%
{\protect \APACyear {1951}}%
}]{%
r3}
\APACinsertmetastar {%
r3}%
\begin{APACrefauthors}%
Kullback, S.%
\BCBT {}\ \BBA {} Leibler, R\BPBI A.%
\end{APACrefauthors}%
\unskip\
\newblock
\APACrefYearMonthDay{1951}{}{}.
\newblock
{\BBOQ}\APACrefatitle {On information and sufficiency} {On information and
  sufficiency}.{\BBCQ}
\newblock
\APACjournalVolNumPages{The annals of mathematical statistics}{22}{1}{79--86}.
\PrintBackRefs{\CurrentBib}

\bibitem [\protect \citeauthoryear {%
Lakshminarayanan%
, Pritzel%
\BCBL {}\ \BBA {} Blundell%
}{%
Lakshminarayanan%
\ \protect \BOthers {.}}{%
{\protect \APACyear {2017}}%
}]{%
laks}
\APACinsertmetastar {%
laks}%
\begin{APACrefauthors}%
Lakshminarayanan, B.%
, Pritzel, A.%
\BCBL {}\ \BBA {} Blundell, C.%
\end{APACrefauthors}%
\unskip\
\newblock
\APACrefYearMonthDay{2017}{}{}.
\newblock
{\BBOQ}\APACrefatitle {Simple and scalable predictive uncertainty estimation
  using deep ensembles} {Simple and scalable predictive uncertainty estimation
  using deep ensembles}.{\BBCQ}
\newblock
\BIn{} \APACrefbtitle {Advances in Neural Information Processing Systems}
  {Advances in neural information processing systems}\ (\BPGS\ 6402--6413).
\PrintBackRefs{\CurrentBib}

\bibitem [\protect \citeauthoryear {%
LeCun%
, Bengio%
\BCBL {}\ \BBA {} Hinton%
}{%
LeCun%
\ \protect \BOthers {.}}{%
{\protect \APACyear {2015}}%
}]{%
lec}
\APACinsertmetastar {%
lec}%
\begin{APACrefauthors}%
LeCun, Y.%
, Bengio, Y.%
\BCBL {}\ \BBA {} Hinton, G.%
\end{APACrefauthors}%
\unskip\
\newblock
\APACrefYearMonthDay{2015}{}{}.
\newblock
{\BBOQ}\APACrefatitle {Deep learning} {Deep learning}.{\BBCQ}
\newblock
\APACjournalVolNumPages{nature}{521}{7553}{436}.
\PrintBackRefs{\CurrentBib}

\bibitem [\protect \citeauthoryear {%
LeCun%
, Cortes%
\BCBL {}\ \BBA {} Burges%
}{%
LeCun%
\ \protect \BOthers {.}}{%
{\protect \APACyear {1998}}%
}]{%
mnist}
\APACinsertmetastar {%
mnist}%
\begin{APACrefauthors}%
LeCun, Y.%
, Cortes, C.%
\BCBL {}\ \BBA {} Burges, C\BPBI J.%
\end{APACrefauthors}%
\unskip\
\newblock
\APACrefYearMonthDay{1998}{}{}.
\newblock
{\BBOQ}\APACrefatitle {The MNIST database of handwritten digits, 1998} {The
  mnist database of handwritten digits, 1998}.{\BBCQ}
\newblock
\APACjournalVolNumPages{URL http://yann. lecun. com/exdb/mnist}{10}{}{34}.
\PrintBackRefs{\CurrentBib}

\bibitem [\protect \citeauthoryear {%
{Liu}%
, {Pokharel}%
\BCBL {}\ \BBA {} {Principe}%
}{%
{Liu}%
\ \protect \BOthers {.}}{%
{\protect \APACyear {2007}}%
}]{%
corr}
\APACinsertmetastar {%
corr}%
\begin{APACrefauthors}%
{Liu}, W.%
, {Pokharel}, P\BPBI P.%
\BCBL {}\ \BBA {} {Principe}, J\BPBI C.%
\end{APACrefauthors}%
\unskip\
\newblock
\APACrefYearMonthDay{2007}{}{}.
\newblock
{\BBOQ}\APACrefatitle {Correntropy: Properties and Applications in Non-Gaussian
  Signal Processing} {Correntropy: Properties and applications in non-gaussian
  signal processing}.{\BBCQ}
\newblock
\APACjournalVolNumPages{IEEE Transactions on Signal
  Processing}{55}{11}{5286-5298}.
\newblock
\begin{APACrefDOI} \doi{10.1109/TSP.2007.896065} \end{APACrefDOI}
\PrintBackRefs{\CurrentBib}

\bibitem [\protect \citeauthoryear {%
Liu%
, Principe%
\BCBL {}\ \BBA {} Haykin%
}{%
Liu%
\ \protect \BOthers {.}}{%
{\protect \APACyear {2011}}%
}]{%
jp}
\APACinsertmetastar {%
jp}%
\begin{APACrefauthors}%
Liu, W.%
, Principe, J\BPBI C.%
\BCBL {}\ \BBA {} Haykin, S.%
\end{APACrefauthors}%
\unskip\
\newblock
\APACrefYear{2011}.
\newblock
\APACrefbtitle {Kernel adaptive filtering: a comprehensive introduction}
  {Kernel adaptive filtering: a comprehensive introduction}\ (\BVOL~57).
\newblock
\APACaddressPublisher{}{John Wiley \& Sons}.
\PrintBackRefs{\CurrentBib}

\bibitem [\protect \citeauthoryear {%
Lo{\`e}ve%
}{%
Lo{\`e}ve%
}{%
{\protect \APACyear {1946}}%
}]{%
loe}
\APACinsertmetastar {%
loe}%
\begin{APACrefauthors}%
Lo{\`e}ve, M.%
\end{APACrefauthors}%
\unskip\
\newblock
\APACrefYearMonthDay{1946}{}{}.
\newblock
{\BBOQ}\APACrefatitle {Fonctions al{\'e}atoires {\`a} d{\'e}composition
  orthogonale exponentielle} {Fonctions al{\'e}atoires {\`a} d{\'e}composition
  orthogonale exponentielle}.{\BBCQ}
\newblock
\APACjournalVolNumPages{La Revue Scientifique}{84}{}{159--162}.
\PrintBackRefs{\CurrentBib}

\bibitem [\protect \citeauthoryear {%
MacKay%
}{%
MacKay%
}{%
{\protect \APACyear {1992}}%
}]{%
mack}
\APACinsertmetastar {%
mack}%
\begin{APACrefauthors}%
MacKay, D\BPBI J.%
\end{APACrefauthors}%
\unskip\
\newblock
\APACrefYearMonthDay{1992}{}{}.
\newblock
{\BBOQ}\APACrefatitle {A practical Bayesian framework for backpropagation
  networks} {A practical bayesian framework for backpropagation
  networks}.{\BBCQ}
\newblock
\APACjournalVolNumPages{Neural computation}{4}{3}{448--472}.
\PrintBackRefs{\CurrentBib}

\bibitem [\protect \citeauthoryear {%
Meng%
, Zhang%
\BCBL {}\ \BBA {} Guo%
}{%
Meng%
\ \protect \BOthers {.}}{%
{\protect \APACyear {2016}}%
}]{%
q1}
\APACinsertmetastar {%
q1}%
\begin{APACrefauthors}%
Meng, X.%
, Zhang, J\BHBI W.%
\BCBL {}\ \BBA {} Guo, H.%
\end{APACrefauthors}%
\unskip\
\newblock
\APACrefYearMonthDay{2016}{}{}.
\newblock
{\BBOQ}\APACrefatitle {Quantum Brownian motion model for the stock market}
  {Quantum brownian motion model for the stock market}.{\BBCQ}
\newblock
\APACjournalVolNumPages{Physica A: Statistical Mechanics and its
  Applications}{452}{}{281--288}.
\PrintBackRefs{\CurrentBib}

\bibitem [\protect \citeauthoryear {%
Muandet%
, Fukumizu%
, Sriperumbudur%
, Sch{\"o}lkopf%
\BCBL {}\ \protect \BOthers {.}}{%
Muandet%
\ \protect \BOthers {.}}{%
{\protect \APACyear {2017}}%
}]{%
emb}
\APACinsertmetastar {%
emb}%
\begin{APACrefauthors}%
Muandet, K.%
, Fukumizu, K.%
, Sriperumbudur, B.%
, Sch{\"o}lkopf, B.%
\BCBL {}\ \BOthersPeriod {.}\end{APACrefauthors}%
\unskip\
\newblock
\APACrefYearMonthDay{2017}{}{}.
\newblock
{\BBOQ}\APACrefatitle {Kernel mean embedding of distributions: A review and
  beyond} {Kernel mean embedding of distributions: A review and beyond}.{\BBCQ}
\newblock
\APACjournalVolNumPages{Foundations and Trends{\textregistered} in Machine
  Learning}{10}{1-2}{1--141}.
\PrintBackRefs{\CurrentBib}

\bibitem [\protect \citeauthoryear {%
Nagel%
}{%
Nagel%
}{%
{\protect \APACyear {2017}}%
}]{%
nag}
\APACinsertmetastar {%
nag}%
\begin{APACrefauthors}%
Nagel, J\BPBI B.%
\end{APACrefauthors}%
\unskip\
\newblock
\APACrefYear{2017}.
\unskip\
\newblock
\APACrefbtitle {Bayesian techniques for inverse uncertainty quantification}
  {Bayesian techniques for inverse uncertainty quantification}\
  \APACtypeAddressSchool {\BUPhD}{}{}.
\unskip\
\newblock
\APACaddressSchool {}{ETH Zurich}.
\PrintBackRefs{\CurrentBib}

\bibitem [\protect \citeauthoryear {%
Neal%
}{%
Neal%
}{%
{\protect \APACyear {2012}}%
}]{%
neal}
\APACinsertmetastar {%
neal}%
\begin{APACrefauthors}%
Neal, R\BPBI M.%
\end{APACrefauthors}%
\unskip\
\newblock
\APACrefYear{2012}.
\newblock
\APACrefbtitle {Bayesian learning for neural networks} {Bayesian learning for
  neural networks}\ (\BVOL~118).
\newblock
\APACaddressPublisher{}{Springer Science \& Business Media}.
\PrintBackRefs{\CurrentBib}

\bibitem [\protect \citeauthoryear {%
Neumann%
, Wigner%
\BCBL {}\ \BBA {} Hofstadter%
}{%
Neumann%
\ \protect \BOthers {.}}{%
{\protect \APACyear {1955}}%
}]{%
vonn}
\APACinsertmetastar {%
vonn}%
\begin{APACrefauthors}%
Neumann, J.%
, Wigner, E\BPBI P.%
\BCBL {}\ \BBA {} Hofstadter, R.%
\end{APACrefauthors}%
\unskip\
\newblock
\APACrefYear{1955}.
\newblock
\APACrefbtitle {Mathematical foundations of quantum mechanics} {Mathematical
  foundations of quantum mechanics}.
\newblock
\APACaddressPublisher{}{Princeton university press}.
\PrintBackRefs{\CurrentBib}

\bibitem [\protect \citeauthoryear {%
Nguyen%
, Yosinski%
\BCBL {}\ \BBA {} Clune%
}{%
Nguyen%
\ \protect \BOthers {.}}{%
{\protect \APACyear {2015}}%
}]{%
adv}
\APACinsertmetastar {%
adv}%
\begin{APACrefauthors}%
Nguyen, A.%
, Yosinski, J.%
\BCBL {}\ \BBA {} Clune, J.%
\end{APACrefauthors}%
\unskip\
\newblock
\APACrefYearMonthDay{2015}{}{}.
\newblock
{\BBOQ}\APACrefatitle {Deep neural networks are easily fooled: High confidence
  predictions for unrecognizable images} {Deep neural networks are easily
  fooled: High confidence predictions for unrecognizable images}.{\BBCQ}
\newblock
\BIn{} \APACrefbtitle {Proceedings of the IEEE conference on computer vision
  and pattern recognition} {Proceedings of the ieee conference on computer
  vision and pattern recognition}\ (\BPGS\ 427--436).
\PrintBackRefs{\CurrentBib}

\bibitem [\protect \citeauthoryear {%
Osband%
, Blundell%
, Pritzel%
\BCBL {}\ \BBA {} Van~Roy%
}{%
Osband%
\ \protect \BOthers {.}}{%
{\protect \APACyear {2016}}%
{\protect \APACexlab {{\protect \BCnt {1}}}}}]{%
osb}
\APACinsertmetastar {%
osb}%
\begin{APACrefauthors}%
Osband, I.%
, Blundell, C.%
, Pritzel, A.%
\BCBL {}\ \BBA {} Van~Roy, B.%
\end{APACrefauthors}%
\unskip\
\newblock
\APACrefYearMonthDay{2016{\protect \BCnt {1}}}{}{}.
\newblock
{\BBOQ}\APACrefatitle {Deep exploration via bootstrapped DQN} {Deep exploration
  via bootstrapped dqn}.{\BBCQ}
\newblock
\BIn{} \APACrefbtitle {Advances in neural information processing systems}
  {Advances in neural information processing systems}\ (\BPGS\ 4026--4034).
\PrintBackRefs{\CurrentBib}

\bibitem [\protect \citeauthoryear {%
Osband%
, Blundell%
, Pritzel%
\BCBL {}\ \BBA {} Van~Roy%
}{%
Osband%
\ \protect \BOthers {.}}{%
{\protect \APACyear {2016}}%
{\protect \APACexlab {{\protect \BCnt {2}}}}}]{%
osband}
\APACinsertmetastar {%
osband}%
\begin{APACrefauthors}%
Osband, I.%
, Blundell, C.%
, Pritzel, A.%
\BCBL {}\ \BBA {} Van~Roy, B.%
\end{APACrefauthors}%
\unskip\
\newblock
\APACrefYearMonthDay{2016{\protect \BCnt {2}}}{}{}.
\newblock
{\BBOQ}\APACrefatitle {Deep exploration via bootstrapped DQN} {Deep exploration
  via bootstrapped dqn}.{\BBCQ}
\newblock
\BIn{} \APACrefbtitle {Advances in neural information processing systems}
  {Advances in neural information processing systems}\ (\BPGS\ 4026--4034).
\PrintBackRefs{\CurrentBib}

\bibitem [\protect \citeauthoryear {%
Paisley%
, Blei%
\BCBL {}\ \BBA {} Jordan%
}{%
Paisley%
\ \protect \BOthers {.}}{%
{\protect \APACyear {2012}}%
}]{%
jord}
\APACinsertmetastar {%
jord}%
\begin{APACrefauthors}%
Paisley, J.%
, Blei, D.%
\BCBL {}\ \BBA {} Jordan, M.%
\end{APACrefauthors}%
\unskip\
\newblock
\APACrefYearMonthDay{2012}{}{}.
\newblock
{\BBOQ}\APACrefatitle {Variational Bayesian inference with stochastic search}
  {Variational bayesian inference with stochastic search}.{\BBCQ}
\newblock
\APACjournalVolNumPages{arXiv preprint arXiv:1206.6430}{}{}{}.
\PrintBackRefs{\CurrentBib}

\bibitem [\protect \citeauthoryear {%
Parzen%
}{%
Parzen%
}{%
{\protect \APACyear {1962}}%
}]{%
parz}
\APACinsertmetastar {%
parz}%
\begin{APACrefauthors}%
Parzen, E.%
\end{APACrefauthors}%
\unskip\
\newblock
\APACrefYearMonthDay{1962}{}{}.
\newblock
{\BBOQ}\APACrefatitle {On estimation of a probability density function and
  mode} {On estimation of a probability density function and mode}.{\BBCQ}
\newblock
\APACjournalVolNumPages{The annals of mathematical
  statistics}{33}{3}{1065--1076}.
\PrintBackRefs{\CurrentBib}

\bibitem [\protect \citeauthoryear {%
Parzen%
}{%
Parzen%
}{%
{\protect \APACyear {1970}}%
}]{%
parze}
\APACinsertmetastar {%
parze}%
\begin{APACrefauthors}%
Parzen, E.%
\end{APACrefauthors}%
\unskip\
\newblock
\APACrefYearMonthDay{1970}{}{}.
\newblock
\APACrefbtitle {STATISTICAL INFERENCE ON TIME SERIES BY RKHS METHODS.}
  {Statistical inference on time series by rkhs methods.}\
  \APACbVolEdTR{}{\BTR{}}.
\newblock
\APACaddressInstitution{}{STANFORD UNIV CALIF DEPT OF STATISTICS}.
\PrintBackRefs{\CurrentBib}

\bibitem [\protect \citeauthoryear {%
Pearce%
, Zaki%
, Brintrup%
\BCBL {}\ \BBA {} Neel%
}{%
Pearce%
\ \protect \BOthers {.}}{%
{\protect \APACyear {2018}}%
}]{%
pearce}
\APACinsertmetastar {%
pearce}%
\begin{APACrefauthors}%
Pearce, T.%
, Zaki, M.%
, Brintrup, A.%
\BCBL {}\ \BBA {} Neel, A.%
\end{APACrefauthors}%
\unskip\
\newblock
\APACrefYearMonthDay{2018}{}{}.
\newblock
{\BBOQ}\APACrefatitle {Uncertainty in neural networks: Bayesian ensembling}
  {Uncertainty in neural networks: Bayesian ensembling}.{\BBCQ}
\newblock
\APACjournalVolNumPages{arXiv preprint arXiv:1810.05546}{}{}{}.
\PrintBackRefs{\CurrentBib}

\bibitem [\protect \citeauthoryear {%
Pradier%
, Pan%
, Yao%
, Ghosh%
\BCBL {}\ \BBA {} Doshi-Velez%
}{%
Pradier%
\ \protect \BOthers {.}}{%
{\protect \APACyear {2018}}%
}]{%
projbnn}
\APACinsertmetastar {%
projbnn}%
\begin{APACrefauthors}%
Pradier, M\BPBI F.%
, Pan, W.%
, Yao, J.%
, Ghosh, S.%
\BCBL {}\ \BBA {} Doshi-Velez, F.%
\end{APACrefauthors}%
\unskip\
\newblock
\APACrefYearMonthDay{2018}{}{}.
\newblock
{\BBOQ}\APACrefatitle {Latent Projection BNNs: Avoiding weight-space
  pathologies by learning latent representations of neural network weights}
  {Latent projection bnns: Avoiding weight-space pathologies by learning latent
  representations of neural network weights}.{\BBCQ}
\newblock
\APACjournalVolNumPages{arXiv preprint arXiv:1811.07006}{}{}{}.
\PrintBackRefs{\CurrentBib}

\bibitem [\protect \citeauthoryear {%
Principe%
}{%
Principe%
}{%
{\protect \APACyear {2010}}%
}]{%
prin}
\APACinsertmetastar {%
prin}%
\begin{APACrefauthors}%
Principe, J\BPBI C.%
\end{APACrefauthors}%
\unskip\
\newblock
\APACrefYear{2010}.
\newblock
\APACrefbtitle {Information theoretic learning: Renyi's entropy and kernel
  perspectives} {Information theoretic learning: Renyi's entropy and kernel
  perspectives}.
\newblock
\APACaddressPublisher{}{Springer Science \& Business Media}.
\PrintBackRefs{\CurrentBib}

\bibitem [\protect \citeauthoryear {%
Principe%
, Xu%
, Fisher%
\BCBL {}\ \BBA {} Haykin%
}{%
Principe%
\ \protect \BOthers {.}}{%
{\protect \APACyear {2000}}%
}]{%
ux}
\APACinsertmetastar {%
ux}%
\begin{APACrefauthors}%
Principe, J\BPBI C.%
, Xu, D.%
, Fisher, J.%
\BCBL {}\ \BBA {} Haykin, S.%
\end{APACrefauthors}%
\unskip\
\newblock
\APACrefYearMonthDay{2000}{}{}.
\newblock
{\BBOQ}\APACrefatitle {Information theoretic learning} {Information theoretic
  learning}.{\BBCQ}
\newblock
\APACjournalVolNumPages{Unsupervised adaptive filtering}{1}{}{265--319}.
\PrintBackRefs{\CurrentBib}

\bibitem [\protect \citeauthoryear {%
Rao%
}{%
Rao%
}{%
{\protect \APACyear {1958}}%
}]{%
rao}
\APACinsertmetastar {%
rao}%
\begin{APACrefauthors}%
Rao, C\BPBI R.%
\end{APACrefauthors}%
\unskip\
\newblock
\APACrefYearMonthDay{1958}{}{}.
\newblock
{\BBOQ}\APACrefatitle {Some statistical methods for comparison of growth
  curves} {Some statistical methods for comparison of growth curves}.{\BBCQ}
\newblock
\APACjournalVolNumPages{Biometrics}{14}{1}{1--17}.
\PrintBackRefs{\CurrentBib}

\bibitem [\protect \citeauthoryear {%
Rasmussen%
}{%
Rasmussen%
}{%
{\protect \APACyear {2003}}%
}]{%
gp}
\APACinsertmetastar {%
gp}%
\begin{APACrefauthors}%
Rasmussen, C\BPBI E.%
\end{APACrefauthors}%
\unskip\
\newblock
\APACrefYearMonthDay{2003}{}{}.
\newblock
{\BBOQ}\APACrefatitle {Gaussian processes in machine learning} {Gaussian
  processes in machine learning}.{\BBCQ}
\newblock
\BIn{} \APACrefbtitle {Summer School on Machine Learning} {Summer school on
  machine learning}\ (\BPGS\ 63--71).
\PrintBackRefs{\CurrentBib}

\bibitem [\protect \citeauthoryear {%
R{\'e}nyi%
\ \protect \BOthers {.}}{%
R{\'e}nyi%
\ \protect \BOthers {.}}{%
{\protect \APACyear {1961}}%
}]{%
r2}
\APACinsertmetastar {%
r2}%
\begin{APACrefauthors}%
R{\'e}nyi, A.%
\BCBT {}\ \BOthersPeriod {.}
\end{APACrefauthors}%
\unskip\
\newblock
\APACrefYearMonthDay{1961}{}{}.
\newblock
{\BBOQ}\APACrefatitle {On measures of entropy and information} {On measures of
  entropy and information}.{\BBCQ}
\newblock
\BIn{} \APACrefbtitle {Proceedings of the Fourth Berkeley Symposium on
  Mathematical Statistics and Probability, Volume 1: Contributions to the
  Theory of Statistics.} {Proceedings of the fourth berkeley symposium on
  mathematical statistics and probability, volume 1: Contributions to the
  theory of statistics.}
\PrintBackRefs{\CurrentBib}

\bibitem [\protect \citeauthoryear {%
Silverman%
}{%
Silverman%
}{%
{\protect \APACyear {2018}}%
}]{%
sil}
\APACinsertmetastar {%
sil}%
\begin{APACrefauthors}%
Silverman, B\BPBI W.%
\end{APACrefauthors}%
\unskip\
\newblock
\APACrefYear{2018}.
\newblock
\APACrefbtitle {Density estimation for statistics and data analysis} {Density
  estimation for statistics and data analysis}.
\newblock
\APACaddressPublisher{}{Routledge}.
\PrintBackRefs{\CurrentBib}

\bibitem [\protect \citeauthoryear {%
Simonyan%
\ \BBA {} Zisserman%
}{%
Simonyan%
\ \BBA {} Zisserman%
}{%
{\protect \APACyear {2014}}%
}]{%
vgg}
\APACinsertmetastar {%
vgg}%
\begin{APACrefauthors}%
Simonyan, K.%
\BCBT {}\ \BBA {} Zisserman, A.%
\end{APACrefauthors}%
\unskip\
\newblock
\APACrefYearMonthDay{2014}{}{}.
\newblock
{\BBOQ}\APACrefatitle {Very deep convolutional networks for large-scale image
  recognition} {Very deep convolutional networks for large-scale image
  recognition}.{\BBCQ}
\newblock
\APACjournalVolNumPages{arXiv preprint arXiv:1409.1556}{}{}{}.
\PrintBackRefs{\CurrentBib}

\bibitem [\protect \citeauthoryear {%
Smith%
}{%
Smith%
}{%
{\protect \APACyear {2013}}%
}]{%
uncdef1}
\APACinsertmetastar {%
uncdef1}%
\begin{APACrefauthors}%
Smith, R\BPBI C.%
\end{APACrefauthors}%
\unskip\
\newblock
\APACrefYear{2013}.
\newblock
\APACrefbtitle {Uncertainty quantification: theory, implementation, and
  applications} {Uncertainty quantification: theory, implementation, and
  applications}\ (\BVOL~12).
\newblock
\APACaddressPublisher{}{Siam}.
\PrintBackRefs{\CurrentBib}

\bibitem [\protect \citeauthoryear {%
Smola%
\ \BBA {} Sch{\"o}lkopf%
}{%
Smola%
\ \BBA {} Sch{\"o}lkopf%
}{%
{\protect \APACyear {1998}}%
}]{%
smola}
\APACinsertmetastar {%
smola}%
\begin{APACrefauthors}%
Smola, A\BPBI J.%
\BCBT {}\ \BBA {} Sch{\"o}lkopf, B.%
\end{APACrefauthors}%
\unskip\
\newblock
\APACrefYear{1998}.
\newblock
\APACrefbtitle {Learning with kernels} {Learning with kernels}\ (\BVOL~4).
\newblock
\APACaddressPublisher{}{Citeseer}.
\PrintBackRefs{\CurrentBib}

\bibitem [\protect \citeauthoryear {%
Stigler%
}{%
Stigler%
}{%
{\protect \APACyear {2005}}%
}]{%
lap}
\APACinsertmetastar {%
lap}%
\begin{APACrefauthors}%
Stigler, S\BPBI M.%
\end{APACrefauthors}%
\unskip\
\newblock
\APACrefYearMonthDay{2005}{}{}.
\newblock
{\BBOQ}\APACrefatitle {PS Laplace, Th{\'e}orie analytique des probabilit{\'e}s,
  (1812); Essai philosophique sur les probabilit{\'e}s, (1814)} {Ps laplace,
  th{\'e}orie analytique des probabilit{\'e}s, (1812); essai philosophique sur
  les probabilit{\'e}s, (1814)}.{\BBCQ}
\newblock
\BIn{} \APACrefbtitle {Landmark Writings in Western Mathematics 1640-1940}
  {Landmark writings in western mathematics 1640-1940}\ (\BPGS\ 329--340).
\newblock
\APACaddressPublisher{}{Elsevier}.
\PrintBackRefs{\CurrentBib}

\bibitem [\protect \citeauthoryear {%
Su%
, Vargas%
\BCBL {}\ \BBA {} Sakurai%
}{%
Su%
\ \protect \BOthers {.}}{%
{\protect \APACyear {2019}}%
}]{%
su}
\APACinsertmetastar {%
su}%
\begin{APACrefauthors}%
Su, J.%
, Vargas, D\BPBI V.%
\BCBL {}\ \BBA {} Sakurai, K.%
\end{APACrefauthors}%
\unskip\
\newblock
\APACrefYearMonthDay{2019}{}{}.
\newblock
{\BBOQ}\APACrefatitle {One pixel attack for fooling deep neural networks} {One
  pixel attack for fooling deep neural networks}.{\BBCQ}
\newblock
\APACjournalVolNumPages{IEEE Transactions on Evolutionary Computation}{}{}{}.
\PrintBackRefs{\CurrentBib}

\bibitem [\protect \citeauthoryear {%
Sullivan%
}{%
Sullivan%
}{%
{\protect \APACyear {2015}}%
}]{%
uncdef2}
\APACinsertmetastar {%
uncdef2}%
\begin{APACrefauthors}%
Sullivan, T\BPBI J.%
\end{APACrefauthors}%
\unskip\
\newblock
\APACrefYear{2015}.
\newblock
\APACrefbtitle {Introduction to uncertainty quantification} {Introduction to
  uncertainty quantification}\ (\BVOL~63).
\newblock
\APACaddressPublisher{}{Springer}.
\PrintBackRefs{\CurrentBib}

\bibitem [\protect \citeauthoryear {%
Theil%
\ \BBA {} Meisner%
}{%
Theil%
\ \BBA {} Meisner%
}{%
{\protect \APACyear {1980}}%
}]{%
r4}
\APACinsertmetastar {%
r4}%
\begin{APACrefauthors}%
Theil, H.%
\BCBT {}\ \BBA {} Meisner, J\BPBI F.%
\end{APACrefauthors}%
\unskip\
\newblock
\APACrefYearMonthDay{1980}{}{}.
\newblock
{\BBOQ}\APACrefatitle {Simultaneous equation estimation based on maximum
  entropy moments} {Simultaneous equation estimation based on maximum entropy
  moments}.{\BBCQ}
\newblock
\APACjournalVolNumPages{Economics Letters}{5}{4}{339--344}.
\PrintBackRefs{\CurrentBib}

\bibitem [\protect \citeauthoryear {%
Tibshirani%
}{%
Tibshirani%
}{%
{\protect \APACyear {1996}}%
}]{%
tib}
\APACinsertmetastar {%
tib}%
\begin{APACrefauthors}%
Tibshirani, R.%
\end{APACrefauthors}%
\unskip\
\newblock
\APACrefYearMonthDay{1996}{}{}.
\newblock
{\BBOQ}\APACrefatitle {A comparison of some error estimates for neural network
  models} {A comparison of some error estimates for neural network
  models}.{\BBCQ}
\newblock
\APACjournalVolNumPages{Neural Computation}{8}{1}{152--163}.
\PrintBackRefs{\CurrentBib}

\bibitem [\protect \citeauthoryear {%
Van~Loan%
\ \BBA {} Golub%
}{%
Van~Loan%
\ \BBA {} Golub%
}{%
{\protect \APACyear {1983}}%
}]{%
icd}
\APACinsertmetastar {%
icd}%
\begin{APACrefauthors}%
Van~Loan, C\BPBI F.%
\BCBT {}\ \BBA {} Golub, G\BPBI H.%
\end{APACrefauthors}%
\unskip\
\newblock
\APACrefYear{1983}.
\newblock
\APACrefbtitle {Matrix computations} {Matrix computations}.
\newblock
\APACaddressPublisher{}{Johns Hopkins University Press Baltimore}.
\PrintBackRefs{\CurrentBib}

\bibitem [\protect \citeauthoryear {%
Vapnik%
}{%
Vapnik%
}{%
{\protect \APACyear {2013}}%
}]{%
g2}
\APACinsertmetastar {%
g2}%
\begin{APACrefauthors}%
Vapnik, V.%
\end{APACrefauthors}%
\unskip\
\newblock
\APACrefYear{2013}.
\newblock
\APACrefbtitle {The nature of statistical learning theory} {The nature of
  statistical learning theory}.
\newblock
\APACaddressPublisher{}{Springer science \& business media}.
\PrintBackRefs{\CurrentBib}

\end{thebibliography}

\end{document}